\documentclass{article}
\usepackage[a4paper,margin=0.75in]{geometry}
\usepackage{natbib}
\usepackage{graphicx}
\usepackage{subcaption}
\usepackage{caption}
\usepackage{amsmath}
\usepackage{amsfonts}
\usepackage{amssymb}
\usepackage{bbm}
\usepackage{xcolor}
\usepackage{listings}
\usepackage{wrapfig}
\usepackage{mathtools}
\usepackage{algorithm}
\usepackage{algpseudocode}

\newtheorem{Lemma}{Lemma}

\newtheorem{Assum}{Assumption}
\newtheorem{Thm}{Theorem}
\newtheorem{Proposition}{Proposition}

\makeatletter
\def\BState{\State\hskip-\ALG@thistlm}
\makeatother

\DeclareMathOperator*{\argmin}{arg\,min}

\algnewcommand\algorithmicswitch{\textbf{switch}}
\algnewcommand\algorithmiccase{\textbf{case}}
\algnewcommand\algorithmicassert{\texttt{assert}}
\algnewcommand\Assert[1]{\State \algorithmicassert(#1)}

\algdef{SE}[SWITCH]{Switch}{EndSwitch}[1]{\algorithmicswitch\ #1\ \algorithmicdo}{\algorithmicend\ \algorithmicswitch}%
\algdef{SE}[CASE]{Case}{EndCase}[1]{\algorithmiccase\ #1}{\algorithmicend\ \algorithmiccase}%
\algtext*{EndSwitch}%
\algtext*{EndCase}%

\title{A linear time method for the detection of point and collective anomalies}
\author{Alexander T. M. Fisch, Idris A. Eckley, and Paul Fearnhead \\
\small{Lancaster University, United Kingdom}}  
\begin{document} 
\maketitle 	
 \begin{abstract}
 	The challenge of efficiently identifying anomalies in data sequences is an important statistical problem that now arises in many applications. Whilst there has been substantial work aimed at making statistical analyses robust to outliers, or point anomalies, there has been much less work on detecting anomalous segments, or collective anomalies, particularly in those settings where point anomalies might also occur. In this article, 
we introduce Collective And Point Anomalies (CAPA), a computationally efficient approach that is suitable when collective anomalies are characterised by either a change in mean, variance, or both, and distinguishes them from point anomalies. Theoretical results establish the consistency of CAPA at detecting collective anomalies and, as a by-product,  the consistency of a popular penalised cost based change in mean and variance detection method. Empirical results show that CAPA has close to linear computational cost as well as being more accurate at detecting and locating collective anomalies than other approaches. We demonstrate the utility of CAPA through its ability to detect exoplanets from light curve data from the Kepler telescope.		
 \end{abstract}

\noindent%
{\it Keywords:}  Epidemic Changepoints, Exoplanets, Dynamic Programming, Outliers, Robust Statistics. 
 	
\section{Introduction}\label{sec:Intro}
 	
Anomaly detection is an area of considerable importance for many time series applications, such as fault detection or fraud prevention, and has been subject to increasing attention in recent years. See \cite{chandola2009anomaly} and  \cite{pimentel2014review} for comprehensive reviews of the area. As \cite{chandola2009anomaly} highlight, anomalies can fall into one of three categories: global anomalies, contextual anomalies, or collective anomalies. Global anomalies and contextual anomalies are defined as single observations which are outliers with regards to the complete dataset and their local context respectively. Conversely, collective anomalies are defined as sequences of observations which are not anomalous when considered individually, but together form an anomalous pattern. 

A number of different approaches can be taken to detect point (i.e.\ contextual and/or global) anomalies. These are observations that do not conform with the pattern of the data. Hence, the problem of detecting point anomalies can be reformulated as inferring the general pattern of the data in a manner that is robust to anomalies. The field of robust statistics offers a wide range of methods aimed at this problem. For instance, \cite{rousseeuw1984} proposed $S$-estimators to robustly estimate the mean and variance. These estimators were later extended to a multivariate setting by \cite{rousseeuw1985}. A wide variety of robust time series models also exist. For example, \cite{muler2009robust} proposed a robust ARMA model, \cite{muler2002robust} a robust ARCH model, and \cite{muler2008robust} a robust GARCH model. A robust non-parametric method, which decomposes time series into trend, seasonal component, and residual was proposed by \cite{cleveland1990stl}. 

The machine learning community has also provided a rich corpus of work for the detection of point anomalies. Commonly used methods include nearest neighbour based approaches, such as the local outlier factor \citep{LOF}, and information theoretical methods such as the one introduced by \cite{guha2016robust}. It is beyond the scope of this paper to review them all. Instead we refer to excellent reviews that can be found in \cite{chandola2009anomaly} and \cite{pimentel2014review}.

One common drawback of several point anomaly approaches is their inability to detect anomalous segments, or collective anomalies. Such features are of significance  in many applications. One example is the analysis of brain imaging data, where periods in which the brain activity deviates from the pattern of the rest state have been associated with sudden shocks \citep{Epidemic:Aston}. Another example is in detecting regions of the genome with unusual copy number \citep{bardwell2017bayesian,siegmund2011detecting,zhang2010detecting}, with such copy number variation being associated with diseases such as cancer \citep{jeng2012simultaneous}. 

Existing work on the detection of collective anomalies can be found in both the statistics and machine learning literature. On the statistical side, hidden Markov models have been proposed, which assume that a hidden state chain determines whether the data produced is anomalous or typical \citep{smyth1994markov}. The underlying assumption that anomalous segments share one or multiple common behaviours is very attractive for some applications, such as brain imaging, where it can be assumed that there is a finite number of states, but can be a constraint in others. Hidden Markov models also suffer from the fact that they are not robust to global anomalies 
Moreover, they tend to be slow to fit, which is an important disadvantage in many modern, big-data applications.  
More generally, it may be of interest to identify anomalous segments and point anomalies simultaneously. This is the setting that we consider in this article.

\begin{figure}
	\begin{subfigure}[b]{0.5\linewidth}
		\includegraphics[width=0.9\linewidth]{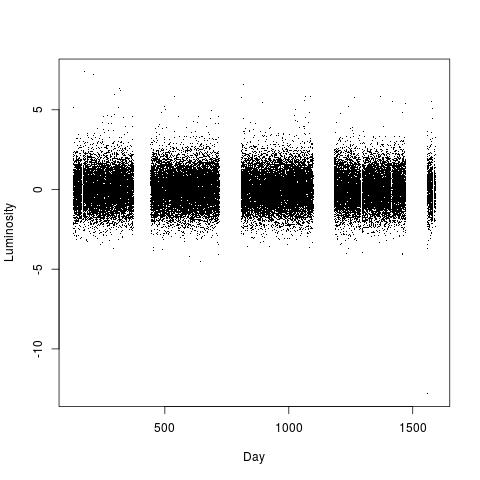}
		\caption{Full data}
	\end{subfigure}
	\begin{subfigure}[b]{0.5\linewidth}
		\includegraphics[width=0.9\linewidth]{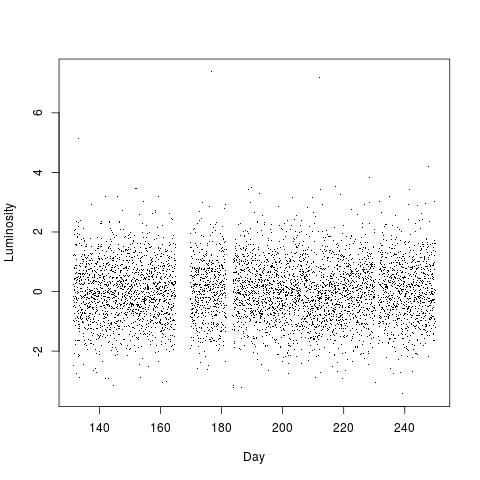}
		\caption{Subset (120 days)}
	\end{subfigure}
	\caption{ Light curve of Kepler 1132, obtained at approximately 30 minute intervals. Missing values are due to periods in which the star was not observed. Note the presence of a point anomaly on day 1550 and the fact that no transit signature is apparent to the eye. This remains true after zooming in on a 120 day subset of the data, despite the known presence of Kepler 1132-b, an exoplanet orbiting this star every 62.9 days}
	\label{fig:Kepler1132First} 
\end{figure}

Consider, by way of example, the problem of detecting exoplanets via the so called transit method first proposed by \cite{struve1952proposal}. The luminosity of a star is measured at regular intervals, with the aim of detecting segments of reduced luminosity. These indicate the transit of a planet \citep{Transit} and can naturally be interpreted as collective anomalies. The light curves are typically preprocessed \citep{Preprocessing} and both the raw and whitened light curves can be accessed online. We have included the whitened light curve of the star Kepler 1132 in Figure \ref{fig:Kepler1132First} to illustrate the nature of this type of data. We note the presence of a global anomaly on day 1550 and the noisy nature of the data. These make the detection of transits challenging given the weak signal induced by planetary transits. Indeed, even the transit of Jupiter past the sun reduces the latter's luminosity by only 1\% \citep{Transit}. The signal to noise ratio can be improved by exploiting the periodic nature of transits. Indeed, as we shall see later, 
the transits signal is visible by eye when the orbital period of a planet is known. However, the period of exoplanets (if any are present) is not known a priori. Consequently, the ability to automate the analysis of a large number of light curves would be beneficial, given that there are currently 40 million light curves, similar to that shown in Figure 1, that have been gathered and need analysing.

This article makes two main contributions. The first is the introduction of an inference procedure that allows for the identification of \textbf{C}ollective \textbf{A}nd \textbf{P}oint \textbf{A}nomalies (CAPA). 
Secondly, we establish finite sample consistency results not only for CAPA, but also for a commonly used penalised cost based method aimed at detecting changes in mean and variance. This setting presents significant additional technical challenge compared to the change in mean setting, to which most existing theoretical results apply. 

The article is organised as follows: We begin by introducing a parametric model with epidemic changes in Section \ref{sec:Model}. This provides a general framework for collective anomalies, the location of which we infer by minimising a penalised cost. 
In Section \ref{sec:Inference}, we introduce an algorithm which minimises an approximation to the penalised cost based on a robust estimate of the parameter of the typical distribution. This approximation can be minimised by a dynamic program. 

We then present theoretical results in Section \ref{sec:Theory}. Specifically, we introduce a proof of consistency for the detection of joint classical changes in mean and variance using a penalised cost approach, which is of independent interest. We then prove that CAPA consistently estimates the number and location of collective anomalies, despite the simplicity of the approach used for the estimation of the parameters of the typical distribution. The section ends with a discussion of penalties. We then compare CAPA to other methods in a simulation study in Section \ref{sec:Simulations} and show that it outperforms them, especially in the presence of point anomalies. We conclude the paper by demonstrating in Section \ref{sec:Application} that CAPA can be used to detect Kepler 1132-b, an exoplanet which orbits Kepler 1132 \citep{Kepler}. The proofs of all propositions and lemmata can be found in the appendix and the supplementary material respectively. Both can be found at the end of the paper. 
CAPA has been implemented in the R package ``anomaly" and is available on CRAN.
 	
\section{A Modelling Framework for Collective Anomalies}\label{sec:Model}

We assume that the data follow a parametric model and model collective anomalies as epidemic changes in the model parameters. Whilst, in practice, it is unlikely that the distribution of the data in an anomalous segment will belong to the same family of distributions as the distribution of the typical data, it can nevertheless be expected that a set of parameters different from the typical distribution's will offer a better fit. We say that data $\textbf{x}_1,...,\textbf{x}_n$ follow a parametric epidemic changepoint model if $\textbf{x}_t$ has probability density function $f(\textbf{x}_t,\theta(t))$ and
\begin{align*}
\theta(t) = \begin{cases}
\theta_1  & s_1 < t \leq e_1, \\
&\vdots \\
\theta_K  & s_K < t \leq e_K, \\
\theta_0  & \text{otherwise},
\end{cases}
\end{align*}
where $\theta_0$ is the usually unknown parameter of the typical distribution, from which the model deviates during the $K$ anomalous segments $(s_1,e_1)$,...,$(s_K,e_K)$. We assume these windows do not overlap, i.e.\ $e_1 \leq s_2, ... , e_{K-1} \leq s_K$. Note that fitting an epidemic changepoint requires only one new set of parameters for $\theta$, since the typical parameter is shared across the non-anomalous segments. This compares favourably with the two additional sets of parameters for $\theta$ introduced when an epidemic changepoint is fitted using two classical changepoints. We therefore gain statistical power. This gain is particularly important when $\theta$ is high dimensional. 

It is possible to infer the number and location of epidemic changes by choosing $\tilde{K}$, $(\tilde{s}_1,\tilde{e}_1)$,...,$(\tilde{s}_{\tilde{K}},\tilde{e}_{\tilde{K}})$, and $\tilde{\theta}_{0}$, which minimise the penalised cost
\begin{equation} \label{eq:TOBEMAXED}
\sum_{t \notin \cup[\tilde{s}_i+1,\tilde{e}_i ]}   \mathcal{C}(\textbf{x}_t ,\tilde{\theta}_{0} )  
+ \sum_{j = 1}^{\hat{K}} \left[  \min_{\tilde{\theta}_{j}}\left(\sum_{t = \tilde{s}_{j}+1}^{\tilde{e}_{j}}  
\mathcal{C}(\textbf{x}_t ,\tilde{\theta}_{j} )\right)  + \beta  \right],
\end{equation}
subject to $e_i - s_i \geq \hat{l}$, where $\hat{l}$ is the minimum segment length for an appropriate cost function $\mathcal{C}(x,\theta)$ and a suitable penalty $\beta$. For example, $\mathcal{C}(x,\theta)$ could be defined as the negative log-likelihood of $x$ under the parametric model using parameter $\theta$. A common choice for the penalty $\beta$ would then be $C\log(n)$ \cite{yao1988estimating,killick2012optimal,fryzlewicz2014wild}, where the constant $C$ depends on the model considered.

Using the formulation in (\ref{eq:TOBEMAXED}), we can infer the location of joint epidemic changes in mean and variance by minimising the penalised cost related to the negative log-likelihood of Gaussian data. In this case $\theta = (\mu,\sigma^2)$ contains both the mean and variance and we estimate $K$, $s_1,...,s_K$, and $e_1,...,e_K$ by minimising 
\begin{equation} \label{eq:TOBEMAXEDnorm}
\sum_{t \notin \cup[\tilde{s}_i+1,\tilde{e}_i ]}  \left[\log(\sigma^2_0) + \left(\frac{x_t-\mu_0}{\sigma_0}\right)^2 \right]
+ \sum_{j = 1}^{\tilde{K}} \left[  
(\tilde{e}_{j}-\tilde{s}_j) \left( \log \left( \frac{\sum_{t = \tilde{s}_{j}+1}^{\tilde{e}_{j}}(x_t - \bar{x}_{(\tilde{s}_{j}+1):\tilde{e}_{j}})^2}{(\tilde{e}_{j}-\tilde{s}_j)}\right) + 1\right)  + \beta  \right],
\end{equation}
subject to $\tilde{e}_i - \tilde{s}_i \geq 2$, i.e. a minimum segment length of 2, to account for the fact that $\theta$ is two dimensional. 

It is well known that many changepoint detection methods struggle in the presence of point anomalies in the data and tend to fit two changepoints around each of them (\cite{fearnhead2016changepoint}). An approach based on minimising the above cost function is not intrinsically immune to it. However, we can modify the model by allowing epidemic changes, in variance only, of length one to address this issue. We therefore choose $\tilde{K}$, $(\tilde{s}_1,\tilde{e}_1)$,...,$(\tilde{s}_{\tilde{K}},\tilde{e}_{\tilde{K}})$, $\mu_{0}$, $\sigma_0$, as well as the set of point anomalies $O \subset \{1,...,n\}$, which minimise the modified penalised cost
\begin{align*} \label{eq:TOBEMAXEDnormoutlier}
\sum_{t \notin \cup[\tilde{s}_i+1,\tilde{e}_i ] \cup O}  \left[\log(\sigma^2_0) + \left(\frac{x_t-\mu_0}{\sigma_0}\right)^2 \right]
+  \sum_{t \in O} \left[ \log\left((x-\mu_0)^2\right) + 1 + \beta'\right] +
\\ \sum_{j = 1}^{\hat{K}} \left[  
(\tilde{e}_{j}-\tilde{s}_j) \left( \log \left( \frac{\sum_{t = \tilde{s}_{j}+1}^{\tilde{e}_{j}}(x_t - \bar{x}_{(\tilde{s}_{j}+1):\tilde{e}_{j}})^2}{(\tilde{e}_{j}-\tilde{s}_j)}\right) + 1\right)  + \beta  \right],
\end{align*}
where $\beta'$ is a penalty smaller than $\beta$. This modification ensures that it is now cheaper to fit an outlier as an epidemic changepoint in variance only than as a full epidemic change. Consequently, the method becomes robust against point anomalies, fitting epidemic changes only around true collective anomalies. This modification has the added benefit that it allows the algorithm to detect and distinguish between point and collective anomalies. This property is important for a range of applications in which collective and point anomalies have different interpretations (see Section \ref{sec:Application} for an example).

\section{Inference for Collective Anomalies}\label{sec:Inference}

\begin{algorithm}
	\caption{CAPA Algorithm (No Pruning)}\label{CAPA}
	\begin{footnotesize}
		\begin{tabular}[h]{ll}
			{\bf Input:} & A set of observations of the form, $(x_1,x_2,\ldots,x_n)$ where $x_i \in \mathbb{R}$. \\ & Penalty constants $\beta$ and $\beta'$ for the introduction of a collective and a point anomaly respectively \\ & A minimum segment length $l \geq 2$\\ {\bf Initialise:} & Set $C(0)=0$, $Anom(0)=NULL$. 
		\end{tabular}

		\begin{algorithmic}[1]
			\State $\hat{\mu} \gets MEDIAN(x_1,x_2,\ldots,x_n)$ \Comment{Obtain robust estimates of the mean and variance}
			\State $\hat{\sigma} \gets IQR(x_1,x_2,\ldots,x_n)$
			\For{$i \in \{1,...,n\} $}
			\State $x_i \gets \frac{x_i - \hat{\mu} }{\hat{\sigma}}$ \Comment{Centralise the data}
			\EndFor
			\For{$m \in \{1,...,n\} $}
			\State $C_1(m) \gets \min_{0 \leq k \leq m - l} \Big[ 
			C(k) + (m -k) \left[\log \left( \frac{1}{m-k}\sum_{t=k+1}^{m}\left(x_t - \bar{x}_{(k+1):m}\right)^2\right) + 1\right] + \beta \Big] $
			\Comment{Collective Anom.}
			\State $ s \gets \argmin_{0 \leq k \leq m - l} \Big[ 
			C(k) + (m -k) \left[\log \left( \frac{1}{m-k}\sum_{t=k+1}^{m}\left(x_t - \bar{x}_{(k+1):m}\right)^2\right) + 1\right] + \beta \Big] $
			\State $C_2(m) \gets 
			C(m-1) + x_m^2 $ \Comment{No Anomaly}
			\State $C_3(m) \gets  
			C(m-1) + 1 + \log \left( \gamma  + x_m ^2 \right) + \beta'
			\Big], $ \Comment{Point Anomaly}
			\State $C(m) \gets \min\left[C_1(m),C_2(m),C_3(m)\right] $
			\Switch{$\argmin\left[C_1(m),C_2(m),C_3(m)\right]$} \Comment{Select type of anomaly giving the lowest cost}
			\Case{$1$} : $Anom(m) \gets [Anom(s),(s+1,m) ]$ 
			\EndCase
			\Case{$2$} : $Anom(m) \gets Anom(m-1) $
			\EndCase
			\Case{$3$} : $Anom(m) \gets[ Anom(m-1),(m)] $
			\EndCase
			\EndSwitch
			\EndFor
		\end{algorithmic}
		{\bf Output} The points and segments recorded in $Anom(n)$
	\end{footnotesize}
\end{algorithm}

We now turn to consider the problem of minimising the penalised cost we introduced in the previous section. Unlike in the classical changepoint problem considered by \cite{jackson2005algorithm}, the penalised cost given by equation (\ref{eq:TOBEMAXED}) can not be minimised using a dynamic program, since the parameter $\theta_{0}$ is shared across multiple segments and typically unknown. We therefore use robust statistics to obtain an estimate $\hat{\theta}_{0}$ for $\theta_{0}$. Such robust estimates can be obtained for a variety of models (\cite{MAINROBUSTSTATSBOOK} \cite{jurevckova2005robust}). For example, the median, $M$-estimators, or the clipped mean can be used to robustly estimate the mean. The inter quantile range, the median absolute deviation, or the clipped standard deviation can be use to estimate the variance. Robust regression is available to estimate the parameters of AR models.  

Having obtained $\hat{\theta}_{0}$, we then minimise 
\begin{equation*}
\sum_{t \notin \cup[\hat{s}_i+1,\hat{e}_i ]}  \mathcal{C}(\textbf{x}_t ,\hat{\theta}_{0} ) 
+ \sum_{j = 1}^{\hat{K}} \left[  \min_{\hat{\theta}_j}\left(\sum_{t = \hat{s}_{j}+1}^{\hat{e}_{j}}  
\mathcal{C}(\textbf{x}_t ,\hat{\theta} _{j} )\right)  + \beta  \right],
\end{equation*}
as an approximation to (\ref{eq:TOBEMAXED}). Since it can be expected that most data belongs to the typical distribution, $\hat{\theta}_{0} $ should be close to $\theta_{0}$. One might therefore expect that using this estimate will have little impact on the performance of the method, which we also show theoretically for joint changes in mean and variance in Section \ref{sec:ECPcons}. 

The approximation to the penalised cost can be minimised exactly by solving the dynamic programme
\begin{align}
C(m) = \min_{0 \leq k \leq m - \hat{l} } \Big[ 
C(m-1) +  \mathcal{C}(\textbf{x}_m , \hat{\theta} _0) , 
C(k) + \min_{\hat{\theta}}\left(\sum_{t=k+1}^{m}  \mathcal{C}(\textbf{x}_t , \hat{\theta})\right) + \beta
\Big],
\end{align}
where $C(m)$ is the cost of the most efficient partition of the first $m$ observations and $C(0) = 0$. For example, solving the dynamic programme
\begin{align*}
C(m) = \min_{0 \leq k \leq m - 2} \Big[ 
&C(m-1) +  \log(\hat{\sigma}_0^2) + \left(\frac{x_m - \hat{\mu}_0}{\hat{\sigma}_0}\right)^2 , \\
&C(k) + (m -k) \left[\log \left( \frac{1}{m-k}\sum_{t=k+1}^{m}\left(x_t - \bar{x}_{(k+1):m}\right)^2\right) + 1\right] + \beta
\Big],
\end{align*}
approximately minimises the penalised cost for joint epidemic changes in mean and variance defined in equation (\ref{eq:TOBEMAXEDnorm}). Similarly, we can minimise its point anomaly robust analogue by solving the dynamic programme
\begin{align*}
C(m) = \min_{0 \leq k \leq m - 2} \Big[ 
&C(m-1) +  \log(\hat{\sigma}_0^2) + \left(\frac{x_m - \hat{\mu}_0}{\hat{\sigma}_0}\right)^2 , \\
&C(k) + (m -k) \left[\log \left( \frac{1}{m-k}\sum_{t=k+1}^{m}\left(x_t - \bar{x}_{(k+1):m}\right)^2\right) + 1\right] + \beta, \\
&C(m-1) + 1 + \log \left( \gamma \hat{\sigma}_0^2 + (x_m - \hat{\mu}_0)^2 \right) + \beta'
\Big], 
\end{align*}
where $\gamma$ is a small constant ensuring that the argument of the logarithm will be larger than 0 (see Algorithm \ref{CAPA} for pseudocode). Adding the $\gamma \hat{\sigma}_0^2$ term is necessary when order statistics are used to obtain $\hat{\mu}_0$. Assuming that the observations $x_t$ are independent and Normal, all sums  $\sum_{t=m-k+1}^{m}\left(x_t - \bar{x}_{(m-k+1):m}\right)^2$ will be non-zero with probability 1, meaning that in theory such a correction is not necessary for the other logarithmic term. In practice, observations are of finite precision and adding $\gamma \hat{\sigma}_0^2$ to the argument of the other logarithmic term, with $\gamma$ set to the level of rounding should be considered. Setting $\gamma > e^{-\beta'}$  has the added benefit of ensuring that no inliers will be fitted as point anomalies, as shown by Proposition \ref{Prop:TPCONTROL} in Section \ref{sec:Theory}.

Solving the full dynamic program is at least $O(n^2)$. However, we can prune the solution space by borrowing ideas from \cite{killick2012optimal}, provided the loss function is such that adding a free changepoint will not increase the cost -- a property which holds for many commonly used cost functions such as the negative log-likelihood. Indeed, the following proposition holds: 

\begin{Proposition}\label{Prop:PRUNING}
	Let the cost function $\mathcal{C}(\cdot,\cdot)$ be such that 
	\begin{equation*}
	\sum_{t=a}^{c}\mathcal{C}(\textbf{x}_t , \hat{\theta}_{a:c}) \geq \sum_{t=a}^{b-1}\mathcal{C}(\textbf{x}_t , \hat{\theta}_{a:(b-1)}) + 
	\sum_{t=b}^{c}\mathcal{C}(\textbf{x}_t , \hat{\theta}_{b:c})
	\end{equation*}
	holds for all $a$, $b$, and $c$ such that $a + \hat{l} \leq b < c -  \hat{l}$. Then, if 
	\begin{equation*}
	C(k) + \sum_{t=k}^{m}  \mathcal{C}(\textbf{x}_t , \hat{\theta}) \geq C(m)
	\end{equation*}
	holds for some $k < m - \hat{l}$, we can disregard $k$ for all future steps $m' \geq m + \hat{l}$ of the dynamic programme.
\end{Proposition}

\noindent \textbf{Proof}: Please see the Appendix. Note that the time after which an option can be discarded also depends on the minimum segment length, something not considered by \cite{killick2012optimal}. 

This results enables us to reduce the computational cost. In practice, we found that it was close to $O(n)$ for the detection of joint epidemic changes in mean and variance when the number of true epidemic changes increased linearly with the number of observations. 

\section{Theory for Joint Changes in Mean and Variance}\label{sec:Theory}

We now introduce some theoretical results fo CAPA. In particular, we establish the consistency of CAPA at detecting collective anomalies and demonstrate that it can be viewed as a corollary of the consistency of a statistical procedure minimising a penalised cost function to detect classical changepoints. Consequently, we will begin by proving that method's consistency for the detection of changes in mean and variance in Section \ref{sec:OPcons}. To the best of our knowledge, no such result exists in the literature, which makes this proof of independent interest. We then proceed to proving the consistency of CAPA at detecting collective anomalies in Section \ref{sec:ECPcons}. Like other cost function based approaches, CAPA is significantly affected by the choice of penalties. We therefore conclude this section by discussing suitable choices for this important hyper parameter in Section \ref{sec:Penalties}. The proofs of all theorems and propositions stated in this section can be found in the appendix and supplementary material respectively.

\subsection{Consistency of Classical Changepoint Detection}\label{sec:OPcons}

Consider the sequence $x_1,...,x_n \in \mathbb{R}^n$ which is normally distributed with $K \in \mathbb{N}$ changepoints. The sequence therefore satisfies
\begin{align}\label{eq:modeltheory1}
x_t = \mu(t) + \sigma(t)\eta_t, 
\; \; 
\text{for}
\; \; \; \; \; \; 
\eta_t \stackrel{i.i.d.}{\sim} N(0,1) 
\; \; \; \; \; \; 
\text{and}
\; \; \; \; \; \; 
(\mu(t), \sigma(t)^2) = \begin{cases}
(\mu_1,\sigma_1^2)  & t_{0} + 1 \leq t \leq t_1, \\
&\vdots  \\
(\mu_{K+1},\sigma_{K+1}^2)  & t_{K} + 1 \leq t \leq t_{K+1}. \\
\end{cases}
\end{align}
Here $0 = t_0 \leq ... \leq t_{K+1} = n$ denote the start of the series, the $K$ changepoints, and the end of the series. Changes in mean and variance can be of varying strength. To quantify this, we define the signal strength $\triangle_{\sigma,k}$ of the change in variance at the $k$th changepoint to be 
\begin{equation*}
\triangle_{\sigma,k}^2 = \left(\sqrt{\frac{\sigma_k}{\sigma_{k+1}}} - \sqrt{\frac{\sigma_{k+1}}{\sigma_{k}}}\right)^2 =  \frac{\sigma_k}{\sigma_{k+1}} + \frac{\sigma_{k+1}}{\sigma_{k}} - 2.
\end{equation*}
We note that $\triangle_{\sigma,k}^2$ is equal to 0 if, and only if, $\sigma_{k+1} = \sigma_{k}$. We also define the signal strength $\triangle_{\mu,k}$ of change in mean at the $k$th changepoint to be
\begin{equation*}
\triangle_{\mu,k} = \frac{|\mu_k-\mu_{k+1}|}{\sqrt{\sigma_k\sigma_{k+1}}}.
\end{equation*}
Note that these two quantities can be combined into a global measure of signal strength
\begin{equation*}
\triangle_k = \log \left(1 + \frac{1}{2}\triangle_{\sigma,k}^2 + \frac{1}{4}\triangle_{\mu,k}^2\right)
\end{equation*}
for the $k$th change (see Lemma \ref{lemma:Mixlemma} in the Appendix for details).

We now define the penalised cost $\tilde{\mathcal{C}}(x_{i:j}, \tau',\alpha)$ of data $x_{i:j}$ under partition $\tau' = \{i-1,\hat{t}_1',...,\hat{t}_{\hat{K}'}',j \}$ to be
\begin{equation*}
\tilde{\mathcal{C}}(x_{i:j}, \tau',\alpha) = \sum_{k = 0}^{\hat{K}'} \tilde{\mathcal{C}}(x_{(\hat{t}_k+1):\hat{t}_{k+1}}) +  \hat{K}'\alpha\log(n)^{1+\delta},
\end{equation*}
for $\delta, \alpha > 0$. Here $\alpha\log(n)^{1+\delta}$ corresponds to the commonly used strengthened SIC-type penalty (\cite{fryzlewicz2014wild, li2016fdr}) for introducing an additional changepoint.  The cost of a segment $x_{a:b}$ is set to be
\begin{equation*}
\tilde{\mathcal{C}}(x_{a:b}) = \tilde{\mathcal{C}}(x_{a:b},\{a-1,b\}) = 
(b-a+1)\left( \log \left( \frac{\sum_{a}^{b}(x_t - \bar{x}_{a:b})^2}{b-a+1}\right) +1\right),
\end{equation*}
which is similar to the segment cost used to infer the location of epidemic changes in mean and variance. Since this leaves two parameters to fit, we impose a minimum segment length of two for all partitions.

We also introduce the following assumption which ensures that the changepoints are sufficiently spaced apart to allow for their detection:
\begin{Assum} \label{ass:TheoryClassic}
	There exists some $\tilde{\delta} >0$ such that 
	\begin{equation*}
	t_k - t_{k-1} \geq \frac{\log(n) ^ {1+\delta+\tilde{\delta}}}{\min(\triangle_k,\triangle_k^2,\triangle_{k-1},\triangle_{k-1}^2)}
	\end{equation*} 
\end{Assum}
Then, the following consistency result holds: 
\begin{Thm} \label{Thm:ConsistencyCP}
	 Assume that observations $x_1,...,x_n$ follow the distribution specified in Equation \ref{eq:modeltheory1} and that Assumption \ref{ass:TheoryClassic} holds. Let $\hat{K}$ and $\hat{t}_1,...,\hat{t}_{\hat{K}}$ be the number and locations of changepoints inferred by minimising the penalised cost function $\tilde{\mathcal{C}}(x_{1:n}, \tau,\alpha)$. Then $\forall \epsilon >0$ there exist constants $A(\alpha,\epsilon)$, $B(\alpha,\tilde{\delta},\delta,\epsilon)$, and $C$ such that
	\begin{equation*}
	\mathbb{P}\left( \hat{K} = K , \;\; |\hat{t}_{k} - t_{k}| < \frac{A(\alpha,\epsilon)}{\min(\triangle_{k},\triangle_{k}^2)}  \log(n)^{1+\delta} \;\;\;\; 1 \leq k \leq K \right)  \geq 1 - Cn^{-\epsilon}
	\end{equation*}
	holds for all $n \geq B(\alpha,\tilde{\delta},\delta,\epsilon)$.
\end{Thm}

Note that the strengthened SIC was used as penalty in order to simplify the exposition of the proof. A very similar proof can be used to show that a $\alpha\log(n)$ type penalty for a sufficiently large $\alpha$ also achieves consistency.

\subsection{Consistency of CAPA}\label{sec:ECPcons}

The results we obtained in the previous section can be extended to prove the consistency of CAPA for the detection of joint epidemic changes in mean and variance. As in the previous section, consider data $x_1,...,x_n$ which is of the form $x_t = \mu(t) + \sigma(t)\eta_t$, where $\eta_t \sim N(0,1)$. Since we now assume epidemic changes, we have
\begin{equation*}
(\mu(t), \sigma(t)^2) = \begin{cases}
(\mu_1,\sigma_1^2)  & s_1 < t \leq e_1, \\
&\vdots \\
(\mu_K,\sigma_K^2)  & s_K < t \leq e_K, \\
(\mu_0,\sigma_0^2)  & \text{otherwise}.
\end{cases}
\end{equation*}
Here, $\mu_0$ and $\sigma_0^2$ are the typical mean and variance respectively and $K$ is the number of epidemic changepoints. The variables $s_k$ and $e_k$ denote the starting and end point of the $k$th anomalous window respectively. Treating the $s_k$ and $e_k$ like classical changepoints allows us to extend the definitions of $\triangle_\sigma$ and $\triangle_\mu$, and therefore $\triangle$, to the epidemic changepoint model by setting

\begin{equation*}
\triangle_{\sigma,k}^2 = \frac{\sigma_k}{\sigma_{0}} + \frac{\sigma_{0}}{\sigma_{k}} - 2 \;\;\;\;\;\;\; \text{and}  \;\;\;\;\;\;\; \triangle_{\mu,k} = \frac{|\mu_k-\mu_{0}|}{\sqrt{\sigma_k\sigma_{0}}}.
\end{equation*}

We make the following assumptions

\begin{Assum} \label{ass:TheoryMV}
	.\\
	a) There exists some $\tilde{\delta} >0$ such that for $1\leq k \leq K$
	\begin{align*}
	e_k - s_{k} &\geq \frac{\log(n) ^ {1+\delta+\tilde{\delta}}}{\min(\triangle_k,\triangle_k^2)}, \\
	s_{k+1} - e_{k} &\geq \frac{\log(n) ^ {1+\delta+\tilde{\delta}}}{\min(\triangle_k,\triangle_k^2,\triangle_{k+1},\triangle_{k+1}^2)}.
	\end{align*}
	b)  $e_k - s_{k} \leq \sqrt{n}$ for $1\leq k \leq K$.
\end{Assum}
Assumption \ref{ass:TheoryMV}a is analogous to Assumption \ref{ass:TheoryClassic}. Assumption \ref{ass:TheoryMV}b is only needed when the parameters of the typical distribution are unknown and guarantee that the robust estimates of the typical parameter will converges quickly enough to the ground truth. The following consistency result then holds for a partition $(\hat{s}_1,\hat{e}_1,...,\hat{s}_{\hat{K}},\hat{e}_{\hat{K}})$ inferred by CAPA using a minimum segment length of 2 and $\alpha\log(n)^{1+\delta}$ as penalty for both point anomalies and epidemic changepoints.
\begin{Thm} \label{Thm:Consistency}
	Let $(\hat{s}_1,\hat{e}_1,...,\hat{s}_{\hat{K}},\hat{e}_{\hat{K}})$ be the partition inferred by CAPA on observations $x_1,...,x_n$ using a minimum segment length of 2 and $\alpha\log(n)^{1+\delta}$ as penalty for both point anomalies and epidemic changepoints. If $x_1,...,x_n$ follow the distribution specified above and Assumption \ref{ass:TheoryMV} holds, then $\forall \epsilon >0$ there exist constants $A(\alpha,\epsilon)$, $B(\alpha,\tilde{\delta},\delta,\epsilon)$, and $C$ such that
	\begin{equation*}
	\mathbb{P}\left( \hat{K} = K ,\;\; |\hat{e}_k - e_k| < \frac{A(\alpha,\epsilon)}{\min(\triangle_{k},\triangle_{k}^2)}  \log(n)^{1+\delta} ,\;\; |\hat{s}_k - s_k| < \frac{A(\alpha,\epsilon)}{\min(\triangle_{k},\triangle_{k}^2)}  \log(n)^{1+\delta} \;\;\;\; 1 \leq k \leq K \right)  \geq 1 - Cn^{-\epsilon}
	\end{equation*}
	holds for all $n \geq B(\alpha,\tilde{\delta},\delta,\epsilon)$.
\end{Thm}

As in Theorem \ref{Thm:ConsistencyCP}, the strengthened SIC was used as penalty to simplify the exposition of the proof. A very similar result could be derived to show that a $\alpha\log(n)$ type penalty for a sufficiently large $\alpha$ achieves consistency. This result suggests that CAPA, fits collective anomalies as such, despite being able to fit them as a mixture of point anomalies and data belonging to the typical distribution. It is also possible to relax Assumption \ref{ass:TheoryMV}b to $e_k - s_{k} \leq D\sqrt{n}$ for some constant $D$, with the constants $A$, $B$, and $C$ then also depending on $D$. 

\subsection{Penalties}\label{sec:Penalties}

We now turn to the problem of tuning the penalties $\beta$ and $\beta'$ introduced in Section \ref{sec:Model}. In the previous section, we showed that CAPA is consistent when using an $\alpha\log(n)^{1+\delta}$ penalty for both collective and point anomalies. The result can be tightened slightly to show that consistency is achieved by using an $\alpha\log(n)$ penalty for collective anomalies and an $\alpha'\log(n)$ penalty for point anomalies for sufficiently large constants $\alpha$ and $\alpha'$. In practice, we recommend choosing $\alpha$ and $\alpha'$ with the aim of controlling the asymptotic rate of false positives under the null hypothesis that no anomalies -- collective or point -- are present. 

Relatively tight results can be derived for the case in which the typical mean and variance are known and the observations are normally distributed. Indeed, the following proposition holds under these assumptions:
\begin{Proposition}\label{Prop:FPCONTROL}
	Let $x_1,...,x_n$ be i.i.d.\ $N(\mu,\sigma^2)$ distributed, for known $\mu$ and $\sigma$. Then there exist constants $C_1$ and $C_2$ such that when the penalty for point anomalies, $\beta'$, and the penalty for collective anomalies, $\beta$, satisfy $\beta' \geq 2t$ and $\beta \geq 2(2+2t+2\sqrt{2t})$, then
	\begin{equation*}
	\mathbb{P}\left(\hat{K} = 0 , \; \hat{O} = \emptyset \right) \geq 1 - C_1ne^{-t} - C_2\left(ne^{-t}\right)^2.
	\end{equation*}
\end{Proposition}

Consequently, setting the penalty for collective anomalies to $(4+\epsilon)\log(n)$ and the penalty for point anomalies to $(2+\epsilon)\log(n)$, where $\epsilon > 0$ controls the asymptotic rate of false positives in this scenario, since the probability of observing false positives then tends to 0 as $n$ tends to $\infty$. Moreover, the following proposition holds under general assumptions

\begin{Proposition}\label{Prop:TPCONTROL}
	Let $(\hat{\mu},\hat{\sigma})$, $\hat{O}$, and $\beta'$ correspond to the estimated parameters of the true distribution, the inferred set of point anomalies, and the penalty for point anomalies used by CAPA respectively. Assume, moreover, that the constant $\gamma$, defined in Section \ref{sec:Inference}, satisfies $1 \geq \gamma \geq \exp\left(-\beta'\right)$. Then, there exists a constant $K(\beta') = \beta' + o(\beta')$ such that
	\begin{equation*}
	i \notin \hat{O} \implies (x_i - \hat{\mu})^2 <\hat{\sigma}^2  K(\beta')
	\end{equation*}
	and 
	\begin{equation*}
	(x_i - \hat{\mu})^2 > \hat{\sigma}^2  K(\beta') \implies i \notin \hat{O} \lor \exists k \in \{1,...,\hat{K}\}: i \in \{\hat{s}_k+1,...,\hat{e}_k\}
	\end{equation*}
\end{Proposition}

This proposition defines a threshold for $|x_i-\mu|$ below which the $i$th observation  will not be considered a point anomaly and above which the $i$th observation will not be considered typical. This result provides a natural way of choosing the penalty $\beta'$ for point anomalies based on how large outliers, as measured by $|x_i-\mu|/\sigma$ are assumed to be. For Gaussian data, extreme value theory places the threshold at $\sqrt{(2+\epsilon)\log(n)}$, which confirms the choice of $\beta'$ derived from Proposition \ref{Prop:FPCONTROL}. 

In practice, the typical parameters are often unknown and the constants $\alpha$ and $\alpha'$ should be slightly inflated to reflect this additional uncertainty. This effect on $\alpha$, is offset by the fact that the Bonferroni correction used in the proof of Proposition \ref{Prop:FPCONTROL} becomes loose when the minimum segment length exceeds 2 as it does not exploit correlations. We chose $\beta' = 3\log(n)$ and $\beta = 4\log(n)$ as default penalties for point and collective anomalies respectively in our software implementation and recommend inflating both penalties, whilst maintaining their ratio, when dealing with heavy tailed or autocorrelated data.   

\section{Simulation Study}\label{sec:Simulations}

To assess the potential of CAPA, we compare its performance to that of other popular anomaly and changepoint methods on simulated data. In particular, we compare with PELT as implemented in \cite{killick2014changepoint}, a commonly used changepoint detection method, luminol (\cite{luminol}), an algorithm developed by LinkedIn to detect segments of atypical behaviour, as well as BreakoutDetection (\cite{breakout}) which was introduced by Twitter to detect changes in mean in a way which is robust to point anomalies. 

The simulation study was conducted over simulated time series each consisting of 5000 observations, for which the typical data follows a $N(0,1)$ distribution. Epidemic changepoints start at a rate of 0.0005 (corresponding to an average of about 2.5 epidemic changes in each series), with their length being i.i.d.\ $Pois(30)$ distributed. In each anomalous segment the data is again normally distributed, with the means being i.i.d.\ $N(0,a^2)$ distributed and standard deviations i.i.d.\ $\Gamma(1/b,1/b)$ distributed. We used
\begin{enumerate}
	\item $a = 1$ and $a = 10$ for weak and strong changes in mean respectively
	\item $b = 1$ and $b = 10$ for weak and strong changes in mean respectively
\end{enumerate} 

We compared the performance of the four methods in the presence of both strong and weak changes in mean and/or variance. We also repeated the analysis with 10 i.i.d.\ $N(0,10^2)$ distributed point anomalies occurring at randomly sampled points in the typical data. The comparison of these methods is made using the three different approaches we detail below.

\begin{figure} 
	\begin{subfigure}[b]{0.5\linewidth}
		\centering
		\includegraphics[width=0.9\linewidth]{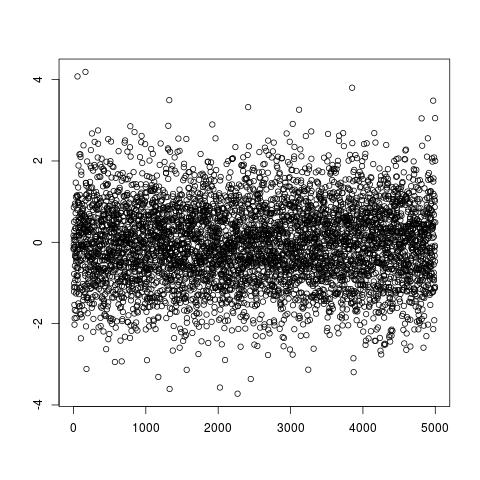} 
		\caption{No point anomalies}
		\label{fig:meanchange_graph} 
		\vspace{4ex}
	\end{subfigure} 
	\begin{subfigure}[b]{0.5\linewidth}
		\centering
		\includegraphics[width=0.9\linewidth]{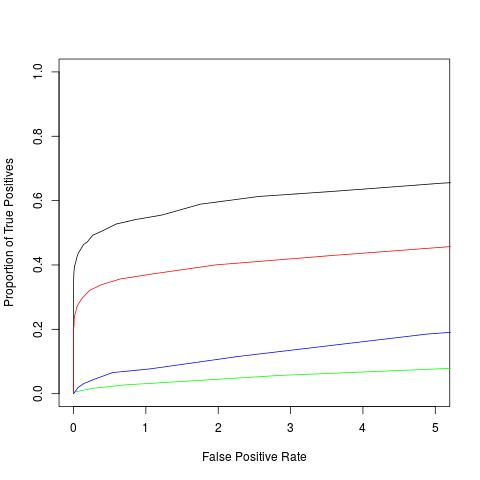} 
		\caption{No point anomalies} 
		\label{fig:meanchange_ROC} 
		\vspace{4ex}
	\end{subfigure} 
	\begin{subfigure}[b]{0.5\linewidth}
		\centering
		\includegraphics[width=0.9\linewidth]{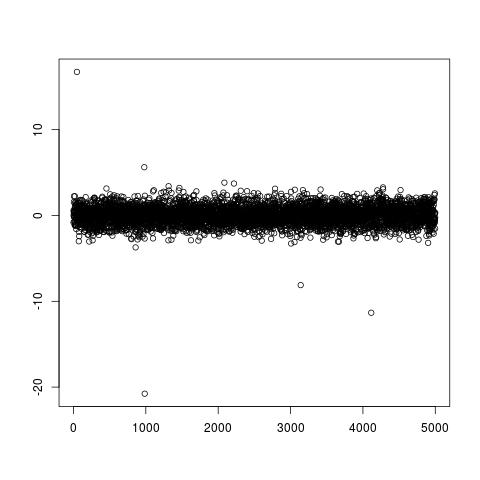} 
		\caption{Point anomalies present} 
		\label{fig:meanANOMchange_graph} 
		\vspace{4ex}
	\end{subfigure} 
	\begin{subfigure}[b]{0.5\linewidth}
		\centering
		\includegraphics[width=0.9\linewidth]{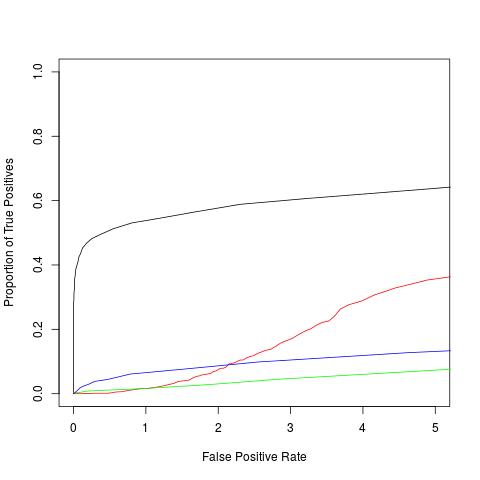} 
		\caption{Point anomalies present} 
		\label{fig:meanANOMchange_ROC} 
		\vspace{4ex}
	\end{subfigure}
	\caption{Data examples and ROC curves for weak changes in mean for CAPA (black), PELT (red), BreakoutDetection (green), and luminol (blue).}
	\label{fig:mean_weak} 
\end{figure}

\begin{figure} 
	\begin{subfigure}[b]{0.5\linewidth}
		\centering
		\includegraphics[width=0.9\linewidth]{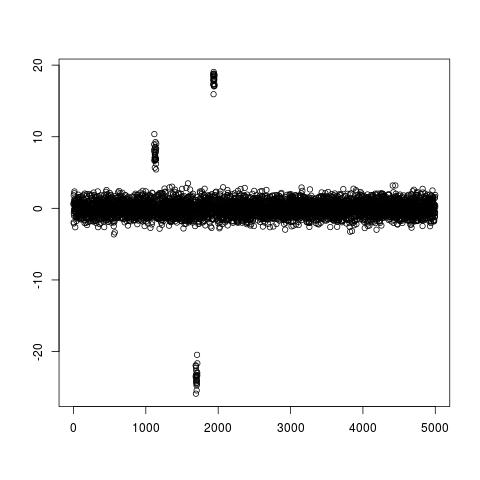} 
		\caption{No point anomalies}
		\label{fig:MEANchange_graph} 
		\vspace{4ex}
	\end{subfigure} 
	\begin{subfigure}[b]{0.5\linewidth}
		\centering
		\includegraphics[width=0.9\linewidth]{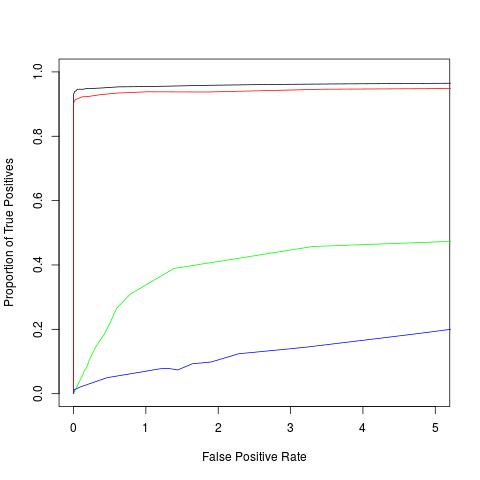} 
		\caption{No point anomalies} 
		\label{fig:MEANchange_ROC} 
		\vspace{4ex}
	\end{subfigure} 
	\begin{subfigure}[b]{0.5\linewidth}
		\centering
		\includegraphics[width=0.9\linewidth]{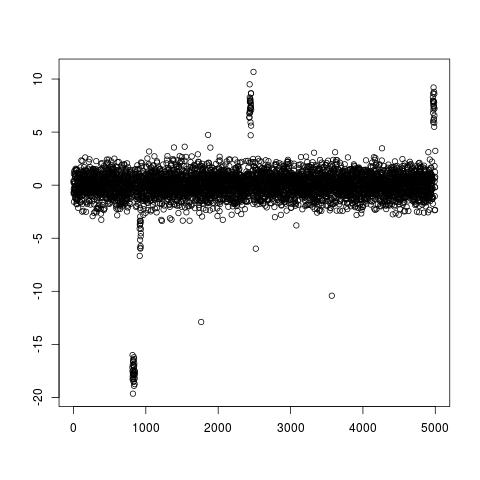} 
		\caption{Point anomalies present} 
		\label{fig:MEANANOMchange_graph} 
		\vspace{4ex}
	\end{subfigure} 
	\begin{subfigure}[b]{0.5\linewidth}
		\centering
		\includegraphics[width=0.9\linewidth]{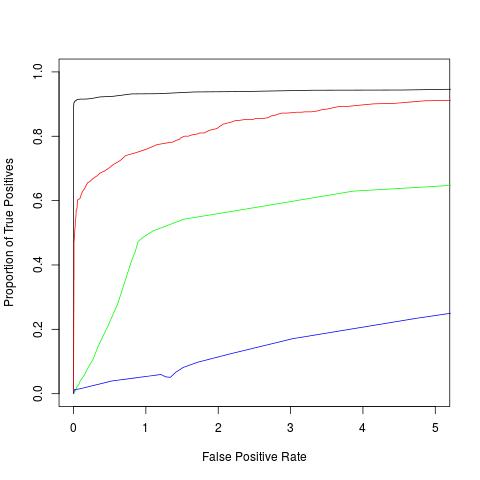} 
		\caption{Point anomalies present} 
		\label{fig:MEANANOMchange_ROC} 
		\vspace{4ex}
	\end{subfigure}
	\caption{Data examples and ROC curves for strong changes in mean for CAPA (black), PELT (red), BreakoutDetection (green), and luminol (blue).}
	\label{fig:MEAN} 
\end{figure}

\subsection{ROC}\label{sec:ROC}

We obtained ROC curves for the four methods. For BreakoutDetection and PELT, we considered detected changes within 20 time points of true changes to be true positives and classified all other detected changes as false positives. For luminol and CAPA, we considered detected starting and end points of epidemic changes to be true positives if they were within 20 observations of a starting and end point respectively. The results regarding the precision of true positives in Section \ref{sec:PREC} suggest that the results in this section are robust with regard to the choice of error tolerance. We set the minimum segment length to ten for PELT, CAPA, and BreakoutDetection. To obtain the ROC curves we varied the penalty for collective anomalies $\beta$ in CAPA, the penalty in PELT, the threshold in luminol, and the beta parameter of BreakoutDetection.

The resulting ROC curves, as well as examples of realisations of the data for the scenario of weak and strong changes in mean can be found in Figures \ref{fig:mean_weak} and \ref{fig:MEAN} respectively. The results for joint changes in mean and variance, as well as changes in variance can be found in the supplementary material. We see that CAPA generally outperforms PELT, even in the absence of point anomalies. This is due to it having more statistical power, by exploiting the epidemic nature of the change. This becomes particularly apparent when the changes are weak. CAPA also outperform BreakoutDetection and luminol for epidemic changes in mean, the scenario for which these methods were developed. Moreover, the performance of CAPA is barely affected by the presence of point anomalies, unlike that of the non-robust methods. This observation remained true when we repeated our analysis with $N(0,1000^2)$ distributed point anomalies. The ROC curves for these additional simulations can be found in the supplementary material.

\begin{figure} 
	\begin{center}
		\begin{tabular}{||c c c ||c | c | c | c ||} 
			\hline
			Mean & Variance & Point anomalies & CAPA & PELT & BreakoutDetection & luminol  \\ [0.5ex] 
			\hline\hline
			weak   & -      & -  & 1.79 & \textbf{1.50} & 3.40 & 9.91 \\ 
			\hline
			weak   & -      & 10  & \textbf{1.72} & 2.27 & 3.75 & 10.70 \\ 
			\hline
			strong & -      & -  & \textbf{0.16} & 0.61 & 5.38 & 15.99\\
			\hline
			strong & -      & 10  & \textbf{0.19} & 0.67 & 4.68 & 15.60 \\ 
			\hline
			-      & weak   & -  & \textbf{1.41} & 1.43 & 4.60 & 9.87 \\
			\hline
			-      & weak   & 10  & \textbf{1.31} & 1.89 & 4.49 &  10.76\\  
			\hline
			-      & strong & -  & \textbf{0.33} & 0.73 & 5.19 & 12.03\\
			\hline
			-      & strong & 10  & \textbf{0.33} & 0.79 & 5.17 & 11.29 \\ 
			\hline
			weak   & weak   & -  & \textbf{1.16} & 1.30 & 4.00 & 11.40\\ 
			\hline
			weak   & weak   & 10  & \textbf{1.22} & 1.63 & 4.00 & 11.30 \\ 
			\hline
			strong & strong & -  & \textbf{0.09} & 0.56 & 3.78 & 16.31\\
			\hline
			strong & strong & 10  & \textbf{0.09} & 0.58 & 3.77 & 15.71 \\ 
			\hline
		\end{tabular}
		\caption{Precision of true positives measured in mean absolute distance for CAPA, PELT, luminol, and BreakoutDetection} 
		\label{tble:Precision}
	\end{center}
\end{figure}

\subsection{Precision}\label{sec:PREC}

We also investigated the precision of the true positives for the four methods. We compared the mean absolute distance between detected changes (i.e.\ true changes which had a detected changes within 20 observations) and the nearest estimated change across all the 12 scenarios. We used the default penalties for all methods (i.e.\ the BIC for PELT) except BreakoutDetection, where we found that the default penalty returned no true positives at all for many scenarios. We therefore used the results we obtained when deriving the ROC curves to set the beta parameter to an appropriate level for each case. 

The results of this analysis can be found in Figure \ref{tble:Precision}. We see that CAPA is generally the most precise one. Moreover, its precision is not too strongly affected by the presence of point anomalies, unlike that of PELT, whose performance is significantly deteriorated by anomalies, especially when the signal is weak. The reason for this is that PELT fits additional changes in the presence of anomalies, which results in shorter segments. This leads to less accurate parameter estimates, which results in poorer estimates for the location of the changepoint. CAPA does not face this problem since the parameter of the typical distribution is shared across all segments. This remains true when the point anomalies are are a lot stronger, as can be seen in the supplementary material.

\begin{figure} 
	\begin{subfigure}[b]{0.5\linewidth}
		\centering
		\includegraphics[width=0.9\linewidth]{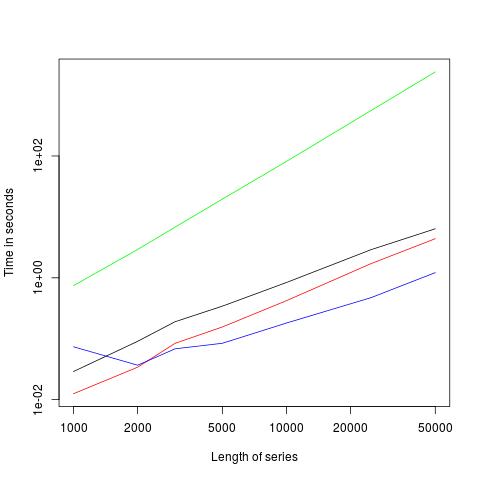} 
		\caption{With epidemic changes} 
		\label{fig:SpeedChange} 
		\vspace{2ex}
	\end{subfigure} 
	\begin{subfigure}[b]{0.5\linewidth}
		\centering
		\includegraphics[width=0.9\linewidth]{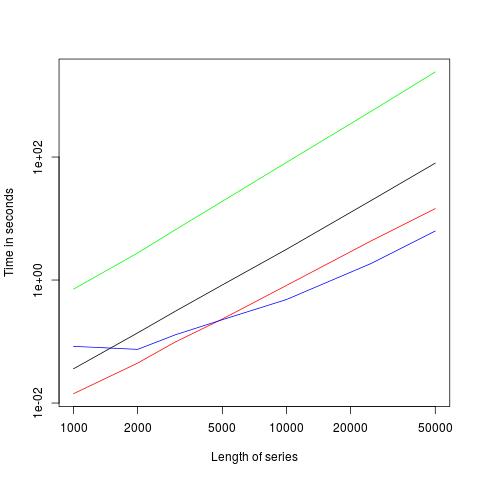} 
		\caption{Stationary data} 
		\label{fig:SpeedFix} 
		\vspace{2ex}
	\end{subfigure}  
	\caption{Runtime of CAPA (black), PELT (red), BreakoutDetection (green), and luminol (blue)}
	\label{fig:Speeds} 
\end{figure}

\subsection{Runtime}

Finally, we investigated the relationship between the runtime of the 4 methods and the number of observations. Our comparison is based on data following a distribution identical to the one we used in Sections \ref{sec:ROC} and \ref{sec:PREC}. Since this type of data favours PELT and CAPA, because the expected number of changes increases with the number of observations, we also compared the runtime of the four methods on stationary $N(0,1)$ data, which represents the worst case scenario for these methods. 

Figure \ref{fig:Speeds} displays the average speed over 50 repetitions for the two cases. When comparing the slopes between 10000 and 50000 datapoints we note that it is very close to 2 for BreakoutDetection in both cases as well as CAPA and PELT for stationary data, suggesting quadratic scaling. In the presence of epidemic changes however, that slope is 1.26 for CAPA -- 1.14 even between 25000 and 50000 datapoints --  thus suggesting near linear runtime.

\section{Application to Kepler Light Curve Data}\label{sec:Application}

\begin{figure}[hb]
	\begin{subfigure}[b]{0.333\linewidth}
		\centering
		\includegraphics[width=\linewidth]{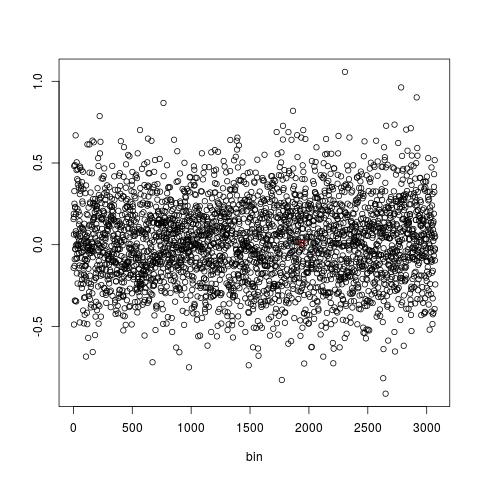} 
		\caption{62.8 days} 
	\end{subfigure}
	\begin{subfigure}[b]{0.333\linewidth}
		\centering
		\includegraphics[width=\linewidth]{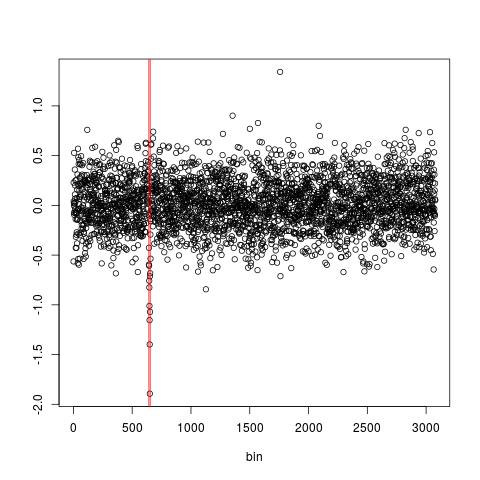} 
		\caption{62.9 days} 
	\end{subfigure}
	\begin{subfigure}[b]{0.333\linewidth}
		\centering
		\includegraphics[width=\linewidth]{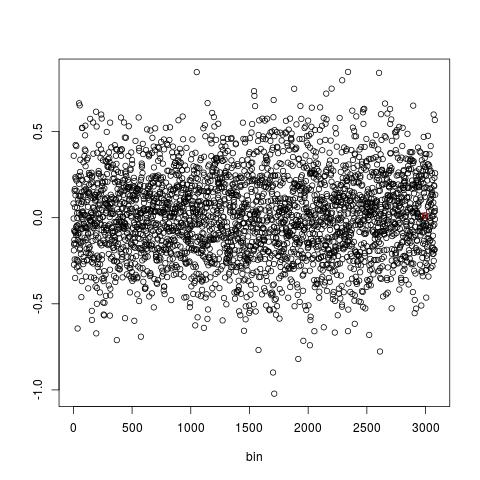} 
		\caption{63.0 days} 
	\end{subfigure}
	\caption{CAPA applied to the light curve of Kepler 1132 preprocessed using different periods.}
	\label{fig:Periodmethod} 
\end{figure}

\begin{figure}
	\centering
	\includegraphics[width=0.8\linewidth]{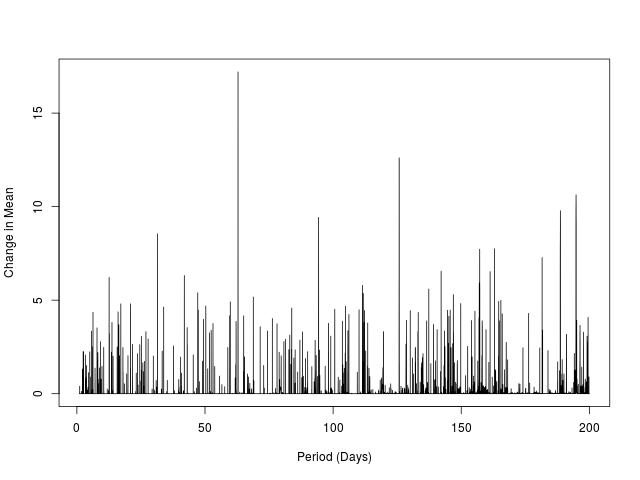} 
	\caption{The strongest change in mean, as measured by $\max_k\left(\triangle_{\mu,k}\right)$, detected by CAPA for the lightcurve of Kepler 1132. All periods from 1 to 200 days at 0.01 day increment were examined}
	\label{fig:Spectrogram} 
\end{figure}

We now apply CAPA to the Kepler light curve data, with the aim of detecting exoplanets via the so called transit method (\cite{Transit}). As described in Section \ref{sec:Intro}, this approach consists of repeatedly measuring a star's brightness for a certain period of time, thus obtaining a so called light curve. Periodically recurring dips in the measurements then point towards the transit of a planet causing a small eclipse. Since the signal of transiting planets is known to be weak, we amplify it by exploiting its periodic nature. If the period of an orbiting planet were known, the signal of its transit could be strengthened by considering all data points to have been gathered at their measurement time modulo that period. We would thus obtain an irregularly sampled time series which we can transform into a regularly sampled time series by binning the data into equally sized bins of length approximately equal to the measurement interval of the Kepler telescope and taking the average within each bin. We could then apply CAPA to this preprocessed data, which would exhibit a stronger signal for any planet with the associated period. Detecting the signal for such a planet involves detecting a collective anomaly with a reduced mean. However we need to do this whilst being robust to the point anomalies in the data, and the potentially other collective anomalies associated with planets with different periods. The results obtained by applying this method, using the default penalties of our software implementation of CAPA, to the light curve of Kepler 1132 using a period of 62.8, 62.9, and 63.0 days can be found in Figure \ref{fig:Periodmethod}. We note that using a period of 62.9 days results in a promising dip, which is not present when using 62.8 or 63 days as period. 

Given a light curve, the periods of exoplanets orbiting the corresponding star (if any are present) are obviously not known a priori. We can, however, apply the above approach for a range of periods, given the fact that the cost of running CAPA is comparable to that of binning the data. Since transits appear as periods of reduced mean, we record the strength of the strongest change in mean as defined by $\max_k \left(\triangle_{\mu,k}\right)$ and estimated using the sample mean and variance in the collective anomalies, and the estimated means and variance of the typical distribution. We expect this quantity to be largest for the periods of exoplanets. We identified the strength of the strongest change in mean for all periods from 1 day to 200 days with increments of 0.01 days for the light curve of Kepler 1132. The result of this analysis can be found in Figure \ref{fig:Spectrogram}. Note that the largest change in mean is recorded at a period of 62.89 days. As with spectral methods, we also observe resonance of the main signal at integer fractions of that period. This result is consistent with the existing literature, which considers Kepler 1132 to be orbited approximately every 62.892 days by the exoplanet Kepler 1132-b whose radius is about 2.5 times that of the earth (\cite{Kepler}). 

We also applied CAPA to the light curves of other stars with confirmed exoplanets and were able to detect their transit signal at the right period. A more detailed exposition of these results can be found in the supplementary material. 

\section{Acknowledgements}
This research has made use of the NASA Exoplanet Archive, which is operated by the California Institute of Technology, under contract with the National Aeronautics and Space Administration under the Exoplanet Exploration Program. The authors would like to thank the Isaac Newton Institute for Mathematical Sciences for support and hospitality during the programme Statistical Scalability when work on this paper was undertaken. This work was supported by EPSRC grant numbers EP/K032208/1 (INI), EP/R014604/1 (INI), EP/N031938/1 (StatScale), and EP/L015692/1 (STOR-i). The authors also acknowledge British Telecommunications plc (BT) for financial support and David Yearling and Kjeld Jensen in BT Research \& Innovation for discussions.

\section{Appendix: Proofs}

This Appendix contains proofs for all the results in this papers. Proofs for Lemmata we use can be found in the supplementary material. 

\subsection{Proof of Proposition \ref{Prop:PRUNING} }

Let $m' \geq m + \hat{l}$. We have 
\begin{align*}
C(m) + \sum_{t=m+1}^{m'}  \mathcal{C}(\textbf{x}_t , \hat{\theta}_{(m+1):m'}) + \beta &\leq C(k) + \sum_{t=k}^{m}  \mathcal{C}(\textbf{x}_t , \hat{\theta}_{(k+1):m}) + \sum_{t=m+1}^{m'}  \mathcal{C}(\textbf{x}_t , \hat{\theta}_{(m+1):m'}) + \beta \\
&\leq C(k) + \sum_{t=k+1}^{m'}  \mathcal{C}(\textbf{x}_t , \hat{\theta}_{(k+1):m'}) + \beta,
\end{align*}
which shows that the cost of choosing $k$ will always be larger than that of choosing $m$. We can thus disregard $k$. 

\subsection{Proof of Theorem \ref{Thm:ConsistencyCP}}

Before proving this theorem, we introduce some notation. We define the cost of a segment $x_{i:j}$ under the true partition $\{0,t_1,...,t_K,n\}$ and true parameters to be  
\begin{equation*}
\mathcal{C}(x_{i:j}) = \sum_{t=i}^{j}\log (\sigma(t)^2) + \sum_{t=i}^{j}\eta_t^2.
\end{equation*}
Note that this cost is additive, i.e.\ for $a<b-1<b+1<c$ we have $\mathcal{C}(x_{a:c}) = \mathcal{C}(x_{a:b}) + \mathcal{C}(x_{(b+1):c})$, whilst the fitted cost satisfies the inequality $\tilde{\mathcal{C}}(x_{a:c}) \geq \tilde{\mathcal{C}}(x_{a:b}) + \tilde{\mathcal{C}}(x_{(b+1):c})$. 

We also define the residual sum of squares $Y_{i:j} = \sum_{k = i}^{j} (\eta_k-\bar{\eta}_{i:j})^2$. Finally, we will work on the event sets $E_1$, $E_2$, $E_3$, $E_4$, $E_5$, and $E_6$ which we define below using notation $a := a(i,j) = j-i+1$
\begin{align*}
E_1 &:= \left\{a\bar{\eta}_{i:j} ^2  < 4(1+\epsilon)\log(n), \, \; \; \, 1 \leq i \leq j \leq n \right\},\\
E_2 &:= \left\{  Y_{i:j} \leq a - 1 + 2 \sqrt{(a-1)(2+\epsilon)\log(n)} + (4+2\epsilon)\log(n), \, \; \; \,  1 \leq i \leq j \leq n \right\}, \\
E_3 &:= \left\{ Y_{i:j} \geq c(a,n)(a-1) , \, \; \; \, 1 \leq i < j \leq n   \right\}, \\
E_4 &:= \left\{ \frac{\sum_{t=t_k-1}^{t_k+1}(x_t-\bar{x}_{(t_k-1):(t_k+1)})^2}{\sigma_k^{2/3}\sigma_{k-1}^{4/3}} > n^{-\epsilon} ,\frac{\sum_{t=t_k}^{t_k+2}(x_t-\bar{x}_{t_k:(t_k+2)})^2}{\sigma_k^{4/3}\sigma_{k-1}^{2/3}} > n^{-\epsilon} ,  \, \; \; \, 1 \leq k \leq K-1 \right\}, \\
E_5 &:= \left\{ \frac{(x_{t_k}-x_{t_k+1})^2}{\sigma_k\sigma_{k-1}} > n^{-\epsilon} ,   \, \; \; \,  1 \leq k \leq K \right\}, \\
E_6 &:= \left\{  Y_{i:j} \geq a - 1 - 2 \sqrt{(a-1)(2+\epsilon)\log(n)}, \, \; \; \, 1 \leq i \leq j \leq n \right\},
\end{align*}
where $c(a,n)<1$ satisfies
\begin{equation*}
\frac{a}{2} \cdot \frac{c(a,n)-1-\log(c(a,n))}{2} = (2+\epsilon)\log(n).
\end{equation*}
Note that $c(a,n)$ is guaranteed to exist by the intermediate value theorem. Indeed, the function $f(x) = x - 1 - \log(x)$ is continuous and satisfies $f(1) = 0$ and $f(x) \rightarrow \infty$ as $x \rightarrow 0^+$.  The motivation for these events is as follows: $E_1$ bounds the error in the estimates of the mean, while $E_2$, $E_3$, and $E_6$ bound the error in the estimates of the variance. $E_5$ and $E_4$ are needed to prevent the existence of segments length two and three respectively in which the observations lie to close to each other, which would encourage the algorithm to erroneously fit them in a short segment of low variance. We also define $E_7$ which guarantees that the signal strength of true changes is utilised: 
\begin{equation*}
E_7 = \left\{\sum_{t_k-D+1}^{t_k+D}(x_t-\bar{x}_{(t_k-D+1):(t_k+D)})^2 \geq 2D\sigma_{k}\sigma_{k+1}c(D,n)\exp\left(\triangle_k\right), \; \; \; 1 \leq D \leq \min(t_{k+1} - t_k,t_{k} - t_{k-1}), \; \; \; 1 \leq k \leq K\right\},
\end{equation*}
where $c(D,n)<1$ satisfies $D \left(c(D,n) - 1 - \log(c(D,n))  \right) = 2(2+\epsilon)\log(n)$. We write $E = \cap E_i$ and now in a position to prove the following lemmata:

\begin{Lemma}\label{lemma:Yao}
	(Yao 1988) $\mathbb{P}(E_1) > 1 - \tilde{K_1}n^{-\epsilon}$, for some constant $\tilde{K_1}$.
\end{Lemma}

\begin{Lemma}\label{lemma:ProbabilityofboundsIgotoftheinternet}
	$\mathbb{P}(E_2) > 1 - \tilde{K_2}n^{-\epsilon}$, $\mathbb{P}(E_3) > 1 - \tilde{K_3}n^{-\epsilon}$, $\mathbb{P}(E_4) > 1 - \tilde{K_4}n^{-\epsilon}$, $\mathbb{P}(E_5) > 1 - \tilde{K_5}n^{-\epsilon}$, $\mathbb{P}(E_6) > 1 - \tilde{K_6}n^{-\epsilon}$, and $\mathbb{P}(E_7) > 1 - \tilde{K_7}n^{-\epsilon}$for some constants $\tilde{K_2}$,$\tilde{K_3}$,$\tilde{K_4}$, $\tilde{K_5}$, $\tilde{K_6}$, and $\tilde{K_7}$.
\end{Lemma}

\begin{Lemma}\label{lemma:Yijbounds}
	There exists a constant $\tilde{C}_1$ such that $Y_{i:j} - a - a\log(Y_{i:j} /a)  \leq \tilde{C}_1 \log(n)$  holds on $E$ for all $1 \leq i < j \leq n$.
\end{Lemma}

\begin{Lemma}\label{lemma:Falsepositiveremoval}
	Let $i,j$ be such that there exists some $k$ such that $t_{k-1} < i < j \leq t_{k}$. The following holds given $E$ :
	\begin{equation*}
	0 \leq \mathcal{C}\left(x_{i:j}\right) - \tilde{\mathcal{C}}\left(x_{i:j}\right) \leq \tilde{C}_2 \log(n).
	\end{equation*}  
\end{Lemma}

\begin{Lemma}\label{lemma:Falsepositiveremovalforspecialcases}
	Let $i,j$ be such that $\exists k$ such that $t_{k-1} = i < j \leq t_{k}$ or $t_{k-1} < i < j = t_{k}+1$. The following then holds given $E$
	\begin{equation*}
	\mathcal{C}\left(x_{i:j}\right) - \tilde{\mathcal{C}}\left(x_{i:j}\right) \leq \tilde{C}_3 \log(n)
	\end{equation*}  
\end{Lemma}

\begin{Lemma}\label{lemma:Combiningstuff}
	Let $a,b,c \in \tau$ for some partition $\tau$ of $x_{i,j}$ such that $\exists k$ such that $t_{k-1} < a < b < c \leq t_{k}$. Then, 
	\begin{equation*}
	\tilde{\mathcal{C}}\left(x_{i:j},\tau,\alpha\right) - \tilde{\mathcal{C}}\left(x_{i:j},\tau_{-b},\alpha \right) \geq \frac{3}{4}\alpha \log(n)^{1+\delta},
	\end{equation*}  
	where $\tau_{-b} = \tau \setminus \{b\} $ holds on $E$ for large enough $n$.
\end{Lemma}

\begin{Lemma}\label{lemma:Mixlemma}
	For all $\alpha > 0$, there exists a constant $\tilde{\kappa}(\alpha,\epsilon)$  such that $\tilde{\mathcal{C}}\left( x_{i:j} \right) -( \mathcal{C}\left( x_{i:t_{k}}\right) + \mathcal{C}\left( x_{(t_{k}+1):j}\right) )\geq \alpha \log(n)^{1+\delta} $ holds on $E$ if 
	\begin{equation*}
	j - t_{k} = t_{k} + 1 - i \geq \frac{\tilde{\kappa}(\alpha,\epsilon)}{ \min(\triangle_{k},\triangle_{k}^2)}\log(n)^{1+\delta}
	\end{equation*} 
	and $j \leq t_{k+1}, i >t_{k-1}$ for all $n>2$. 
\end{Lemma}

We now define
\begin{equation*}
\tilde{\kappa}_k = 2\frac{\tilde{\kappa}(3\alpha,\epsilon)}{\min(\triangle_{k},\triangle_{k}^2)}
\end{equation*}
and the set of partitions
\begin{equation*}
\mathcal{B}:= \left\{\{0,t'_1,t'_2,...,t'_K,n\} \; \; \; |  \; \; \; |t'_k - t_k| \leq \tilde{\kappa}_k\log(n)^{1+\delta}  \; \; \; 1 \leq k \leq K \right\},
\end{equation*}
which are within $\tilde{\kappa}_k\log(n)^{1+\delta}$ of the true partition. 

We will show that, for large enough $n$, the optimal partition lies in $\mathcal{B}$ given the event set $E$. Given the probability of $E$, this proves Theorem \ref{Thm:ConsistencyCP}.  Our approach will consist of showing that the cost of a partition $\tau \notin \mathcal{B}$ is higher than that of the true partition with the true parameters (see Proposition \ref{prop:Mainproposition}). We will achieve this by adding free changes to $\tau$ thus splitting up the series into multiple sub-segments each containing a single true changepoint and $\tilde{\kappa}_k\log(n)^{1+\delta}$ points either side of it. This also defines a projection of $\tau$ onto the partitions of the sub-segments. We define the set of partitions
\begin{equation*}
\mathcal{B}_k:= \left\{\{i-1,t'_k,j\} \; \; \; |  \; \; \; |t'_k - t_k| \leq \tilde{\kappa}_k\log(n)^{1+\delta}  \right\}
\end{equation*}
for segments $x_{i:j}$ for which there exist a $k$ such that: $t_{k-1} +1 \leq i  \leq t_k - \tilde{\kappa}_k\log(n)^{1+\delta} <  t_k + \tilde{\kappa}_k\log(n)^{1+\delta} \leq j \leq t_{k+1}$ as an analogue of $\mathcal{B}$ for the whole of $x$.

If $\tau \notin \mathcal{B}$, there must be at least one sub-segment for which the projection of $\tau$ does not lie in $\mathcal{B}_k$. We will show in Proposition \ref{Prop:Notinmakeitworsebylogn(1+delta)}, that the cost of the true partition using the true parameters is at least $O(\log(n)^{1+\delta})$ lower than that of the projection of $\tau$ on such a segment. We will also show in Proposition \ref{Prop:Improvementboundedbylogn} that the projections of $\tau$ which are in $\mathcal{B}_{k}$ have a cost which is at most $O(\log(n))$ lower than that of the true partition with true parameters.  

\begin{Proposition}\label{Prop:Improvementboundedbylogn}
	Let $i,j \in N$, be such that there exists a $k$ such that: $t_{k-1} +1 \leq i  <  t_k  < j \leq t_{k+1}$, then there exists a constant $\tilde{C}_4$ such that given $E$, 
	\begin{equation*}
	\left[  \mathcal{C}\left( x_{i:j}\right) + \alpha\log(n)^{1+\delta} \right] - \tilde{\mathcal{C}}(x_{i:j}, \tau, \alpha)  \leq  \tilde{C}_4\log(n)
	\end{equation*}
	for all valid partitions $\tau$ of the form  $\tau = \{i-1, \hat{t}, j \}$, if $n$ is large enough. 
\end{Proposition}

\textbf{Proof of Proposition \ref{Prop:Improvementboundedbylogn}}: The following cases are possible: 

\textbf{Case 1:} $\hat{t} = t_k$. Then: 
\begin{align*}
\left[  \mathcal{C}\left( x_{i:j}\right) + \alpha\log(n)^{1+\delta} \right] - \tilde{\mathcal{C}}(x_{i:j}, \{i-1,\hat{t},j\},\alpha) = \mathcal{C}\left( x_{i:j}\right) - \left[ \tilde{\mathcal{C}}(x_{i:t_k}) + \tilde{\mathcal{C}}(x_{(t_k+1):j})  \right] \leq 2\tilde{C}_2\log(n),
\end{align*}
where the inequality follows from Lemma \ref{lemma:Falsepositiveremoval}. 

\textbf{Case 2:} $\hat{t} = t_k+1$. Then:
\begin{align*}
&\left[  \mathcal{C}\left( x_{i:j}\right) + \alpha\log(n)^{1+\delta} \right] - \tilde{\mathcal{C}}(x_{i:j}, \{i-1,\hat{t},j\},\alpha )  \\ & =  \mathcal{C}\left( x_{i:j}\right) - \left[ \tilde{\mathcal{C}}(x_{i:(t_k+1)}) + \tilde{\mathcal{C}}(x_{(t_k+2):j})  \right]\leq  (\tilde{C}_2+\tilde{C}_3)\log(n),
\end{align*}
where the inequality follows from Lemmata \ref{lemma:Falsepositiveremoval} and \ref{lemma:Falsepositiveremovalforspecialcases}. 

\textbf{Case 3:} $\hat{t} > t_k+1$. Then:
\begin{align*}
&\left[  \mathcal{C}\left( x_{i:j}\right) + \alpha \log(n)^{1+\delta} \right] - \tilde{\mathcal{C}}(x_{i:j}, \{i-1,\hat{t},j\}) \leq  \mathcal{C}\left( x_{i:j}\right) + 2\alpha\log(n)^{1+\delta}  - \tilde{\mathcal{C}}(x_{i:j}, \{i-1,t_k,\hat{t},j\},\alpha) \\
&= 
\mathcal{C}\left( x_{i:j}\right) - \left[ \tilde{\mathcal{C}}(x_{i:t_k}) + \tilde{\mathcal{C}}(x_{(t_k+1):\hat{t}}) + \tilde{\mathcal{C}}(x_{(\hat{t}+1):j})  \right] \leq 3\tilde{C}_2\log(n),
\end{align*}
where the first inequality follows from the fact that introducing an unpenalised changepoint reduces cost and the second is a consequence of Lemma \ref{lemma:Falsepositiveremoval}. 

\textbf{Case 4:} $\hat{t} = t_k-1$. Symmetrical to case 2.

\textbf{Case 5:} $\hat{t} < t_k-1$. Symmetrical to case 3.

This finishes our proof.

\begin{Proposition}\label{Prop:Notinmakeitworsebylogn(1+delta)}
	There exists a constant $n_4(\alpha,\delta,\epsilon)$, such that $\forall i,j$ for which $\exists k$ such that $t_{k-1} +1 \leq i  \leq t_k - \tilde{\kappa}_k\log(n)^{1+\delta} <  t_k + \tilde{\kappa}_k\log(n)^{1+\delta} \leq j \leq t_{k+1}$
	\begin{equation*}
	\tilde{\mathcal{C}}(x_{i:j}, \tau,\alpha) - \left[  \mathcal{C}\left( x_{i:j}\right) + \alpha\log(n)^{1+\delta} \right] \geq  \frac{1}{3}\alpha\log(n)^{1+\delta}
	\end{equation*}
	holds for all $\tau \notin \mathcal{B}_k$ given $E$ and $n>n_4(\alpha,\delta,\epsilon)$.
\end{Proposition}

\textbf{Proof of Proposition \ref{Prop:Notinmakeitworsebylogn(1+delta)}:} Consider $\tau' \notin \mathcal{B}_k$. We consider the following three cases and denote $H:= \lceil \frac{1}{2}\tilde{\kappa}_k\log(n)^{1+\delta} \rceil$, noting that it is larger than
\begin{equation*}
\frac{\tilde{\kappa}(3\alpha,\epsilon)}{\min(\triangle_{k},\triangle_{k}^2)}\log(n)^{1+\delta}
\end{equation*} 
\textbf{Case 1}: $|\tau'| = 2$. We have $\tau' = \{i-1,j\}$. Hence: 
\begin{align*}
&\tilde{\mathcal{C}}(x_{i:j}, \tau',\alpha)  \geq \tilde{\mathcal{C}}(x_{i:(t_k - H)}) +  \tilde{\mathcal{C}}(x_{(t_k - H + 1):(t_k + H)})  + \tilde{\mathcal{C}}(x_{(t_k + H + 1):j})  \\
\geq &\tilde{\mathcal{C}}(x_{i:(t_k - H)}) +  \mathcal{C}(x_{(t_k - H + 1):(t_k + H)}) + 3\alpha\log(n)^{1+\delta}  + \tilde{\mathcal{C}}(x_{(t_k + H + 1):j}) \\ \geq & 2\alpha\log(n)^{1+\delta} - 2\tilde{C}_2 \log(n) + \left[  \mathcal{C}\left( x_{i:j}\right) + \alpha\log(n)^{1+\delta} \right],
\end{align*}
where the second inequality follows from the definition of $H$ and Lemma \ref{lemma:Mixlemma} and the third from Lemma \ref{lemma:Falsepositiveremoval}.

\textbf{Case 2}: $|\tau'| = 3$. We have $\tau' = \{i-1,t_k+L,j\}$, where $|L| > \tilde{\kappa}_k\log(n)^{1+\delta}$. We assume $L>0$, the other case being very similar. We have: 
\begin{align*}
&\tilde{\mathcal{C}}(x_{i:j}, \{i-1,t_k+L,j\},\alpha) = \tilde{\mathcal{C}}(x_{i:(t_k+L)}) +  \tilde{\mathcal{C}}(x_{(t_k+L+1):j}) +  \alpha\log(n)^{1+\delta}\\
\geq &\tilde{\mathcal{C}}(x_{i:(t_k-H-1)}) + \tilde{\mathcal{C}}(x_{(t_k-H):(t_k+H)}) + \tilde{\mathcal{C}}(x_{(t_k+H+1):(t_k+L)})  - \tilde{C}_2 \log(n) + \mathcal{C}(x_{(t_k+L+1):j})  + \alpha \log(n)^{1+\delta}\\
\geq & 3\alpha\log(n)^{1+\delta} - 3\tilde{C}_2 \log(n) + \left[  \mathcal{C}\left( x_{i:j}\right) + \alpha\log(n)^{1+\delta} \right],
\end{align*}
where the inequalities follow from of the definition of $H$ as well as Lemmata \ref{lemma:Mixlemma} and \ref{lemma:Falsepositiveremoval}.

\textbf{Case 3}: $|\tau'| \geq 4$. Let $\tau' = \{a_1,a_2,...,a_{|\tau'|} \}$, where $a_1 = i - 1$ and $a_{|\tau'|} = j$. There must exist a $l \in \{2,...,|\tau'|-1\}$, such that $a_{l-1} < t_k$ and $a_{l+1} > t_k +1$. We thus have: 
\begin{align*}
&\tilde{\mathcal{C}}(x_{i:j}, \tau',\alpha) = (|\tau'|-3)\alpha\log(n)^{1+\delta}
+ \left(\sum_{m=1}^{l-2}  + \sum_{m=l+1}^{|\tau'| - 1} \right)\left[ \tilde{\mathcal{C}}(x_{a_{m}+1,a_{m+1}}) \right] + \tilde{\mathcal{C}}(x_{(a_{l-1}+1):a_{l+1}}, \{  a_{l-1},a_{l},a_{l+1} \},\alpha) \\
&\geq (|\tau'|-2)\alpha\log(n)^{1+\delta} + \left(\sum_{m=1}^{l-2}  + \sum_{m=l+1}^{|\tau'| - 1} \right)\left[ \mathcal{C}(x_{a_{m}+1,a_{m+1}}) \right] +  \mathcal{C}(x_{(a_{l-1}+1):a_{l+1}} ) - (|\tau'|-3)\tilde{C}_2\log(n)  - \tilde{C}_4\log(n) \\
& = \mathcal{C}(x_{i:j}, \tau,\alpha) + \alpha\log(n)^{1+\delta} + (|\tau'|-3)\alpha\log(n)^{1+\delta} - \left[(|\tau'|-3)\tilde{C}_2 + \tilde{C}_4\right]\log(n),
\end{align*}
by Lemma \ref{lemma:Falsepositiveremoval} and Proposition \ref{Prop:Improvementboundedbylogn}. This finishes the proof. 

\begin{Proposition}\label{prop:Mainproposition}
	There exists a constant $\tilde{n}_5(\alpha,\delta,\tilde{\delta},\epsilon)$ such that given $E$, we have 
	\begin{equation*}
	\tilde{\mathcal{C}} (x_{1:n},\tau,\alpha) - \left[\mathcal{C}(x_{1,n}) +K\alpha\log(n)^{1+\delta} \right] \geq \frac{1}{4}\alpha \log(n)^{1+\delta}
	\end{equation*}
	for all $\tau \notin \mathcal{B}$ if $n\geq \tilde{n}_5(\alpha,\delta,\tilde{\delta},\epsilon)$.
\end{Proposition}

\textbf{Proof of Proposition \ref{prop:Mainproposition}:} 
First, consider the special case $K=0$. For this case, $\tau \notin \mathcal{B}$ implies that $\hat{K}\geq1$. We have
\begin{equation*}
\tilde{\mathcal{C}} (x_{1:n},\tau,\alpha) \geq \tilde{\mathcal{C}} (x_{1:n},\{0,n\},\alpha )+ \hat{K}\frac{3}{4}\alpha\log(n)^{1+\delta} \geq \mathcal{C}(x_{1:n}) + \hat{K}\frac{3}{4}\alpha\log(n)^{1+\delta} - \tilde{C_2}\log(n),
\end{equation*}
where the first inequality follows from Lemma \ref{lemma:Combiningstuff} and the second from Lemma \ref{lemma:Falsepositiveremoval}.

Next assume $K \geq 1$. Let $\tau \notin \mathcal{B}$. We now introduce free changepoints $l_0,l_1,...,l_K$ to break up the series into multiple sub-series with one true changepoint each. We impose $l_0 = 0$, $l_K = n$, $|l_k-t_k| > 4\tilde{\kappa}_k\log(n)^{1+\delta}$ for $0<k\leq K$ and $|l_k-t_{k+1}| > 4\tilde{\kappa}_k\log(n)^{1+\delta}$ for $0\leq k < K$. We also require that $\tau \cup \{l_0,...,l_K\}$ is a valid partition (i.e.\ one which has segments of length at least two) and that there exists a $\hat{k}$ such that $\tau_{\hat{k}} := \tau \cap \{l_{\hat{k}-1}+1,l_{\hat{k}-1}+2,...,l_{\hat{k}}\} \notin \mathcal{B}_{\hat{k}}$. We are guaranteed to find such points $l_0,l_1,...,l_K$ if $n$ is such that
\begin{equation*}
\frac{1}{\min(\triangle_{k},\triangle_{k}^2)}\log(n)^{1+\delta+\tilde{\delta}} \geq 12 \tilde{\kappa}_k \log(n)^{1+\delta},
\end{equation*}
which is satisfied if $n > \tilde{n}_5(\alpha,\tilde{\delta},\epsilon)$. Indeed, we can choose points near the middle of the true segments which are not in $\tau$, or by select points in $\tau$ if the former is impossible because there are too many point in $\tau$ near the middle of some segment. 

Since introducing free changes reduces the cost we then have  
\begin{align*}
\tilde{\mathcal{C}} (x_{1:n},\tau,\alpha) &\geq \sum_{k=1}^{K} \tilde{\mathcal{C}} (x_{(l_{k-1}+1):l_k},\tau_k,\alpha) = \tilde{\mathcal{C}} (x_{(l_{\hat{k}-1}+1):l_{\hat{k}}},\tau_{\hat{k}},\alpha) + \sum_{k \neq \hat{k}} \tilde{\mathcal{C}} (x_{(l_{k-1}+1):l_k},\tau_k,\alpha) \\
&\geq \mathcal{C} (x_{1:n},\tau,\alpha) + \frac{1}{3}\alpha\log(n)^{1+\delta} - (K-1)\tilde{C}_4\log(n),
\end{align*}
where the second inequality follows from Propositions \ref{Prop:Improvementboundedbylogn} and \ref{Prop:Notinmakeitworsebylogn(1+delta)}. This finishes the proof.

\textbf{Proof of Theorem \ref{Thm:ConsistencyCP}}: $\mathcal{B}$ contains the true partition with fitted parameters which is cheaper than the true partition with true parameters. Proposition \ref{prop:Mainproposition} shows that conditional on $E$ the true partition with true parameters will be cheaper than all $\tau \notin \mathcal{B}$ fo $n > \tilde{n}_5(\alpha,\delta,\tilde{\delta},\epsilon)$. The optimal partition must therefore be in $\mathcal{B}$, given event set $E$. This proves Theorem \ref{Thm:ConsistencyCP}, since Lemmata \ref{lemma:Yao} and \ref{lemma:ProbabilityofboundsIgotoftheinternet} imply that $\mathbb{P}( E) \geq 1 - (\tilde{K}_1 + \tilde{K}_2 + \tilde{K}_3 + \tilde{K}_4 + \tilde{K}_5 + \tilde{K}_6 + \tilde{K}_7)n^{-\epsilon}$.

\subsection{Proof of Theorem \ref{Thm:Consistency}}

In order to prove this result, we will use the following notation in this section: We define $\tilde{\mathcal{C}}_E \left( x_{1:n} , \tau_E , \alpha , \mu , \sigma \right)$ to be the cost of an epidemic partition $\tau_E = \{\hat{s}_1 ,\hat{e}_1, ... \hat{s}_{\hat{K}} ,\hat{e}_{\hat{K}}\}$ under a penalty $\alpha\log(n)^{1+\delta}$ and inferred parameters of the typical distribution $\mu , \sigma$. We define $\mathcal{C}_E \left( x_{1:n} , \alpha , \mu , \sigma \right)$, to be the cost under the true partition using the true parameters for the epidemic segments and $\mu , \sigma$ as estimates for the parameters of the typical distribution. We also define the set of epidemic partitions
\begin{equation*}
\mathcal{B}_E = \left\{\{\hat{s}_1 ,\hat{e}_1, ... ,\hat{s}_K ,\hat{e}_K\} \; \; | \; \; |\hat{e}_k - e_k| < \tilde{\kappa}_k \log(n)^{1+\delta} , |\hat{s}_k - s_k| < \tilde{\kappa}_k \log(n)^{1+\delta} \;\;\;\;\;\; 1 \leq k \leq K \right\}
\end{equation*}
as an epidemic equivalent of $\mathcal{B}$. Finally, we note that we can extend the definition of the event set $E$ to epidemic changepoints by treating the $s_k$ and $e_k$ like classical changepoints.

We will begin by proving a simplified version of the theorem in which we run our epidemic changepoint detection algorithm without allowing for epidemic changes of length one in variance only and imposing that each segment of the data allocated to the typical distribution is of length at least two. The reason for this is that this allows us to define an equivalent non-epidemic partition, whose segments must be of length at least two, for each epidemic partition. We also begin by assuming that the parameter of the typical distribution is known. 

This simplified version captures the main ideas of the full proof. We will proceed to showing that the result also holds when the typical mean and variance are unknown. This will be followed by  a proof of the full result by means of introducing and proving the consistency of a modified version of the classical changepoint detection algorithm described in the previous section which also allows for segments of length one. 

For now, we assume that all segments are of length at least two and that the true parameters $\mu_0$ and $\sigma_0$ are known. This allows us to use the fact that the cost of the true epidemic partition using the true parameters is exactly the same as the cost of the corresponding true non-epidemic partition using the true parameters with twice the penalty. We can therefore prove the following proposition, as a corollary of Proposition \ref{prop:Mainproposition}.
\begin{Proposition}\label{prop:MainPARAMKNOWNLENGTHATLEAST2}
	There exists a constant $\tilde{n}_6(\alpha,\delta,\tilde{\delta},\epsilon)$, such that for all $\tau_E' \notin \mathcal{B}_E$
	\begin{equation*}
	\tilde{\mathcal{C}}_E \left( x_{1:n} , \tau_E' , \alpha , \mu_0 , \sigma_0 \right) - \mathcal{C}_E \left( x_{1:n} , \alpha , \mu_0 , \sigma_0 \right) \geq \frac{1}{5}\alpha\log(n)^{1+\delta/2}
	\end{equation*}	
	holds on $E$ for  $n > \tilde{n}_6(\alpha,\delta,\tilde{\delta},\epsilon)$.
\end{Proposition}

\textbf{Proof of Proposition \ref{prop:MainPARAMKNOWNLENGTHATLEAST2}:} We note that 
\begin{equation*}
\tilde{\mathcal{C}}_E \left( x_{1:n} , \tau_E' , \alpha , \mu_0 , \sigma_0 \right) \geq 
\tilde{\mathcal{C}}\left( x_{1:n} , \tau_E' \cup \{0,n\} , \frac{1}{2}\alpha \right) + \frac{\alpha}{2} \sum_{k=2}^{\hat{K}} \mathbb{I}\{s_k = e_{k-1}\} \log(n)^{1+\delta},
\end{equation*}
because using fitted parameters instead of $\mu_0$ and $\sigma_0$ for segments allocated to the typical distribution under $\tau_E'$ can only reduce the cost. Additionally, two epidemic changes correspond to three classical changepoints if their end and starting points coincide. Moreover, 
\begin{equation*}
\mathcal{C}_E \left( x_{1:n} , \alpha , \mu_0 , \sigma_0\right) = \mathcal{C}\left( x_{1:n}\right) + K\alpha\log(n)^{1+\delta}. 
\end{equation*}
Therefore: 
\begin{align*}
&\tilde{\mathcal{C}}_E \left( x_{1:n} , \tau_E' , \alpha , \mu_0 , \sigma_0 \right) - \mathcal{C}_E \left( x_{1:n} , \alpha , \mu_0 , \sigma_0 \right) \\ &\geq 
\tilde{\mathcal{C}}\left( x_{1:n} , \tau_E' \cup \{0,n\} , \frac{1}{2}\alpha \right) + \frac{\alpha}{2} \sum_{k=2}^{\hat{K}} \mathbb{I}\{s_k = e_{k-1}\} \log(n)^{1+\delta} - \left[\mathcal{C}\left( x_{1:n}\right) + 2K\frac{\alpha}{2}\log(n)\right]. 
\end{align*}
This leaves two possibilities. If $\tau_E' \cup \{0,n\} \notin \mathcal{B}$ then the above will exceed 
\begin{equation*}
\frac{1}{4}\alpha \log(n)^{1+\delta},
\end{equation*}
by proposition \ref{prop:Mainproposition}. Since $\tau_E' \notin \mathcal{B}_E$, the only way we can have $\tau_E' \cup \{0,n\} \in \mathcal{B}$ is if there exists a $k$ such that $s_k = e_{k-1}$. In that case the difference will exceed
\begin{equation*}
\frac{1}{2}\alpha\log(n)^{1+\delta} - (2K+1)\tilde{C_4}\log(n),
\end{equation*}
by Proposition \ref{Prop:Improvementboundedbylogn}. This finishes the proof. 

We can now use this proposition to prove Theorem \ref{Thm:Consistency} in the same way we used \ref{prop:Mainproposition} to prove Theorem \ref{Thm:ConsistencyCP}.

\textbf{Proof of Theorem \ref{Thm:Consistency}:} Proposition \ref{prop:MainPARAMKNOWNLENGTHATLEAST2} proves Theorem \ref{Thm:Consistency} as Lemmata \ref{lemma:Yao} and \ref{lemma:ProbabilityofboundsIgotoftheinternet} imply that $\mathbb{P}( E) \geq 1 - (\tilde{K}_1 + \tilde{K}_2 + \tilde{K}_3 + \tilde{K}_4 + \tilde{K}_5 + \tilde{K}_6 + \tilde{K}_7)n^{-\epsilon}$.
\\
\\
We now introduce the following lemma about the distribution of the median and inter-quantile range. It will allow us to prove Theorem \ref{Thm:Consistency} when the true parameters are unknown.

\begin{Lemma}\label{lemma:Robuststats}
	There exists a constants $\tilde{K}_8$, $D_1$, and $D_2$ such that for large enough $n$
	\begin{equation*}
	\mathbb{P}\left(|\hat{\mu}-\mu_0| \leq D_1\sigma_0\sqrt{\frac{\log(n)}{n}}, \left|\frac{\hat{\sigma}^2}{\sigma_0^2}-1\right| \leq D_2\sqrt{\frac{\log(n)}{n}}\right)
	\geq 1 - \tilde{K}_8n^{-\epsilon}
	\end{equation*}
\end{Lemma}

We can use this Lemma above to introduce a new event $E_8$ stating that the estimated parameters are close to the true parameters.
\begin{equation*}
E_8 := \left\{|\hat{\mu}-\mu_0| \leq D_1\sigma_0\sqrt{\frac{\log(n)}{n}}, \left|\frac{\hat{\sigma}^2}{\sigma_0^2}-1 \right| \leq D_2\sqrt{\frac{\log(n)}{n}}\right\}.
\end{equation*}

This event bounds the effect of using the estimated typical parameters instead of the true parameters for the cost of the true distribution with true non-typical parameters. Indeed, the following lemma holds:

\begin{Lemma}\label{lemma:CostoftruevsCostofFalse}
	There exists a constant $\tilde{C}_7$ such that given $E$ and $E_8$ and n large enough we have:
	\begin{equation*}
	\tilde{\mathcal{C}}_E \left( x_{1:n} , \alpha , \hat{\mu} , \hat{\sigma} \right) - \mathcal{C}_E \left( x_{1:n} , \alpha , \mu_0 , \sigma_0 \right) \leq \tilde{C}_7\log(n).
	\end{equation*}	
\end{Lemma}

We can use this lemma to prove the following extension of Proposition \ref{prop:MainPARAMKNOWNLENGTHATLEAST2} to the case when the typical parameters are inferred.
\begin{Proposition}\label{prop:MainLENGTHATLEAST2}
	There exists a constant $\tilde{n}_7(\alpha,\delta,\tilde{\delta},\epsilon)$ such that for all $\tau_E' \notin \mathcal{B}_E$
	\begin{equation*}
	\tilde{\mathcal{C}}_E \left( x_{1:n} , \tau_E' , \alpha , \hat{\mu} , \hat{\sigma} \right) - \mathcal{C}_E \left( x_{1:n} , \alpha , \hat{\mu} , \hat{\sigma} \right) \geq \frac{1}{5}\alpha\log(n)^{1+\delta/2}
	\end{equation*}	
	holds on $E \cap E_8$ for $n > \tilde{n}_7(\alpha,\delta,\tilde{\delta},\epsilon)$.
\end{Proposition}

\textbf{Proof of Proposition \ref{prop:MainLENGTHATLEAST2}:} We note that, as before, 
\begin{equation*}
\tilde{\mathcal{C}}_E \left( x_{1:n} , \tau_E' , \alpha , \hat{\mu} , \hat{\sigma} \right) \geq 
\tilde{\mathcal{C}}\left( x_{1:n} , \tau_E' \cup \{0,n\} , \frac{1}{2}\alpha \right) + \frac{\alpha}{2} \sum_{k=2}^{\hat{K}} \mathbb{I}\{s_k = e_{k-1}\} \log(n)^{1+\delta}
\end{equation*}
\begin{equation*}
\mathcal{C}_E \left( x_{1:n} , \alpha , \mu , \sigma \right) = \mathcal{C}\left( x_{1:n}\right) + K\alpha\log(n)^{1+\delta}, 
\end{equation*}
Therefore we now have
\begin{align*}
&\tilde{\mathcal{C}}_E \left( x_{1:n} , \tau_E' , \alpha , \hat{\mu} , \hat{\sigma} \right) - \mathcal{C}_E \left( x_{1:n} , \alpha , \hat{\mu} , \hat{\sigma} \right) \\ &\geq 
\tilde{\mathcal{C}}\left( x_{1:n} , \tau_E' \cup \{0,n\} , \frac{1}{2}\alpha \right) + \frac{\alpha}{2} \sum_{k=2}^{\hat{K}} \mathbb{I}\{s_k = e_{k-1}\} \log(n)^{1+\delta} - \left[\mathcal{C}\left( x_{1:n}\right) + 2K\frac{\alpha}{2}\log(n)^{1+\delta}\right] - \tilde{C}_7\log(n),
\end{align*}
by applying Lemma \ref{lemma:CostoftruevsCostofFalse}. The rest of the proof is identical to that of Proposition \ref{prop:MainPARAMKNOWNLENGTHATLEAST2}, with an added $O(\log(n))$ term. 

In order to be able to extend Proposition \ref{prop:MainLENGTHATLEAST2} to the case in which we allow epidemic changes of length one in variance only, as well as segments of the typical distribution which are of length one, we will prove the consistency of the following adaptation of the algorithm detecting classical changepoints we introduced in the previous section. We now let the segment costs be
\begin{equation*}
\tilde{\mathcal{C}}(x_{i:j}) = \tilde{\mathcal{C}}(x_{i:j},\{i-1,j\}) = \begin{cases}
(\hat{t}_{k+1}-\hat{t}_{k})\left( \log \left( \frac{\sum_{\hat{t}_{k}+1}^{\hat{t}_{k+1}}(x_t - \bar{x}_{(\hat{t}_{k}+1):\hat{t}_{k+1}})^2}{(\hat{t}_{k+1}-\hat{t}_{k})}\right) +1\right)  & i < j, \\
\min \left\{ \log(\tilde{\sigma}^2) + \frac{(x_i - \tilde{\mu})^2}{\tilde{\sigma}^2} , 
1 + \log( \gamma \tilde{\sigma}^2 + (x_t - \tilde{\mu})^2  )  \right\} & i = j,
\end{cases}
\end{equation*}
where $|\tilde{\mu} - \mu_{k'}| \leq D_1\sigma_{k'}\sqrt{\frac{\log(n)}{n}}$ and $|\frac{\tilde{\sigma}^2}{\sigma_{k'}^2} - 1| < D_2\sqrt{\frac{\log(n)}{n}}$ for $k'$ either $k-1$,$k$, or $k+1$, when $i$ belongs to the $k$th segment. Given $E_8$ the range of allowed $\tilde{\sigma}^2$ and $\tilde{\mu}$ is therefore guaranteed to contain the estimated typical parameters  $\hat{\sigma}^2$ and $\hat{\mu}$ when applied to $x$. The algorithm can obviously not be implemented in practice, as it requires knowledge of the true parameters. It is nevertheless a consistent method. 

To prove this, we need to define a last event set $E_9$ which controls the newly introduced segments of length one: 
\begin{equation*}
E_9 := \left\{  |x_t - \mu_{k+1}| \geq \sigma_k n^{-2+\epsilon}, \, \;  \; \, |x_t - \mu_{k-1}| \geq \sigma_k n^{-2+\epsilon}, \, \; \; \, 1 \leq t \leq n \right\},
\end{equation*}
We can prove the following probability bounds
\begin{Lemma}\label{lemma:Probabilitylength1}
	There exists a constant $\tilde{K}_9$ such that $\mathbb{P}(E_9) \geq 1 - \tilde{K}_9n^{-\epsilon}$
\end{Lemma}

We can now prove the following proposition, adapted from Proposition \ref{prop:Mainproposition} for this modified penalised cost approach: 
\begin{Proposition}\label{prop:MainpropositionMODIFIED}
	There exists a constant $\tilde{n}_7(\alpha,\delta,\tilde{\delta},\epsilon)$ such that given $E \cap E_9$, we have 
	\begin{equation*}
	\tilde{\mathcal{C}} (x_{1:n},\tau,\alpha) - \left[\mathcal{C}(x_{1,n}) +K\alpha\log(n)^{1+\delta} \right] \geq \frac{1}{5}\alpha \log(n)^{1+\delta}
	\end{equation*}
	for all $\tau \notin B$ if $n\geq \tilde{n}_7(\alpha,\delta,\tilde{\delta},\epsilon)$
\end{Proposition}

\textbf{Proof of Proposition \ref{prop:MainpropositionMODIFIED}:} Identical to the proof of Proposition \ref{prop:Mainproposition}. We just need to replace Lemma \ref{lemma:Falsepositiveremoval} by 
\begin{Lemma}\label{lemma:FalsepositiveremovalSPECIAL}
	There exists a constant $\tilde{C}_2'$ such that if $i,j$ are such that there exists some $k$ such that $t_{k-1} < i \leq j \leq t_{k}$, then given $E \cap E_8$ and $n$ large enough 
	\begin{equation*}
	\mathcal{C}\left(x_{i:j}\right) - \tilde{\mathcal{C}}\left(x_{i:j}\right) \leq \tilde{C}_2' \log(n).
	\end{equation*}  
\end{Lemma}
\noindent to also account for the newly added segments of length one. We can now prove that 
\begin{Proposition}\label{prop:Main}
	There exists a constant $\tilde{n}_8(\alpha,\delta,\tilde{\delta},\epsilon)$ such that for all $\tau_E' \notin \mathcal{B}_E$
	\begin{equation*}
	\tilde{\mathcal{C}}_E \left( x_{1:n} , \tau_E' , \alpha , \hat{\mu} , \hat{\sigma} \right) - \mathcal{C}_E \left( x_{1:n} , \alpha , \hat{\mu} , \hat{\sigma} \right) \geq \frac{1}{5}\alpha\log(n)^{1+\delta/2}
	\end{equation*}	
	holds on $E \cap E_8 \cap E_9$ for $n > \tilde{n}_8(\alpha,\delta,\tilde{\delta},\epsilon)$.
\end{Proposition}

\noindent holds even when we allow for epidemic changes of length one in variance only and do not impose that segments allocated to the typical distribution have to be of length at least two. 

\textbf{Proof of Proposition \ref{prop:Main}}: Identical to the proof of Proposition \ref{prop:MainLENGTHATLEAST2} using Proposition \ref{prop:MainpropositionMODIFIED} instead of Proposition \ref{prop:Mainproposition}.

\textbf{Proof of Theorem \ref{Thm:Consistency}}: Proposition \ref{prop:Main} proves Theorem \ref{Thm:Consistency} since Lemmata \ref{lemma:Yao}, \ref{lemma:ProbabilityofboundsIgotoftheinternet}, \ref{lemma:Robuststats}, and \ref{lemma:Probabilitylength1} show that $\mathbb{P}(E \cap E_8 \cap E_9) \geq 1 - (\tilde{K}_1 + \tilde{K}_2 + \tilde{K}_3 + \tilde{K}_4 + \tilde{K}_5 + \tilde{K}_6 + \tilde{K}_7 + \tilde{K}_8 + \tilde{K}_9)n^{-\epsilon}$.

\bibliographystyle{Chicago}
\bibliography{EPELT}

\section{Supplementary Material}

\subsection{Additional Lemmata}

\begin{Lemma}\label{lemma:BoundsOnChisquared}
	Let weights $W_1, ... , W_p > 0 $ and $A_1,...,A_p > 0$. Define $\mu = \sum_{1}^{p}W_iA_i$. Then 
	\begin{equation*}
	\min_{\lambda < 0}\left[\left(\prod_{i=1}^{p} \left(\frac{1}{1-A_i\lambda}\right)^{W_i}\right) e^{-\lambda c \mu}\right] \leq \exp \left( \left(\sum_{i=1}^pW_i\right) \left( \log(c) + 1 - c \right) \right)
	\end{equation*}
	holds for $0 < c < 1$
\end{Lemma}

\textbf{Proof of Lemma \ref{lemma:BoundsOnChisquared}}: This Lemma proves a multiplicative Chernoff lower bound for a weighted sum of chi-squared random variables. We define 
\begin{equation*}
Z(\lambda) = \left(\prod_{i=1}^{p} \left(\frac{1}{1-A_i\lambda}\right)^{W_i}\right) e^{-\lambda c \mu}
\end{equation*}
and note that it has derivative 
 \begin{equation*}
\frac{d}{d\lambda}\left(Z(\lambda)\right) = \left(\sum_{i=1}^{p} \frac{A_iW_i}{1-A_i\lambda} - c \mu\right) Z(\lambda).
 \end{equation*}
The minimise $\lambda^*$ of $Z(\lambda)$ thus satisfies
\begin{equation*}
\sum_{i=1}^{p} \frac{A_iW_i}{1-A_i\lambda^*} = c \mu.
\end{equation*}
Note that the LHS is strictly increasing from $0$ to $\mu$ as $\lambda^*$ increases from $-\infty$ to $0$. Consequently $\lambda^*$ is well defined and unique. Moreover, 
\begin{align*}
c \mu \lambda^* = \sum_{i=1}^{p} \frac{A_iW_i\lambda^*}{1-A_i\lambda^*} = - \sum_{i=1}^{p}W_i + \sum_{i=1}^{p} \frac{W_i}{1-A_i\lambda^*} \geq - \sum_{i=1}^{p}W_i + \frac{\left(\sum_{i=1}^{p}W_i\right)^2}{ \sum_{i=1}^{p} \left(W_i - W_iA_i\lambda^*\right)}& = -  \sum_{i=1}^{p}W_i + \frac{\left(\sum_{i=1}^{p}W_i\right)^2}{\sum_{i=1}^{p}W_i - \mu \lambda^*} \\
&= \left( \sum_{i=1}^{p}W_i\right) \frac{\mu \lambda^*}{\sum_{i=1}^{p}W_i -  \mu \lambda^*},
\end{align*}
with the inequality following from the fact that the arithmetic mean exceeds the harmonic mean (a special case of Jensen's inequality). This can be recomposed to yield
\begin{equation}\label{eq:lambdastarabound}
\lambda^* c \mu \leq (c-1) \sum_{i=1}^p W_i.
\end{equation}
We can now use these results to bound $Z(\lambda^*)$ by noting 
\begin{align*}
Z(\lambda^*) &= \left(\prod_{i=1}^{p} \left(\frac{1}{1-A_i\lambda^*}\right)^{W_i}\right) e^{-\lambda* c \mu} \leq \left( \frac{1}{\sum_{i=1}^pW_i}\sum_{i=1}^{p} \frac{W_i}{1-A_i\lambda^*}\right)^{\sum_{i=1}^pW_i} e^{-\lambda^* c \mu} \\ &= \left( 1 + \frac{1}{\sum_{i=1}^pW_i}\sum_{i=1}^{p} \frac{W_iA_i\lambda^*}{1-A_i\lambda^*}\right)^{\sum_{i=1}^pW_i} e^{-\lambda^* c \mu} = 
\left[\left( 1 + \frac{\lambda^* c \mu}{\sum_{i=1}^pW_i}\right)e^{-\frac{\lambda^* c \mu}{\sum_{i=1}^pW_i}}\right]^{\sum_{i=1}^pW_i},
\end{align*}
where the first inequality follows form The AMGM inequality. It can be shown by differentiation that the above bound is increasing in $\lambda^*$. Consequently, using (\ref{eq:lambdastarabound}) we have that
\begin{equation*}
Z(\lambda^*) \leq \exp \left( \left(\sum_{i=1}^pW_i\right) \left( \log(c) + 1 - c \right) \right)
\end{equation*}

\begin{Lemma}\label{lemma:c}
	Let $c < 1$ solve the equation $ c - 1 - \log(c) = t$ for some $t > 0$. Then 
	$ c-1 \geq -\sqrt{2t} $
\end{Lemma}

\textbf{Proof of Lemma \ref{lemma:c}}: This Lemma helps bound $c(D,n)$ from the event set $E_7$.
\begin{equation*}
t = c - 1 - \log(c) = c - 1 - \log(1 - (1-c)) \geq  c - 1 + (1-c) + \frac{1}{2} (1-c)^2 = \frac{1}{2} (1-c)^2
\end{equation*}
which implies that $ c-1 \geq -\sqrt{2t} $. This finishes the proof.

\setcounter{Lemma}{0} 
\setcounter{Proposition}{1} 

\subsection{Proposition \ref{Prop:FPCONTROL} and Proof}

\begin{Proposition}
	Let $x_1,...,x_n$ be i.i.d.\ $N(\mu,\sigma^2)$ distributed, for known $\mu$ and $\sigma$. Then there exist constants $C_1$ and $C_2$ such that when the penalty for point anomalies, $\beta'$, and the penalty for collective anomalies, $\beta$, satisfy $\beta' \geq 2t$ and $\beta \geq 2(2+2t+2\sqrt{2t})$, then
	\begin{equation*}
	\mathbb{P}\left(\hat{K} = 0 , \; \hat{O} = \emptyset \right) \geq 1 - C_1ne^{-t} - C_2\left(ne^{-t}\right)^2.
	\end{equation*}
\end{Proposition}

\textbf{Proof of Proposition \ref{Prop:FPCONTROL}}: Assume, without loss of generality, that $\mu=0$ and $\sigma = 1$. This Proof has two parts. The first one consist of showing that $\mathbb{P}\left(\hat{O} = \emptyset \right) \geq 1 - C_1'ne^{-t}$, the second one consists of showing that $\mathbb{P}\left(\hat{K} = 0\right) \geq 1 - C_1'ne^{-t} - C_2'\left(ne^{-t}\right)^2$

\textbf{Part 1:} We begin by proving that $\mathbb{P}\left(\hat{O} = \emptyset \right) \geq 1 - C_1'ne^{-t}$. Note that $x_i^2 \sim \chi^2_1$. We define $a_+(t)$ and $a_-(t)$ via the equation
\begin{equation*}
\mathbb{P}\left(\chi^2_1 > a_+(t) \right) = \mathbb{P}\left(\chi^2_1 < a_-(t) \right) = e^{-t}
\end{equation*}
for $t > \log(1/2)$ Therefore, by a Bonferroni correction,
\begin{equation*}
\mathbb{P}\left( a_-(t)<x_i^2<a_+(t)\right) \geq 1 - 2e^{-t} . 
\end{equation*}
Note that 
\begin{equation*}
x_i^2 - \log\left(\gamma + x_i^2 \right) - 1 < x_i^2 - \log\left( x_i^2 \right) - 1
\end{equation*}
and that the function $f(x) = (x - 1) - \log(x) $ is decreasing for $x \leq 1 $ and increasing thereafter. Consequently
\begin{equation*}
x_i^2 - \log\left( x_i^2 \right) - 1 \leq \max(f(a_+(t)),f(a_-(t)))
\end{equation*}
with probability $1-2e^{-t}$. We also know that the Chernoff bounds
\begin{equation*}
\mathbb{P}\left(\chi^2_1 > a \right) \leq \exp\left( \frac{-a + 1 + \log(a)}{2}\right)  \; \; \; \; \; \; \equiv \; \; \; \; \; \; a-1-\log(a) \leq -2 \log(\mathbb{P}\left(\chi^2_1 > a \right) )
\end{equation*}
and 
\begin{equation*}
\mathbb{P}\left(\chi^2_1 < a \right) \leq \exp\left( \frac{-a + 1 + \log(a)}{2}\right)  \; \; \; \; \; \; \equiv \; \; \; \; \; \; a-1-\log(a) \leq -2 \log(\mathbb{P}\left(\chi^2_1 < a \right) )
\end{equation*}
hold for $a \geq 1$ and $a \leq 1$ respectively. Hence, $\max(f(a_+),f(a_-)) \leq 2t$ and thus
\begin{equation*}
x_i \notin \hat{O} \; \; \; \equiv \; \; \;  x_i^2 - \log\left(\gamma + x_i^2 \right) - 1 \leq 2t.
\end{equation*}
holds with probability with probability $1-2e^{-t}$. A Bonferroni correction over $x_1,...,x_n$ then gives the result. 

\textbf{Part 2:} We now prove that $\mathbb{P}\left(\hat{K} = \emptyset \right) \geq 1 - C_1'ne^{-t} - C_2'\left(ne^{-t}\right)^2$. First of all, note that this is equivalent to showing that \small
\begin{equation*}
\mathbb{P}\left( \sum_{s=i}^{j} x_s^2 - (j-i+1)\left[1 + \log \left( \frac{\sum_{s=i}^{j} (x_s-\bar{x}_{i:j})^2}{(j-i+1)}\right)\right] < 4+4t+4\sqrt{2t}\;\;\;\; 1 \leq i \leq j - l + 1 < j \leq n \right) \geq 1 - C_1'ne^{-t} - C_2'\left(ne^{-t}\right)^2.
\end{equation*} \normalsize
For a fixed $i$ and $j$, writing $a = j-i+1$, we have that $\sum_{s=i}^{j} x_s^2 = \sum_{s=i}^{j} (x_s-\bar{x}_{i:j})^2 + a\left(\bar{x}_{i:j}\right)^2$.  Note that $a \left(\bar{x}_{i:j}\right)^2 \sim \chi_1^2$ and $\sum_{s=i}^{j} \left(x_s -  \bar{x}_{i:j} \right)^2 \sim \chi_{a-1}^2$. Moreover, these two random variables are independent. Consequently, the MGF of 
\begin{equation*}
a \left(\bar{x}_{i:j}\right)^2 - a \log\left(\frac{\sum_{s=i}^{j} \left(x_s -  \bar{x}_{i:j} \right)^2}{a}\right)  + \sum_{s=i}^{j} \left(x_s -  \bar{x}_{i:j} \right)^2 - a
\end{equation*}
is given by 
\begin{align*}
&\sqrt{\frac{1}{1-2\lambda}} \int_{0}^{\infty} \left(\frac{a}{x}\right)^{a\lambda}e^{\lambda x -\lambda a} \left( \Gamma \left(\frac{a-1}{2}\right)2^{\frac{a-1}{2}}\right)^{-1}  x ^{\frac{a-1}{2} - 1} e^{-x/2}dx \\
&=\left(\frac{1}{1-2\lambda}\right)^{\frac{1}{2}a(1-2\lambda)} \frac{a^{\lambda a}}{e^{\lambda a} 2^{\lambda a} } \frac{\Gamma \left(\frac{a-1}{2} - \lambda a \right)}{\Gamma \left(\frac{a-1}{2}\right)} = \left(\frac{1}{1-2\lambda}\right)^{\frac{1}{2}a(1-2\lambda)} \frac{a^{\lambda a}}{e^{\lambda a} 2^{\lambda a} } \frac{\frac{a-1}{2}}{\frac{a-1}{2} - \lambda a } \frac{\Gamma \left(\frac{a+1}{2} - \lambda a \right)}{\Gamma \left(\frac{a+1}{2}\right)}
\end{align*}
We now use the following Stirling bounds from \cite{artin2015gamma} 
\begin{equation*}
1 \leq \Gamma(x) \frac{e^x}{\sqrt{2\pi}x^{x-\frac{1}{2}}} \leq e^{\frac{1}{12x}}
\end{equation*}
which imply that: 
\begin{equation*}
\frac{\Gamma \left(\frac{a+1}{2} - \lambda a \right)}{\Gamma \left(\frac{a+1}{2}\right)} \leq e^{\frac{1}{12\left(\frac{a+1}{2} - \lambda a \right)}} e^{\lambda a} \frac{\left(\frac{a+1}{2} - \lambda a \right) ^{\left(\frac{a}{2} - \lambda a\right)}}{\left(\frac{a+1}{2} \right) ^{\left(\frac{a}{2} \right)}} \leq e^{1/6} e^{\lambda a}\left( 1 - 2 \frac{a}{a+1}\lambda\right)^{a/2} \left(\frac{a+1}{2} - \lambda a \right) ^{ - \lambda a},
\end{equation*}
since $\lambda \leq \frac{1}{2}$. Consequently, the MGF is bounded by:
\begin{align*}
&e^{1/6}\left(\frac{1}{1-2\lambda}\right)^{\frac{1}{2}a(1-2\lambda)} \frac{a^{\lambda a}}{ 2^{\lambda a} } \frac{ 1 }{1 - 2 \frac{a}{a-1}\lambda }\left( 1 - 2 \frac{a}{a+1}\lambda\right)^{a/2} \left(\frac{a+1}{2} - \lambda a \right) ^{ - \lambda a} \\
&= e^{1/6} \frac{ 1 }{1 - 2 \frac{a}{a-1}\lambda } \left[ \left( \frac{1-2\lambda}{\frac{a+1}{a} - 2 \lambda}\right)^{2\lambda} \left(\frac{1-2\frac{a}{a+1}\lambda}{1 - 2 \lambda} \right)\right] ^{a/2} = e^{1/6} \frac{ 1 }{1 - 2 \frac{a}{a-1}\lambda } \left[ \left( \frac{1}{1 + \frac{1}{a(1-2\lambda)} }\right)^{2\lambda} \left(1 + \frac{2\frac{1}{a+1}\lambda}{1 - 2 \lambda} \right)\right] ^{a/2} \\
& \leq e^{1/6} \frac{ 1 }{1 - 2 \frac{a}{a-1}\lambda } \left[ \left( \frac{1}{1 + \frac{1}{a(1-2\lambda)} }\right)^{2\lambda} \left(1 + \frac{1}{a(1 - 2 \lambda)} \right)\right] ^{a/2} = e^{1/6} \frac{ 1 }{1 - 2 \frac{a}{a-1}\lambda } \left[  1 + \frac{1}{a(1 - 2 \lambda)} \right] ^{a(1-2\lambda)/2} \\
&  \leq e^{1/6}e^{1/2} \frac{ 1 }{1 - 2 \frac{a}{a-1}\lambda } \leq e \frac{ 1 }{1 - 2 \frac{a}{a-1}\lambda }.
\end{align*}
This implies the following Chernoff bound 
\begin{equation*}
\mathbb{P}\left(\sum_{s=i}^{j}x_s^2 - a \log\left(\frac{\sum_{s=i}^{j} \left(x_s -  \bar{x}_{i:j} \right)^2}{a}\right)  - a > \frac{a}{a-1}\left(2 + 2t + 2\sqrt{2t}\right) \right) \leq ee^{-t}
\end{equation*}
and therefore, as $a \geq l \geq 2$
\begin{equation*}
\mathbb{P}\left(\sum_{s=i}^{j}x_s^2 - a \log\left(\frac{\sum_{s=i}^{j} \left(x_s -  \bar{x}_{i:j} \right)^2}{a}\right)  - a > 2 \cdot 2 + 2\frac{a}{a-1}t' + 2\sqrt{2\cdot 2\frac{a}{a-1}t'} \right) \leq ee^{-t'}.
\end{equation*}
substituting $t' = 2\frac{a-1}{a}t$ yields 
\begin{equation*}
\mathbb{P}\left(\sum_{s=i}^{j}x_s^2 - a \log\left(\frac{\sum_{s=i}^{j} \left(x_s -  \bar{x}_{i:j} \right)^2}{a}\right)  - a > 2\cdot 2 + 2\cdot 2t + 2\sqrt{2\cdot 2t} \right) \leq ee^{-2\frac{a-1}{a}t}.
\end{equation*}
Consequently, by a Bonferroni correction,
\begin{equation*}
\mathbb{P}\left(\exists i,j : \sum_{s=i}^{j}x_s^2 - (j-i+1) \left( \log\left(\frac{\sum_{s=i}^{j} \left(x_s -  \bar{x}_{i:j} \right)^2}{j-i+1}\right)  - 1 \right)> 4 + 4t + 4\sqrt{t} \right) \leq \sum_{a = l}^{n} n  ee^{-2\frac{a-1}{a}t} \leq n  e\sum_{a = 2}^{n} e^{-2\frac{a-1}{a}t}.
\end{equation*}
Now, 
\begin{align*}
& \sum_{a = 2}^{n} e^{-2\frac{a-1}{a}t} = e^{-t} + \sum_{a = 3}^{n} e^{-2\frac{a-1}{a}t} \leq e^{-t} + \int_{a = 2}^{n} e^{-2\frac{a-1}{a}t}da = e^{-t} + e^{-2t}\int_{a = 2}^{n}e^{\frac{2}{a}t}da = e^{-t} + e^{-2t}\int_{x = \frac{1}{n}}^{\frac{1}{2}}\frac{1}{x^2}e^{2tx}dx.
\end{align*}
Next note that 
\begin{equation*}
e^{-2t}\int_{x = \frac{1}{n}}^{\frac{1}{2t}}\frac{1}{x^2}e^{2tx}dx \leq e e^{-2t} \int_{x = \frac{1}{n}}^{\infty}\frac{1}{x^2}e^{2tx}dx = e n e^{-2t} ,
\end{equation*} 
which proves the result if $t \leq 1$. If $t > 1$ , the proof can be obtained by noting that
\begin{equation*}
e^{-2t}\int_{x =\frac{1}{2t}}^{\frac{1}{2}}\frac{1}{x^2}e^{2tx}dx \leq e^{-2t}\frac{1}{2} \max_{\frac{1}{2t} \leq x \leq \frac{1}{2}}\left(\frac{1}{x^2}e^{2tx}\right) \leq e^{-2t}\frac{1}{2}4e^{t} = 2e^{-t}.
\end{equation*}

\subsection{Proposition \ref{Prop:TPCONTROL} and Proof}

\begin{Proposition}
	Let $(\hat{\mu},\hat{\sigma})$ and $\beta'$ correspond to the estimated parameters of the true distribution and the penalty for point anomalies used by CAPA respectively. Assume, moreover, that $\gamma$ satisfies $1 \geq \gamma \geq \exp\left(-\beta'\right)$. Then, there exists a constant $K(\beta') \asymp \beta'$ such that
	\begin{equation*}
	i \notin \hat{O} \implies (x_i - \hat{\mu})^2 <\hat{\sigma}^2  K(\beta')
	\end{equation*}
	and 
	\begin{equation*}
	(x_i - \hat{\mu})^2 > \hat{\sigma}^2  K(\beta') \implies i \notin \hat{O} \lor \exists k \in \{1,...,\hat{K}\}: i \in \{\hat{s}_k+1,...,\hat{e}_k\}
	\end{equation*}
\end{Proposition}

\textbf{Proof of Proposition \ref{Prop:TPCONTROL}}: This proposition follows from the fact that CAPA will not fit $x_i$ as a point anomaly if
\begin{equation*}
\left(\frac{x-\hat{\mu}}{\hat{\sigma}}  \right)^2 - 1 - \log\left(\gamma+\left(\frac{x-\hat{\mu}}{\hat{\sigma}}  \right)^2\right) < \beta'
\end{equation*}
and not fit not $x_i$ as typical if 
\begin{equation*}
\left(\frac{x-\hat{\mu}}{\hat{\sigma}}  \right)^2 - 1 - \log\left(\gamma+\left(\frac{x-\hat{\mu}}{\hat{\sigma}}  \right)^2\right) > \beta'.
\end{equation*}
It can be show by differentiation that the function $f(y) = y - 1 - \log(y+\gamma)$ is decreasing from $0$ to $(y+\gamma)^{-1}$ and increasing thereafter. Since $f(0) < \beta'$, by the lower bound on $\gamma$ and since $f(y) \rightarrow \infty$ as $y \rightarrow \infty$, there exists a constant $K(\beta')$ solving the equation $f(K(\beta')) = \beta'$ such that $f(y) < \beta'$ if $y < K(\beta')$ and $f(y) > \beta'$ if $y > K(\beta')$. Next note that
\begin{equation}\label{eq:boundlow}
f(\beta') = \beta' - 1 - \log\left( \gamma + \beta' \right) <  \beta' - \log\left( e^{-\beta'} + \beta' \right) < \beta' - \log\left( 1 - \beta' + \beta' \right) = \beta'.
\end{equation}
Moreover, we can show that
\begin{equation*}
f(1+\beta'+\sqrt{2(\beta'+\gamma)})  - \beta' = \sqrt{2(\beta'+\gamma)} - \log(1+(\beta'+\gamma)+\sqrt{2(\beta'+\gamma)})
\end{equation*}
is equal to 0 when $z = \beta'+\gamma = 0 $. Moreover the derivative of the above expression with respect to $z$ is given by 
\begin{equation*}
\frac{z}{1+z+\sqrt{2z}}
\end{equation*}
which is positive for all $z > 0$. Since $z = \beta'+\gamma>0$ the following result also holds:
\begin{equation}\label{eq:boundhigh}
f(1+\beta'+\sqrt{2(\beta'+\gamma)}) > \beta'.
\end{equation}
This can be deduced from the fact that equality holds  $\beta' = 0$ and comparing the derivatives. Equations (\ref{eq:boundlow}) and (\ref{eq:boundhigh}) show that
\begin{equation*}
\beta' < K(\beta') < \beta' + 1 +\sqrt{2(\beta'+\gamma)},
\end{equation*} 
which finishes the proof. 

\subsection{Proofs of Lemmata Stated in the Appendix}

\begin{Lemma}
	(Yao 1988) $\mathbb{P}(E_1) > 1 - \tilde{K_1}n^{-\epsilon}$, for some constant $\tilde{K_1}$.
\end{Lemma}

\textbf{Proof of Lemma \ref{lemma:Yao}:} See \cite{yao1988estimating}.

\begin{Lemma}
	$\mathbb{P}(E_2) > 1 - \tilde{K_2}n^{-\epsilon}$, $\mathbb{P}(E_3) > 1 - \tilde{K_3}n^{-\epsilon}$, $\mathbb{P}(E_4) > 1 - \tilde{K_4}n^{-\epsilon}$, $\mathbb{P}(E_5) > 1 - \tilde{K_5}n^{-\epsilon}$, $\mathbb{P}(E_6) > 1 - \tilde{K_6}n^{-\epsilon}$, and $\mathbb{P}(E_7) > 1 - \tilde{K_7}n^{-\epsilon}$for some constants $\tilde{K_2}$,$\tilde{K_3}$,$\tilde{K_4}$, $\tilde{K_5}$, $\tilde{K_6}$, and $\tilde{K_7}$.
\end{Lemma}

\textbf{Proof of Lemma \ref{lemma:ProbabilityofboundsIgotoftheinternet}:}  We note that $Y_{i:j}\sim \chi^2_{a-1}$. \cite{laurent2000adaptive} proved that
\begin{equation*}
\mathbb{P}\left(-2\sqrt{kx} \leq \chi^2_k - k \leq 2\sqrt{kx} +2x \right) \geq 1 - 2e^{-x}. 
\end{equation*}
Therefore: 
\begin{equation*}
\mathbb{P}\left(-2\sqrt{(a-1)(2+\epsilon)\log(n)} \leq Y_{i:j} - (a-1) \leq 2\sqrt{(a-1)(2+\epsilon)\log(n)} +2(2+\epsilon)\log(n) \right) \geq 1 - 2n^{-(2+\epsilon)}. 
\end{equation*}
A Bonferroni correction therefore gives $\mathbb{P}(E_2 \cap E_6) > 1 - 2n^{-\epsilon}$. 

We can derive the following Chernoff bound for $k \geq 1$ and $0 \leq \tilde{c} < 1$:
\begin{align*}
\mathbb{P}\left( \chi ^2 _k \leq k \tilde{c}\right) =  \mathbb{P}\left( \exp\left[\theta(\chi ^2 _k - k \tilde{c} )\right]\geq 1 \right) \leq \mathbb{E}\left( \exp\left[\theta(\chi ^2 _k - k \tilde{c} )\right]\right) = e^{-k\tilde{c}\theta}\mathbb{E}\left( e^{\theta\chi ^2 _k}\right) = e^{-k\tilde{c}\theta} \left(\frac{1}{1-2\theta}\right)^{k/2},
\end{align*}
holds for all $\theta<0$. Setting $\theta = \frac{1}{2} (1 -  \frac{1}{\tilde{c}})$ we thus get 
\begin{equation*}
\mathbb{P}\left( \chi ^2 _k \leq k \tilde{c}\right) \leq \exp \left( - \frac{k}{2} \left( \tilde{c} -1-\log(\tilde{c})\right) \right).
\end{equation*} 
Thus if we let $c(a,n) < 1$ be such that 
\begin{equation*}
\frac{a}{2} \cdot \frac{c(a,n)-1-\log(c(a,n))}{2} = (2+\epsilon)\log(n),
\end{equation*}
and write $c := c(a,n)$ for simplicity, we have 
\begin{equation*}
\mathbb{P}\left( Y_{i:j} \leq c(a-1)  \right) \leq \exp \left( -\frac{a-1}{2}\left( c - 1 - \log(c) \right) \right) \leq \exp \left( -\frac{a}{4}\left( c - 1 - \log(c) \right) \right) = n^{-(2+\epsilon)},
\end{equation*}
for $ a \geq 2$. A Bonferroni correction then gives $\mathbb{P}(E_3) > 1 - n^{-\epsilon}$.

Next we note that 
\begin{equation*}
\frac{\sigma_k \eta_{t_{k}+1} + \mu_{k+1} - \mu_{k} - \sigma_{k} \eta_{t_{k}}}{\sqrt{\sigma_{k+1}\sigma_{k}}} \sim N \left( \frac{\mu_{k+1} - \mu_{k}}{\sqrt{\sigma_{k+1}\sigma_{k}}} ,\frac{\sigma_{k+1}^2 + \sigma_{k}^2}{\sigma_{k+1}\sigma_{k}} \right).
\end{equation*}
Consequently, we have that: 
\begin{equation*}
\mathbb{P} \left(\frac{|\sigma_{k+1} \eta_{t_{k}+1} + \mu_{k+1} - \mu_{k} - \sigma_{k} \eta_{t_{k}}| }{\sqrt{\sigma_{k+1}\sigma_{k}}}\leq  n^{-\epsilon}\right) \leq \sqrt{\frac{2\sigma_{k+1}\sigma_{k}}{\pi(\sigma_{k+1}^2 + \sigma_{k-1}^2)}} n^{-\epsilon} \leq \sqrt{\frac{1}{\pi}}n^{-\epsilon}.
\end{equation*}
A Bonferroni correction then gives $\mathbb{P}(E_5) > 1 - K/\sqrt{\pi}n^{-\epsilon}$.

Finally, we have: 
\begin{equation*}
x_t-\bar{x}_{t_k:(t_k+2)} \sim \begin{cases}
N\left(\frac{2\mu_{k} - 2\mu_{k+1}}{3} , \frac{4\sigma_{k}^2 + 2\sigma_{k+1}^2}{9}\right)  & t = t_k, \\
N\left(\frac{\mu_{k+1} - \mu_{k}}{3} , \frac{2\sigma_{k}^2 + 5\sigma_{k+1}^2}{9}\right)  & t = t_k+1, t_k+2,
\end{cases}
\end{equation*}
which means that 
\begin{align*}
(x_t-\bar{x}_{t_k:(t_k+2)})^2 \geq \frac{4\sigma_{k}^2 + 2\sigma_{k+1}^2}{9}n^{-\epsilon} , \; \; \; & t = t_k, \\
(x_t-\bar{x}_{t_k:(t_k+2)})^2 \geq \frac{1\sigma_{k}^2 + 5\sigma_{k+1}^2}{9}n^{-\epsilon} , \; \; \; & t = t_k+1, t_k+2 ,
\end{align*}
holds with probability exceeding $1-3n^{-\epsilon}$. Adding up the three inequalities then gives
\begin{equation*}
\sum_{t=t_k}^{t_k+2}(x_t-\bar{x}_{t_k:(t_k+2)})^2 \geq \frac{12\sigma_{k+1}^2 + 6\sigma_{k}^2 }{9}n^{-\epsilon} \geq 2\sigma_{k+1}^{4/3}\sigma_{k}^{2/3}n^{-\epsilon}.
\end{equation*}
By a similar argument,
\begin{equation*}
\sum_{t=t_k-1}^{t_k+1}(x_t-\bar{x}_{(t_k-1):(t_k+1)})^2 \geq 2\sigma_{k+1}^{2/3}\sigma_{k}^{4/3}n^{-\epsilon}
\end{equation*}
must also hold with probability $1-3n^{-\epsilon}$. A Bonferroni correction then gives $\mathbb{P}(E_4) \geq 1-6K_4n^{-\epsilon}$. 

Next, note that the MGF of
\begin{equation*}
\sum_{t_k-D+1}^{t_k+D}(x_t-\bar{x}_{(t_k-D+1):(t_k+D)})^2 = \sigma_{k}^2Y_{(t_k-D+1):t_k}+\sigma_{k+1}^2Y_{(t_k+1):(t_k+D)} + \frac{D}{2}\left(\mu_{k} + \sigma_{k}\bar{\eta}_{(t_k-D+1):t_k} - \mu_{k+1} - \sigma_{k+1} \bar{\eta}_{(t_k+1):(t_k+D)}\right)^2
\end{equation*} 
is given by
\begin{equation*}
\left(\frac{1}{1-2\sigma_{k}^2\lambda}\right)^{\frac{D-1}{2}}\left(\frac{1}{1-2\sigma_{k+1}^2\lambda}\right)^{\frac{D-1}{2}}\left(\frac{1}{1-(\sigma_{k}^2+\sigma_{k+1}^2)\lambda}\right)^\frac{1}{2} \exp \left( \frac{\frac{D}{2}(\mu_{k+1} - \mu_{k})^2\lambda}{1 - (\sigma_{k}^2+\sigma_{k+1}^2)\lambda}\right)
\end{equation*}
since $Y_{(t_k-D+1):t_k}$ and $Y_{(t_k+1):(t_k+D)}$ are independently $\chi^2_{D-1}$ distributed. Moreover, $\bar{\eta}_{(t_k-D+1):t_k}$ and $\bar{\eta}_{(t_k+1):(t_k+D)}$ are normally distributed and independent of the chi-squared random variables. The MGF is therefore the product of two chi-squared random variables and that of a noncentral chi-squared. Noting that 
\begin{equation*}
e^x = \sqrt{\frac{1}{e^{-2x}}} \leq \sqrt{\frac{1}{1-2x}}
\end{equation*}
holds for all $x < 1/2$ shows that the MGF can be bounded by
\begin{equation*}
\left(\frac{1}{1-2\sigma_{k}^2\lambda}\right)^{\frac{D-1}{2}}\left(\frac{1}{1-2\sigma_{k+1}^2\lambda}\right)^{\frac{D-1}{2}}\left(\frac{1}{1-2\left(\frac{\sigma_{k}^2+\sigma_{k+1}^2}{2}+\frac{D}{2}(\mu_{k+1} - \mu_{k})^2\right)\lambda}\right)^\frac{1}{2}.
\end{equation*}
Lemma \ref{lemma:BoundsOnChisquared} then proves the following Chernoff bound for $0 < c < 1$: 
\begin{equation*}
\mathbb{P}\left(\sum_{t_k-D+1}^{t_k+D}(x_t-\bar{x}_{(t_k-D+1):(t_k+D)})^2 < 2D\sigma_{k+1}\sigma_{k}c \left(1+\left(1-\frac{1}{2D}\right)\triangle_{\sigma,k}^2 +\frac{\triangle_{\mu,k}^2}{4} \right) \right) < \exp \left(\frac{2D-1}{2}(\log(c)-1-c)\right).
\end{equation*}
Since $D \geq 1$ this implies,
\begin{equation*}
\mathbb{P}\left(\sum_{t_k-D+1}^{t_k+D}(x_t-\bar{x}_{(t_k-D+1):(t_k+D)})^2 < 2D\sigma_{k+1}\sigma_{k}c \left(1+\frac{1}{2}\triangle_{\sigma,k}^2 +\frac{1}{4}\triangle_{\mu,k}^2 \right) \right) < \exp \left(\frac{D}{2}(\log(c)-1-c)\right).
\end{equation*}
and therefore, 
\begin{equation*}
\mathbb{P}\left(\sum_{t_k-D+1}^{t_k+D}(x_t-\bar{x}_{(t_k-D+1):(t_k+D)})^2 < 2D\sigma_{k+1}\sigma_{k}c(D,n) \exp\left(\triangle_{k}^2 \right) \right) < \exp \left(\frac{D}{2}(\log(c(D,n))-1-c(D,n))\right) = n^{-2+\epsilon}.
\end{equation*}
A Bonferroni correction over all possible $t_k$ and $D$, of which there are guaranteed to be fewer than $n^2$ gives $\mathbb{P}\left(E_7\right) \geq 1 - n^{-\epsilon}$. This finishes the proof.

\begin{Lemma}
	There exists a constant $\tilde{C}_1$ such that $Y_{i:j} - a - a\log(Y_{i:j} /a)  \leq \tilde{C}_1 \log(n)$  holds on $E$ for all $1 \leq i < j \leq n$.
\end{Lemma}

\textbf{Proof of Lemma \ref{lemma:Yijbounds}:} Consider the function $f(x) = x - a - a\log(x/a)$. This function decreases monotonically on $(0,a)$ and increases monotonically on $(a,\infty)$. Since $E_2$ and $E_3$ bound $Y_{i:j}$ from above and below respectively we only have to show that $Y_{i:j} - a - a\log(Y_{i:j} /a)  \leq \tilde{C}_1 \log(n)$ holds for the bounds in order to prove the lemma.  

\textbf{Part 1: Upper bound}: By $E_2$ there exist constants $M$ and $M'$ such that $Y_{i:j} \leq a + M \sqrt{a\log(n)} + M'\log(n)$. Substituting this upper bound for $Y_{i:j}$ gives: 
\begin{align}\label{eq:upperbound}
Y_{i:j} - a - a\log\left(\frac{Y_{i:j}}{a}\right) \leq M \sqrt{a\log(n)} + M'\log(n) - a\log\left(1 + \frac{M \sqrt{a\log(n)} + M'\log(n)}{a}\right).
\end{align}
\textbf{Case 1}: $a \leq \log(n)$. In that case we can bound equation (\ref{eq:upperbound}) by 
\begin{align*}
M \sqrt{a\log(n)} + M'\log(n) \leq (M+M')\log(n).
\end{align*}
\textbf{Case 2}: $a \geq \log(n)$. We can use the fact that $\log(1+x) \geq x - x^2, \; \; \forall x>0$ to bound equation (\ref{eq:upperbound}) by 
\begin{align*}
\frac{(M \sqrt{a\log(n)} + M'\log(n))^2}{a} \leq (M+M')^2\log(n).
\end{align*}

\textbf{Part 2: Lower bound}: $E_3$ implies that $Y_{i:j} \geq c(a,n)(a-1)$. Substituting this bound gives 
\begin{align*}
&Y_{i:j} - a - a\log\left(\frac{Y_{i:j}}{a}\right) \leq a(c(a,n)-1-\log(c(a,n)) - c -a\log\left(\frac{a-1}{a}\right) \\ &\leq  4(4+\epsilon)\log(n) + a \log\left(\frac{a}{a-1}\right) \leq 4(4+\epsilon)\log(n) + \frac{a}{a-1} \leq 4(4+\epsilon)\log(n)+2.
\end{align*}
This finishes the proof.
\begin{Lemma}
	Let $i,j$ be such that there exists some $k$ such that $t_{k-1} < i < j \leq t_{k}$. The following holds given $E$ :
	\begin{equation*}
	0 \leq \mathcal{C}\left(x_{i:j}\right) - \tilde{\mathcal{C}}\left(x_{i:j}\right) \leq \tilde{C}_2 \log(n).
	\end{equation*}  
\end{Lemma}

\textbf{Proof of Lemma \ref{lemma:Falsepositiveremoval}: } This lemma bounds the reduction in cost we can obtain by using a mean and variance fitted to a segment rather than the true mean and variance of the segment. The left bound follows from the fact $\tilde{\mathcal{C}}\left(x_{i:j}\right)$ fits the mean and variance to minimise the log likelihood on the segment $x_{i:j}$. The right bound follows from Lemma \ref{lemma:Yijbounds} and $E_1$. Indeed,
\begin{align*}
\mathcal{C}\left(x_{i:j}\right) - \tilde{\mathcal{C}}\left(x_{i:j}\right) &=(j-i+1)\log(\sigma_k^2) +\sum_{t=i}^{j}\eta_t^2 - (j-i+1)\left( \log\left(\frac{\sigma_k^2Y_{i:j}}{j-i+1}\right)  + 1\right) \\
& =  a \bar{\eta_{i:j}}^2 + Y_{i:j} - a \log \left( \frac{Y_{i:j}}{a} \right) - a \leq \left(\tilde{C}_1 + 4 + \epsilon\right) \log(n),
\end{align*}
which finishes the proof. 

\begin{Lemma}
	Let $i,j$ be such that $\exists k$ such that $t_{k-1} = i < j \leq t_{k}$ or $t_{k-1} < i < j = t_{k}+1$. The following then holds given $E$
	\begin{equation*}
	\mathcal{C}\left(x_{i:j}\right) - \tilde{\mathcal{C}}\left(x_{i:j}\right) \leq \tilde{C}_3 \log(n)
	\end{equation*}  
\end{Lemma}

\textbf{Proof of Lemma \ref{lemma:Falsepositiveremovalforspecialcases}:} This Lemma is very similar to Lemma \ref{lemma:Falsepositiveremoval}, except that we slightly relax the constraint that all the data has to be located between two changepoints. This is needed because of the minimum segment length of two. We will prove this lemma for the case where $t_{k-1} = i$, the other case being very similar. We consider 3 cases: 

\textbf{Case 1}: $j=t_{k-1}+1$. We have that: 
\begin{align*}
&\mathcal{C}\left(x_{i:j}\right) - \tilde{\mathcal{C}}\left(x_{i:j}\right) = \log(\sigma_k^2)+\log(\sigma_{k-1}^2)+\eta_{t_{k-1}}^2+\eta_{t_{k-1}+1}^2 - 2\log\left({\frac{(x_{t_k+1} - x_{t_k})^2}{4}}\right) - 2\\
&\leq (8+2\epsilon)\log(n) - 2\log\left({\frac{(x_{t_k+1} - x_{t_k})^2}{4\sigma_{k-1}\sigma_k}}\right) -2 \leq( 8+4\epsilon)\log(n) + 2\log(4) -2,
\end{align*}
where the first inequality follows from $E_1$ and the second from $E_5$.

\textbf{Case 2}: $j=t_{k-1}+2$. We have: 
\begin{align*}
&\mathcal{C}\left(x_{i:j}\right) - \tilde{\mathcal{C}}\left(x_{i:j}\right) = 2\log(\sigma_k^2)+\log(\sigma_{k-1}^2)+ \sum_{t=t_{k-1}}^{t_{k-1}+2}\eta_{t}^2 - 3\log\left({\frac{\sum_{t=t_{k-1}}^{t_{k-1}+2}(x_t - \bar{x}  _{(t_{k-1}):(t_{k-1}+2)})^2}{3}}\right) - 3\\
&\leq 2\log(\sigma_k^2)+\log(\sigma_{k-1}^2) + (12+3\epsilon)\log(n) - 3 \log \left(\frac{n^{-\epsilon} \sigma_k^{4/3}\sigma_{k-1}^{2/3}}{3}\right) -3 \\
& = (12+6\epsilon)\log(n) + 3\log(3) - 3,
\end{align*}
where the inequality follows from $E_1$ and $E_4$.

\textbf{Case 3}: $j>t_{k-1}+2$. We have: 
\begin{align*}
\mathcal{C}\left(x_{i:j}\right) - \tilde{\mathcal{C}}\left(x_{i:j}\right) &\leq \left[\mathcal{C}\left(x_{i:(i+1)}\right) - \tilde{\mathcal{C}}\left(x_{i:(i+1)}\right)\right] + \left[\mathcal{C}\left(x_{(i+2):j}\right) - \tilde{\mathcal{C}}\left(x_{(i+2):j}\right)\right] \\ &\leq ( 8+4\epsilon)\log(n) + 2\log(4) -2 + \tilde{C}_2 \log(n),
\end{align*}
where the second inequality follows from case 1 and Lemma \ref{lemma:Falsepositiveremoval}.

\begin{Lemma}
	Let $a,b,c \in \tau$ for some partition $\tau$ of $x_{i,j}$ such that $\exists k$ such that $t_{k-1} < a < b < c \leq t_{k}$. Then, 
	\begin{equation*}
	\tilde{\mathcal{C}}\left(x_{i:j},\tau,\alpha\right) - \tilde{\mathcal{C}}\left(x_{i:j},\tau_{-b},\alpha \right) \geq \frac{3}{4}\alpha \log(n)^{1+\delta},
	\end{equation*}  
	where $\tau_{-b} = \tau \setminus \{b\} $ holds on $E$ for large enough $n$.
\end{Lemma}

\textbf{Proof of Lemma \ref{lemma:Combiningstuff}:} This lemma applies Lemma \ref{lemma:Falsepositiveremoval} to show that removing false positives reduces the overall cost. 
\begin{align*}
&\tilde{\mathcal{C}}\left(x_{i:j},\tau,\alpha\right) - \tilde{\mathcal{C}}\left(x_{i:j},\tau_{-b},\alpha \right) = \tilde{\mathcal{C}}\left( x_{(a+1):b} \right) + \tilde{\mathcal{C}}\left( x_{(b+1):c} \right) - \tilde{\mathcal{C}}\left( x_{(a+1):c} \right) + \alpha \log(n)^{1+\delta} \\
&\geq  \mathcal{C}\left( x_{(a+1):b} \right) + \mathcal{C}\left( x_{(b+1):c} \right) - \mathcal{C}\left( x_{(a+1):c} \right) + \alpha \log(n)^{1+\delta} - 2\tilde{C}_2 \log(n) \geq \frac{3}{4}\alpha \log(n)^{1+\delta},
\end{align*}
for large enough $n$. 
\begin{Lemma}
	For all $\alpha > 0$, there exists a constant $\tilde{\kappa}(\alpha,\epsilon)$  such that $\tilde{\mathcal{C}}\left( x_{i:j} \right) -( \mathcal{C}\left( x_{i:t_{k}}\right) + \mathcal{C}\left( x_{(t_{k}+1):j}\right) )\geq \alpha \log(n)^{1+\delta} $ holds on $E$ if 
	\begin{equation*}
	j - t_{k} = t_{k} + 1 - i \geq \frac{\tilde{\kappa}(\alpha,\epsilon)}{ \min(\triangle_{k},\triangle_{k}^2)}\log(n)^{1+\delta}
	\end{equation*} 
	and $j \leq t_{k+1}, i >t_{k-1}$ for all $n>2$. 
\end{Lemma}

\textbf{Proof of Lemma \ref{lemma:Mixlemma} }: This lemma shows that not having an estimated changepoint near a true changepoint leads to high costs. Let $j - t_{k} = t_{k} + 1 - i = D$. We have
\begin{align*}
&\tilde{\mathcal{C}}\left( x_{i:j} \right) -( \mathcal{C}\left( x_{i:t_{k}}\right) + \mathcal{C}\left( x_{(t_{k}+1):j}\right) ) = 2D \log \left( \frac{1}{2D\sigma_{k}\sigma_{k+1}}\sum_{t=i}^{j}(x_t-\bar{x}_{i:j})^2\right) +2D - Y_{i:j} - 2D\bar{\eta}_{i:j}.
\end{align*}
We note that $E_1$ and $E_2$ imply that
\begin{equation*}
2D - Y_{i:j} - 2D\bar{\eta}_{i:j} \geq 1 - 2\sqrt{2(2+\epsilon)D\log(n)} - (4+2\epsilon)\log(n) - (8+2\epsilon)\log(n).
\end{equation*}
This in conjunction with $E_7$ implies that $ \tilde{\mathcal{C}}\left( x_{i:j} \right) -( \mathcal{C}\left( x_{i:t_{k}}\right) + \mathcal{C}\left( x_{(t_{k}+1):j}\right) )$ is bounded below by
\begin{align*}
&2D\triangle_k +2D\log(c(D,n))- 2\sqrt{2(2+\epsilon)D\log(n)} - (4+2\epsilon)\log(n) - (8+2\epsilon)\log(n) \\
&\geq 2D\triangle_k -4(2+\epsilon)\log(n) +2D(c(D,n)-1) - 2\sqrt{2(2+\epsilon)D\log(n)} - 4(3+\epsilon)\log(n) \\
&\geq 2D\triangle_k -4(5+2\epsilon)\log(n)- 2\sqrt{2(2+\epsilon)D\log(n)} - 2\sqrt{2(2+\epsilon)D\log(n)} ,
\end{align*}
where the first inequality follows from $E_7$ and the second one from Lemma \ref{lemma:c}.
Writing the above lower bound as 
\begin{equation*}
D\triangle_k -4(5+2\epsilon)\log(n)  + \left(\sqrt{D\triangle_k^2} - 4\sqrt{2(2+\epsilon)\log(n)}\right)\sqrt{D} 
\end{equation*}
proves the result for 
\begin{equation*}
\tilde{\kappa}(\alpha,\epsilon) = a + 4\sqrt{2(2+\epsilon)}.
\end{equation*}

\begin{Lemma}
	There exists a constants $\tilde{K}_8$, $D_1$, and $D_2$ such that for large enough $n$
	\begin{equation*}
	\mathbb{P}\left(|\hat{\mu}-\mu_0| \leq D_1\sigma_0\sqrt{\frac{\log(n)}{n}}, \left|\frac{\hat{\sigma}^2}{\sigma_0^2}-1\right| \leq D_2\sqrt{\frac{\log(n)}{n}}\right)
	\geq 1 - \tilde{K}_8n^{-\epsilon}
	\end{equation*}
\end{Lemma}

\textbf{Proof of Lemma \ref{lemma:Robuststats}:} Without loss of generality, we assume that $\mu_0 = 0$ and $\sigma_0 = 1$. Since $\hat{\mu}$ and $\hat{\sigma}$ only depend upon $x_{(0.25n)}$,$x_{(0.5n)}$, and $x_{(0.75n)}$ it is sufficient to show that there exists a constant $D_3$ such that
\begin{equation*}
\mathbb{P}\left(|x_{(cn)} - q_c| < D_3 \sqrt{\frac{\log(n)}{n}}\right) \geq 1-n^{-\epsilon},
\end{equation*}
where $q_c$ is the $c$th quantile of the normal, holds for $c = 0.25,0.5,0.75$. 

In order to do so, we first define $y_{(i)}$ to be the $i$th largest observation belonging to the typical distribution. We note that $y_{(cn - m)}<x_{(cn)}<y_{(cn + m)}$, where $m = O(K\sqrt{n})$ is the number of points belonging to one of the anomalous windows. Since $q_{(cn \pm m)/(n-m)} - q_{c} = O(Kn^{-\frac{1}{2}})$, it is sufficient to show that there exists a constant $D_4$ such that
\begin{equation*}
\mathbb{P}\left( |y_{(a(n - m))} - q_a| \leq D_4 \sqrt{\frac{\log(n)}{n}} \right) \geq 1 - n^{-\epsilon}
\end{equation*}
for $a = (cn \pm m)/(n-m)$. We note that 
\begin{equation*}
y_{(a(n - m))} \sim \Phi^{-1}\Big(U_{(a(n - m)),(n - m)}\Big),
\end{equation*}
where $\Phi$ is the CDF of the normal distribution and $U_{s,t}$ the $s$th largest of $t$ i.i.d.\ $U(0,1)$ random variables. The following concentration inequality (\cite{reiss2012approximate}) applies to the uniform distribution  
\begin{equation*}
\mathbb{P}\left( \frac{\sqrt{n}}{v}\left| U_{r,n} - \frac{r}{n}\right| > t \right) \leq 
\exp \left( - \frac{t^2}{3(1+ \frac{t}{v}\sqrt{n})}\right),
\end{equation*}
where $v^2 = (r/n)\left( 1 - r/n\right) \leq 1/4$ by the AMGM inequality. This means that the event  
\begin{equation*}
\left\{ \left| U_{a(n - m),(n-m)}  - a \right| \leq  \sqrt{\epsilon} \sqrt{\frac{\log(n)}{n}}   \right\}
\end{equation*}
for the six values of $a$ which are of interest to us holds with probability at least 
\begin{equation*}
1 - 6 \exp\left( - \epsilon\log(n) \left( \frac{3}{4} + 3\sqrt{\frac{\epsilon \log(n)}{n}} \right) ^{-1} \right), 
\end{equation*}
by a Bonferroni correction, which is $1 - O(n^{-\epsilon})$. We note that this event implies that
\begin{equation*}
\left| \Phi(y_{a(n - m)})  - a \right| = O \left( \sqrt{\frac{\log(n)}{n}}  \right)
\end{equation*}
holds for all six $a$ of interest, which will be confined to the interval $[0.1,0.9]$ for large enough $n$. Hence we must also have
\begin{equation*}
\left|  \Phi^{-1}(\Phi(y_{a(n - m)})) -\Phi^{-1}(a)  \right| = O\left(\sqrt{\frac{\log(n)}{n}}\right)
\end{equation*}
for large enough $n$. This finishes our proof. 

\begin{Lemma}
	There exists a constant $\tilde{C}_7$ such that given $E$ and $E_8$ and n large enough we have:
	\begin{equation*}
	\tilde{\mathcal{C}}_E \left( x_{1:n} , \alpha , \hat{\mu} , \hat{\sigma} \right) - \mathcal{C}_E \left( x_{1:n} , \alpha , \mu_0 , \sigma_0 \right) \leq \tilde{C}_7\log(n).
	\end{equation*}	
\end{Lemma}

\textbf{Proof of Lemma \ref{lemma:CostoftruevsCostofFalse}:} First of all we note that
\begin{align*}
&\mathcal{C}_E \left( x_{1:n} , \alpha , \hat{\mu} , \hat{\sigma} \right) - \mathcal{C}_E \left( x_{1:n} , \alpha , \mu_0 , \hat{\sigma} \right) = \frac{1}{\hat{\sigma}^2}\sum_{i=1}^{K+1} \sum_{t=e_i+1}^{s_{i+1}}\left[   (x_t -  \hat{\mu})^2 - (x_t -  \mu_0)^2  \right]
\\
& \leq \frac{1}{\hat{\sigma}^2} \sum_{i=1}^{K+1} \left[ (s_{i+1}- e_i) (\hat{\mu} - \mu_0)^2 +  2\sigma_0(s_{i+1}- e_i)|\bar{\eta}_{(s_{i+1}+1):e_i} | |(\hat{\mu} - \mu_0)|   \right] \\
& \leq \frac{2}{\sigma_0^2} n (\hat{\mu} - \mu_0)^2 + \frac{2}{\sigma_0}|(\hat{\mu} - \mu_0)| \sum_{i=1}^{K+1} \sqrt{(s_{i+1}- e_i) (4+\epsilon)\log(n)} \leq 2D_1^2 \log(n) + 2(2+\epsilon)(K+1)D_1\log(n),
\end{align*}
where the second inequality follows from $E_1$ and the third from $E_8$. Moreover, 
\begin{align*}
&\mathcal{C}_E \left( x_{1:n} , \alpha , \mu_0, \hat{\sigma} \right) - \mathcal{C}_E \left( x_{1:n} , \alpha , \mu_0 , \sigma_0 \right) =  \sum_{i=1}^{K+1} \sum_{t=e_i+1}^{s_{i+1}}\left[  \log(\hat{\sigma}^2) - \log(\sigma_0^2) + \left(\frac{1}{\hat{\sigma}^2} - \frac{1}{\sigma_0^2}\right)(x_t -  \mu_0)^2  \right]  \\
&= \sum_{i=1}^{K+1}\left[ - (s_{i+1}- e_i)\log\left( \frac{\sigma_0^2}{\hat{\sigma}^2}\right) + \left( \frac{\sigma_0^2}{\hat{\sigma}^2}-1 \right)\sum_{t=e_i+1}^{s_{i+1}}\eta_t^2\right] \\
&\leq \sum_{i=1}^{K+1}\left[-(s_{i+1}- e_i)\left[\left( \frac{\sigma_0^2}{\hat{\sigma}^2}-1\right) + O\left(\left(\frac{\sigma_0^2}{\hat{\sigma}^2}-1\right)^2 \right)\right] + \left( \frac{\sigma_0^2}{\hat{\sigma}^2}-1 \right)\left(Y_{(e_i+1):(s_{i+1})} + (4+\epsilon)\log(n)\right)\right] \\
& \leq \sum_{i=1}^{K+1}\left[\left( \frac{\sigma^2_0}{\hat{\sigma}^2}-1 \right)\left(Y_{(e_i+1):(s_{i+1})} - (s_{i+1}- e_i) \right)   +O(\log(n))\right] = O(K\log(n)),
\end{align*}
where the first inequality follows from expanding $\log(x)$ around $x=1$ and $E_1$, while the second inequality uses $E_8$ and $E_2$.

\begin{Lemma}
	There exists a constant $\tilde{K}_9$ such that $\mathbb{P}(E_9) \geq 1 - \tilde{K}_9n^{-\epsilon}$
\end{Lemma}
\textbf{Proof of Lemma \ref{lemma:Probabilitylength1}:} Let $k'=k \pm 1$. Clearly, $x_t - \mu_{k'} \sim N(\mu_k - \mu_{k'},\sigma_k^2)$. Consequently, 
\begin{align*}
\mathbb{P}\left( |x_t - \mu_{k'}| < n^{-(2+\epsilon)}\sigma_k\right) \leq  2n^{-(2+\epsilon)}\sigma_k \sqrt{\frac{1}{2\pi\sigma_k^2}} = \sqrt{\frac{2}{\pi}}n^{-(2+\epsilon)}.
\end{align*}
A Bonferroni correction therefore gives $\mathbb{P}(E_9) > 1 - \sqrt{\frac{8}{\pi}}n^{-\epsilon}$.

\begin{Lemma}
	There exists a constant $\tilde{C}_2'$ such that if $i,j$ are such that there exists some $k$ such that $t_{k-1} < i \leq j \leq t_{k}$, then given $E \cap E_7$ and $n$ large enough 
	\begin{equation*}
	\mathcal{C}\left(x_{i:j}\right) - \tilde{\mathcal{C}}\left(x_{i:j}\right) \leq \tilde{C}_2' \log(n).
	\end{equation*}  
\end{Lemma}
\noindent to also account for the newly added segments of length one. 

\textbf{Proof of Lemma \ref{lemma:FalsepositiveremovalSPECIAL}:} We have to consider two cases: 

\textbf{Case 1}: $i < j$. The result holds by Lemma \ref{lemma:Falsepositiveremoval}.

\textbf{Case 2}: $i = j$, with the proxy for segments of length one. We have: 
\begin{align*}
\mathcal{C}\left(x_{i:j}\right) - \tilde{\mathcal{C}}\left(x_{i:j}\right) &=\log(\sigma_k^2) +\eta_i^2 - \log(\tilde{\sigma}^2) - \frac{(x_i - \tilde{\mu})^2}{\tilde{\sigma}^2}  \leq (4+\epsilon)\log(n) + \log\left( \frac{\sigma_k^2}{\tilde{\sigma}^2}\right) - \frac{(x_i - \tilde{\mu})^2}{\tilde{\sigma}^2},
\end{align*}
where the inequality follows from $E_1$. We now bound the above for all choices of $\tilde{\mu}$ and $\tilde{\sigma}$. First of all we consider the case $|\tilde{\mu}- \mu_k| < D_1\sqrt{\frac{\log(n)}{n}} \sigma_k$ and $|\frac{\tilde{\sigma}^2}{\sigma_k^2}-1| < D_2\sqrt{\frac{\log(n)}{n}}$. Then for large enough $n$:
\begin{align*}
\log\left( \frac{\sigma_k^2}{\tilde{\sigma}^2}\right) - \frac{(x_i - \tilde{\mu})^2}{\tilde{\sigma}^2} \leq \log \left( 1 + 2D_2\sqrt{\frac{\log(n)}{n}}\right) \leq 2D_2\sqrt{\frac{\log(n)}{n}}.
\end{align*}

Next we consider the cases $|\tilde{\mu}- \mu_{k'}| < D_1 \sqrt{\frac{\log(n)}{n}} \sigma_{k'}$ and $\left|\frac{\tilde{\sigma}^2}{\sigma_{k'}^2}-1\right| < D_2 \sqrt{\frac{\log(n)}{n}}$, where $k' = k+1$ or $k' = k-1$. We have: 
\begin{align*}
\log\left( \frac{\sigma_k^2}{\tilde{\sigma}^2}\right) - \frac{(x_i - \tilde{\mu})^2}{\tilde{\sigma}^2} \leq 2D_2\sqrt{\frac{\log(n)}{n}} + \log\left( \frac{\sigma_k^2}{\sigma_{k'}^2}\right) - \frac{(x_i - \tilde{\mu})^2}{2\sigma_{k'}^2}
\end{align*}
for large enough $n$. If $\frac{\sigma_k^2}{\sigma_{k'}^2} < n^4$ the above is bounded by $5 \log(n)$ for large enough $n$. Otherwise we have: 
\begin{align*}
&\log\left( \frac{\sigma_k^2}{\sigma_{k'}^2}\right) - \frac{(x_i - \tilde{\mu})^2}{2\sigma_{k'}^2} \leq \log\left( \frac{\sigma_k^2}{\sigma_{k'}^2}\right)  +  \frac{(|x_i-\mu_{k'}| - |\mu_{k'} - \mu|)^2_0}{2\sigma_{k'}^2}  \\
&\leq \log\left( \frac{\sigma_k^2}{\sigma_{k'}^2}\right) - \frac{1}{2}\left( \frac{\sigma_k}{\sigma_{k'}}n^{-(2+\epsilon)}  - D_1\sqrt{\frac{\log(n)}{n}} \right)^2_0 
\leq \log\left( \frac{\sigma_k^2}{\sigma_{k'}^2}\right) - \frac{1}{8}\left( \frac{\sigma_k}{\sigma_{k'}}n^{-(2+\epsilon)} \right)^2  \\
&\leq \log(8n^{4+2\epsilon}) - 1  = (4+2\epsilon)\log(n) + \log(8) - 1.
\end{align*}

\textbf{Case 3}: $i = j$, with the proxy for epidemic changes. We have: 
\begin{align*}
\mathcal{C}\left(x_{i:j}\right) - \tilde{\mathcal{C}}\left(x_{i:j}\right) &=\log(\sigma_k^2) +\eta_i^2 - \log(\tilde{\sigma}^2 \gamma + (x_i - \tilde{\mu})^2) -1  \\
&\leq (4+\epsilon)\log(n) - \log \left( \frac{\tilde{\sigma}^2}{\sigma_k^2}\gamma + \frac{(x_i - \tilde{\mu})^2}{\sigma_k^2}\right), 
\end{align*}
by $E_1$. We again bound the above for all choices of $\tilde{\mu}$ and $\tilde{\sigma}$. First of all we consider the case $|\tilde{\mu}- \mu_k| < D_1\sqrt{\frac{\log(n)}{n}} \sigma_k$ and $\left|\frac{\tilde{\sigma}^2}{\sigma_k^2}-1 \right| < D_2\sqrt{\frac{\log(n)}{n}}$. Then, for large enough $n$
\begin{equation*}
- \log \left( \frac{\tilde{\sigma}^2}{\sigma_k^2}\gamma + \frac{(x_i - \tilde{\mu})^2}{\sigma_k^2}\right) \leq - \log \left( \left[ 1 - D_2 \sqrt{\frac{\log(n)}{n}} \right]\gamma \right) \leq -\log(\gamma) + 2D_2 \sqrt{\frac{\log(n)}{n}}
\end{equation*}

Next, we consider the cases $|\tilde{\mu}- \mu_{k'}| < D_1\sqrt{\frac{\log(n)}{n}} \sigma_{k'}$ and $|\frac{\tilde{\sigma}^2}{\sigma_{k'}^2}-1| < D_2\sqrt{\frac{\log(n)}{n}}$, where $k' = k+1$ or $k' = k-1$. We have 
\begin{align*}
- \log \left( \frac{\tilde{\sigma}^2}{\sigma_k^2}\gamma + \frac{(x_i - \tilde{\mu})^2}{\sigma_k^2}\right) &\leq -\log\left( \frac{\tilde{\sigma}^2}{\sigma_{k'}^2}\right)  - \log \left( \frac{\sigma_{k'}^2}{\sigma_k^2}\gamma + \frac{\sigma_{k'}^2}{\tilde{\sigma}^2}\frac{(x_i - \tilde{\mu})^2}{\sigma_k^2}\right) \\
&\leq 2 D_2 \sqrt{\frac{\log(n)}{n}} - \log \left( \frac{\sigma_{k'}^2}{\sigma_k^2}\gamma + \frac{1}{2}\frac{(|x_i-\mu_{k'}| - |\mu_{k'} - \mu|)^2}{\sigma_k^2}\right)
\end{align*}
for large enough $n$. If $\frac{\sigma_k^2}{\sigma_{k'}^2} < n^4$ the above is bounded by $4 \log(n) - \log(\gamma)$. Otherwise we have: 
\begin{align*}
&- \log \left( \frac{\sigma_{k'}^2}{\sigma_k^2}\gamma + \frac{1}{2}\frac{(|x_i-\mu_{k'}| - |\mu_{k'} - \mu|)^2_0}{\sigma_k^2}\right) \leq - \log \left( \frac{\sigma_{k'}^2}{\sigma_k^2}\gamma + \frac{1}{2}\left(n^{-2+\epsilon} - D_1 \sqrt{\frac{\log(n)}{n}} \frac{\sigma_{k'}}{\sigma_k}  \right)^2_0\right) \\
& \leq - \log \left( \frac{1}{2}\left(n^{-2+\epsilon} - D_1 \sqrt{\frac{\log(n)}{n}} n^{-2} \right)^2_0\right)
\leq - \log \left( \frac{1}{8}\left(n^{-2+\epsilon}\right)^2\right) = \log(8) + (4+2\epsilon) \log(n),
\end{align*}
for large enough $n$. This finishes the proof.

\subsection{Further Simulation Study Results}

\begin{figure} 
	\begin{center}
		\begin{tabular}{||c c c ||c | c | c | c ||} 
			\hline
			Mean & Variance & Point anomalies & CAPA & PELT & BreakoutDetection & luminol  \\ [0.5ex] 
			\hline\hline
			weak   & -      & 10 strong & \textbf{1.71} & 2.98 & 3.70 & 9.91 \\
			\hline
			strong & -      & 10 strong & \textbf{0.18} & 0.72 & 5.50 & 10.33\\ 
			\hline
			-      & weak   & 10 strong & \textbf{1.26} & 2.15 & 4.59 & 9.99\\
			\hline
			-      & strong & 10 strong & \textbf{0.32} & 0.80 & 4.98 & 9.65 \\ 
			\hline
			weak   & weak   & 10 strong & \textbf{1.19} & 1.76 & 4.26 & 10.12\\
			\hline
			strong & strong & 10 strong & \textbf{0.09} & 0.57 & 3.98 & 9.91 \\ 
			\hline
		\end{tabular}
		\caption{Precision of true positives measured in mean absolute distance for CAPA, PELT, luminol, and BreakoutDetection when strong poit anomalies are present} 
		\label{tble:Precisionstronganomalies}
	\end{center}
\end{figure}

\begin{figure} 
	\begin{subfigure}[b]{0.5\linewidth}
		\centering
		\includegraphics[width=0.9\linewidth]{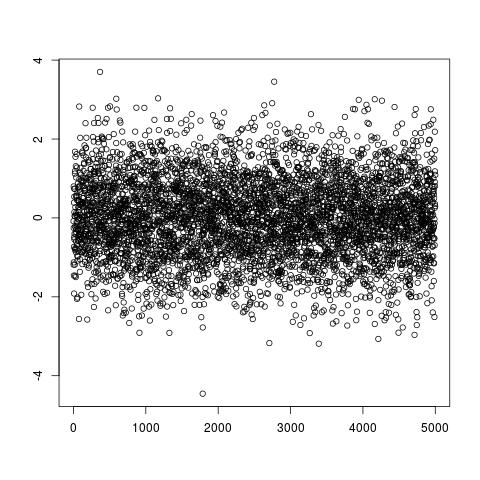} 
		\caption{No point anomalies}
		\label{fig:varchange_graph} 
		\vspace{4ex}
	\end{subfigure} 
	\begin{subfigure}[b]{0.5\linewidth}
		\centering
		\includegraphics[width=0.9\linewidth]{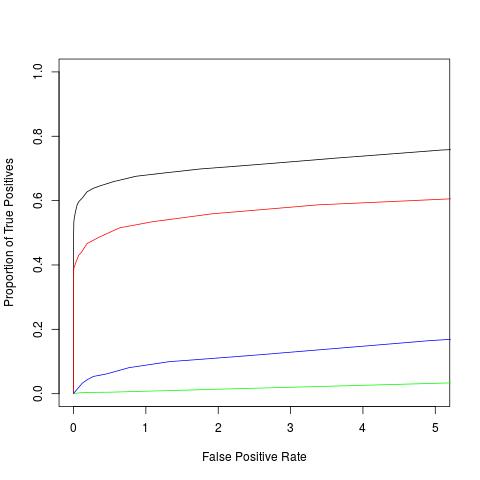} 
		\caption{No point anomalies} 
		\label{fig:varchange_ROC} 
		\vspace{4ex}
	\end{subfigure} 
	\begin{subfigure}[b]{0.5\linewidth}
		\centering
		\includegraphics[width=0.9\linewidth]{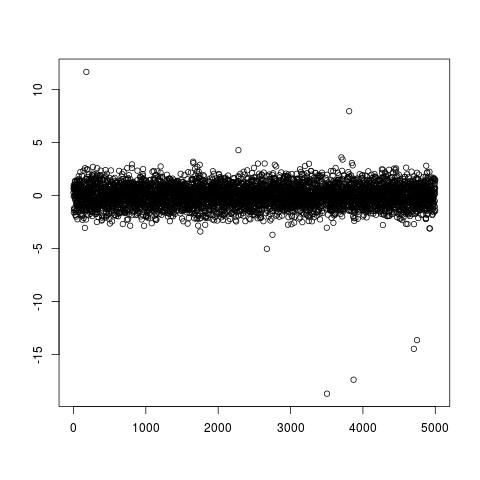} 
		\caption{Point anomalies present} 
		\vspace{4ex}
	\end{subfigure} 
	\begin{subfigure}[b]{0.5\linewidth}
		\centering
		\includegraphics[width=0.9\linewidth]{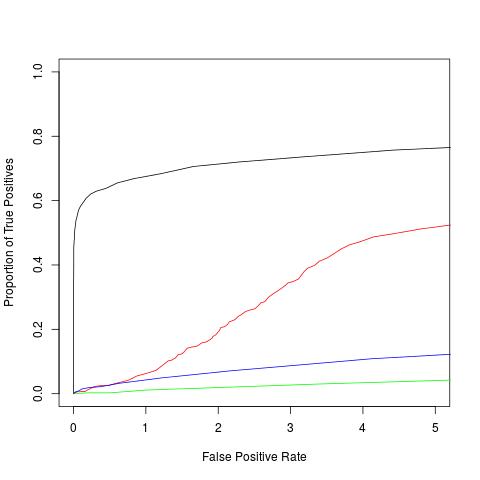} 
		\caption{Point anomalies present} 
		\vspace{4ex}
	\end{subfigure}
	\caption{Data examples and ROC curves for weak changes in variance for CAPA (black), PELT (red), BreakoutDetection (green), and luminol (blue).}
	\label{fig:var_weak} 
\end{figure}

\begin{figure} 
	\begin{subfigure}[b]{0.5\linewidth}
		\centering
		\includegraphics[width=0.9\linewidth]{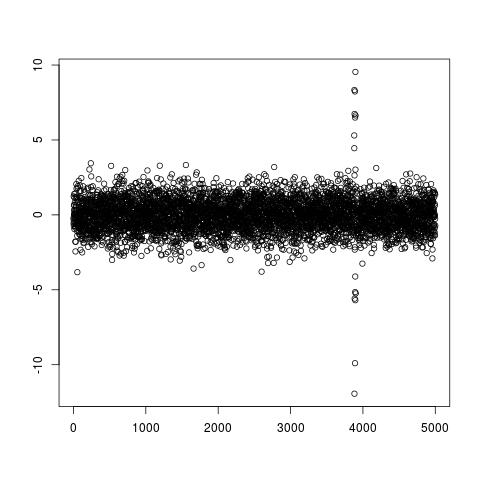} 
		\caption{No point anomalies}
		\label{fig:VARchange_graph} 
		\vspace{4ex}
	\end{subfigure} 
	\begin{subfigure}[b]{0.5\linewidth}
		\centering
		\includegraphics[width=0.9\linewidth]{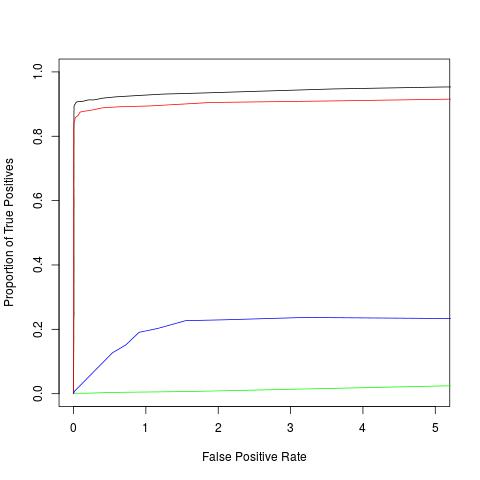} 
		\caption{No point anomalies} 
		\label{fig:VARchange_ROC} 
		\vspace{4ex}
	\end{subfigure} 
	\begin{subfigure}[b]{0.5\linewidth}
		\centering
		\includegraphics[width=0.9\linewidth]{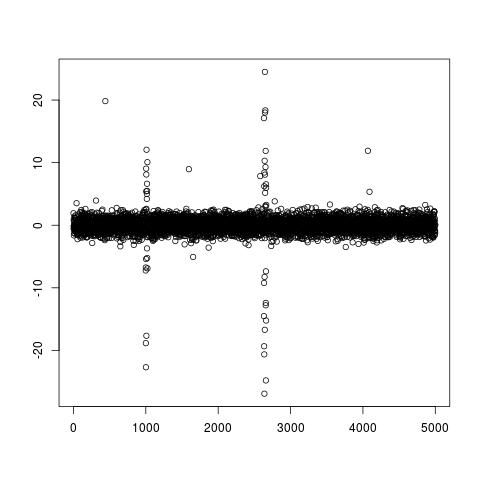} 
		\caption{Point anomalies present} 
		\label{fig:VARANOMchange_graph} 
		\vspace{4ex}
	\end{subfigure} 
	\begin{subfigure}[b]{0.5\linewidth}
		\centering
		\includegraphics[width=0.9\linewidth]{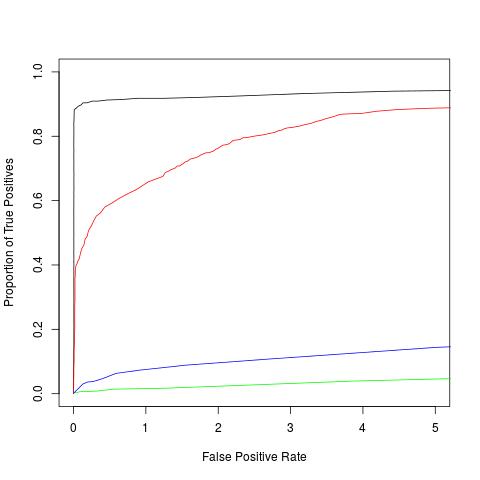}
		\caption{Point anomalies present} 
		\label{fig:VARANOMchange_ROC} 
		\vspace{4ex}
	\end{subfigure}
	\caption{Data examples and ROC curves for strong changes in variance for CAPA (black), PELT (red), BreakoutDetection (green), and luminol (blue).}
	\label{fig:VAR} 
\end{figure}

\begin{figure} 
	\begin{subfigure}[b]{0.5\linewidth}
		\centering
		\includegraphics[width=0.9\linewidth]{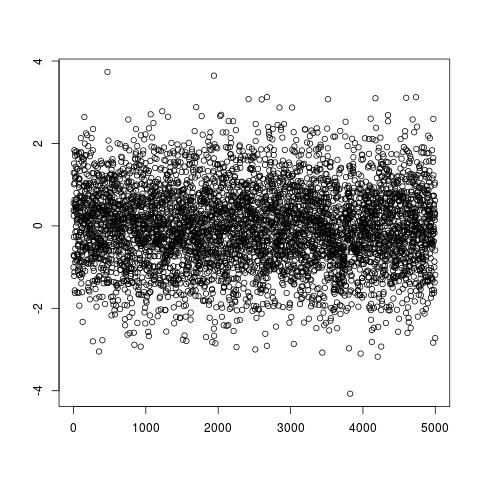} 
		\caption{No point anomalies} 
		\label{fig:meanvarchange_graph} 
		\vspace{4ex}
	\end{subfigure} 
	\begin{subfigure}[b]{0.5\linewidth}
		\centering
		\includegraphics[width=0.9\linewidth]{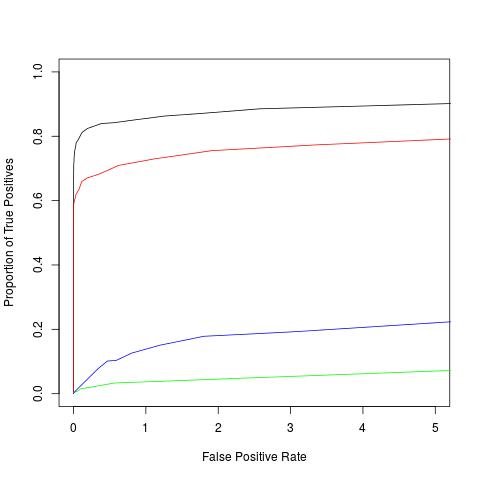} 
		\caption{No point anomalies} 
		\label{fig:meanvarchange_ROC} 
		\vspace{4ex}
	\end{subfigure} 
	\begin{subfigure}[b]{0.5\linewidth}
		\centering
		\includegraphics[width=0.9\linewidth]{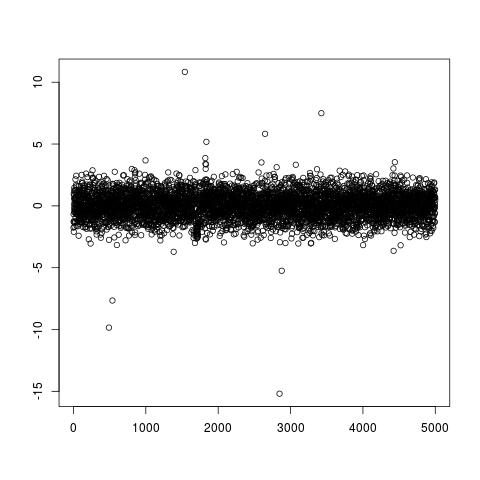} 
		\caption{Point anomalies present} 
		\label{fig:meanvarANOMchange_graph} 
		\vspace{4ex}
	\end{subfigure} 
	\begin{subfigure}[b]{0.5\linewidth}
		\centering
		\includegraphics[width=0.9\linewidth]{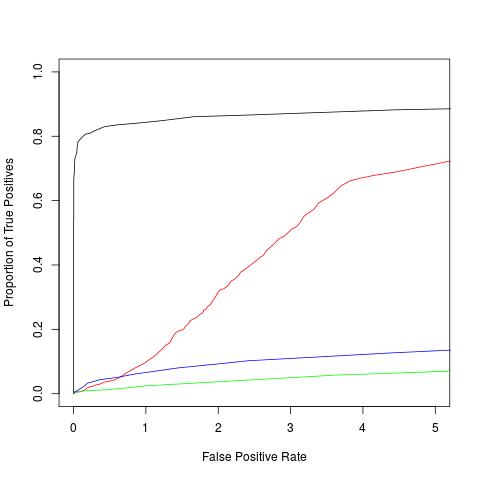} 
		\caption{Point anomalies present} 
		\label{fig:meanvarANOMchange_ROC} 
		\vspace{4ex}
	\end{subfigure}
	\caption{Data examples and ROC curves for weak changes in mean and variance for CAPA (black), PELT (red), BreakoutDetection (green), and luminol (blue).}
	\label{fig:meanvar_weak} 
\end{figure}

\begin{figure} 
	\begin{subfigure}[b]{0.5\linewidth}
		\centering
		\includegraphics[width=0.9\linewidth]{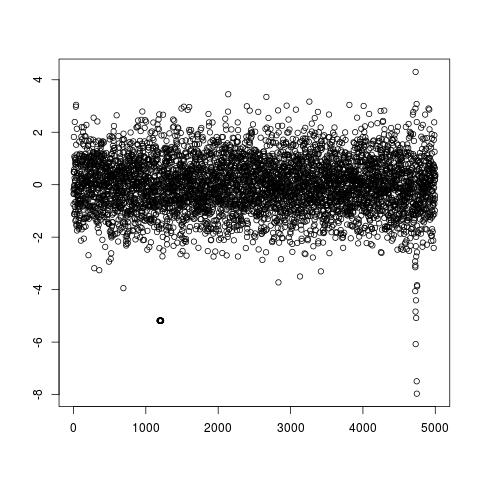} 
		\caption{No point anomalies} 
		\label{fig:MEANVARchange_graph} 
		\vspace{4ex}
	\end{subfigure} 
	\begin{subfigure}[b]{0.5\linewidth}
		\centering
		\includegraphics[width=0.9\linewidth]{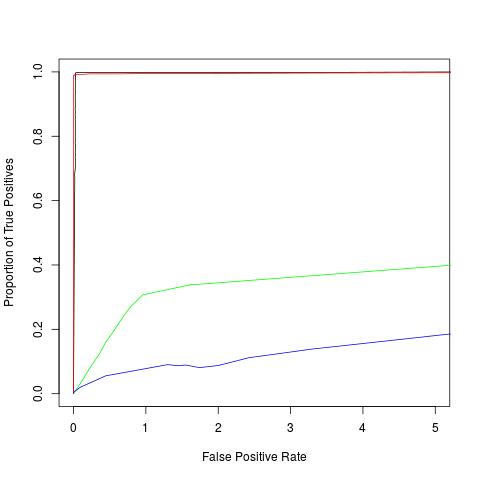} 
		\caption{No point anomalies}
		\label{fig:MEANVARchange_ROC} 
		\vspace{4ex}
	\end{subfigure} 
	\begin{subfigure}[b]{0.5\linewidth}
		\centering
		\includegraphics[width=0.9\linewidth]{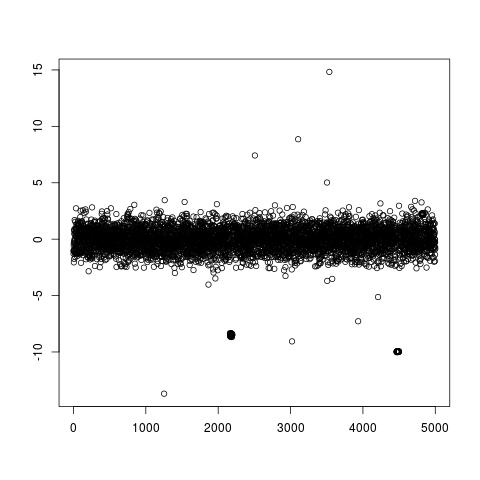} 
		\caption{Point anomalies present} 
		\label{fig:MEANVARANOMchange_graph} 
		\vspace{4ex}
	\end{subfigure} 
	\begin{subfigure}[b]{0.5\linewidth}
		\centering
		\includegraphics[width=0.9\linewidth]{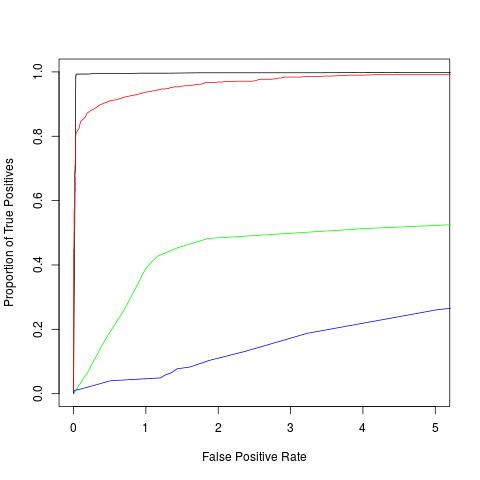} 
		\caption{Point anomalies present} 
		\label{fig:MEANVARANOMchange_ROC} 
		\vspace{4ex}
	\end{subfigure}
	\caption{Data examples and ROC curves for strong changes in mean and variance for CAPA (black), PELT (red), BreakoutDetection (green), and luminol (blue).}
	\label{fig:MEANVAR} 
\end{figure}

\begin{figure} 
	\begin{subfigure}{\linewidth}
		\begin{subfigure}[b]{0.5\linewidth}
			\centering
			\includegraphics[width=0.9\linewidth]{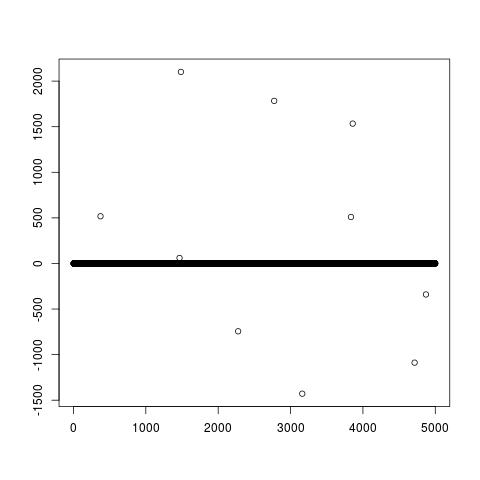} 
			\label{fig:varANOMchange_graph} 
			\caption{Weak changes}
		\end{subfigure} 
		\begin{subfigure}[b]{0.5\linewidth}
			\centering
			\includegraphics[width=0.9\linewidth]{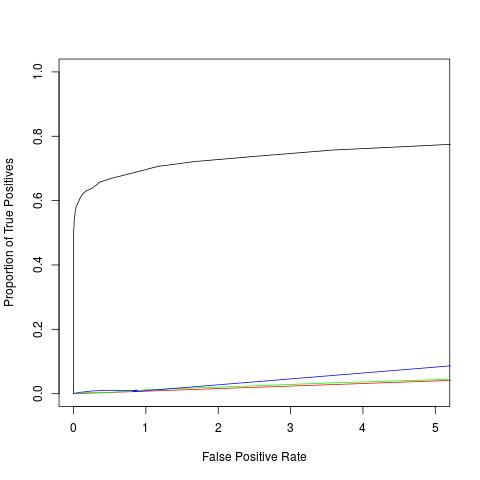} 
			\label{fig:varANOMchange_ROC} 
			\caption{Weak changes}
		\end{subfigure} 
	\end{subfigure}
	\begin{subfigure}{\linewidth}
		\begin{subfigure}[b]{0.5\linewidth}
			\centering
			\includegraphics[width=0.9\linewidth]{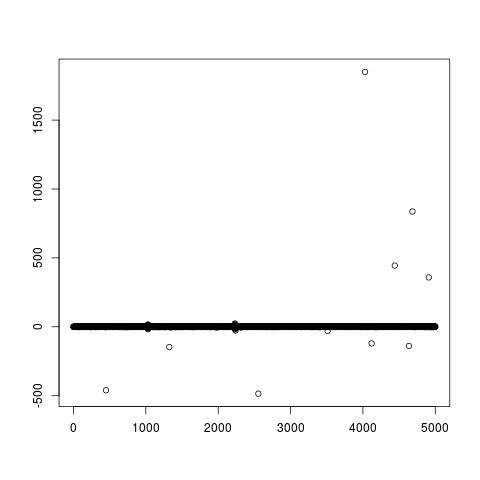} 
			\label{fig:varANOMchange_graph} 
			\caption{Strong changes}
		\end{subfigure} 
		\begin{subfigure}[b]{0.5\linewidth}
			\centering
			\includegraphics[width=0.9\linewidth]{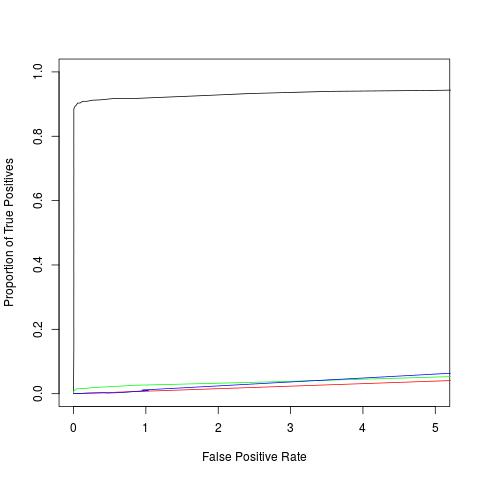} 
			\label{fig:varANOMchange_ROC} 
			\caption{Strong changes}
		\end{subfigure} 
	\end{subfigure}
	\caption{Data examples and effect of strong point anomalies on ROC curves for the detection of changes in variance of CAPA (black), PELT (red), BreakoutDetection (green), and luminol (blue).}
	\label{fig:VARANOM} 
\end{figure}

\begin{figure} 
	\begin{subfigure}{\linewidth}
		\begin{subfigure}[b]{0.5\linewidth}
			\centering
			\includegraphics[width=0.9\linewidth]{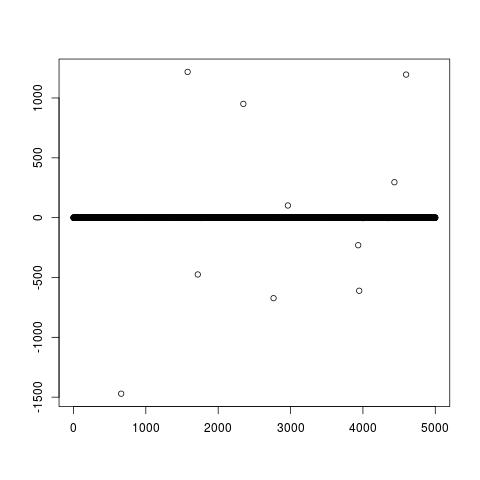} 
			\caption{Weak changes}
		\end{subfigure} 
		\begin{subfigure}[b]{0.5\linewidth}
			\centering
			\includegraphics[width=0.9\linewidth]{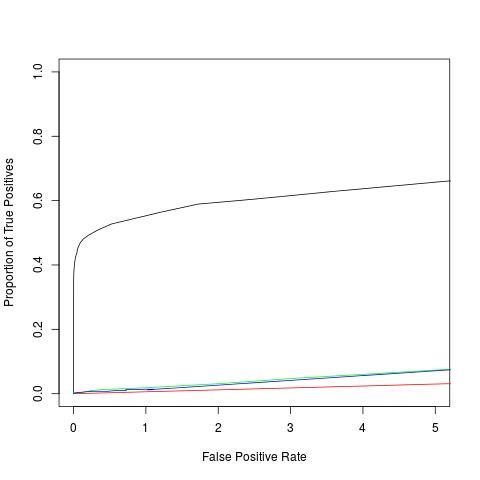} 
			\caption{Weak changes}
		\end{subfigure} 
	\end{subfigure}
	\begin{subfigure}{\linewidth}
		\begin{subfigure}[b]{0.5\linewidth}
			\centering
			\includegraphics[width=0.9\linewidth]{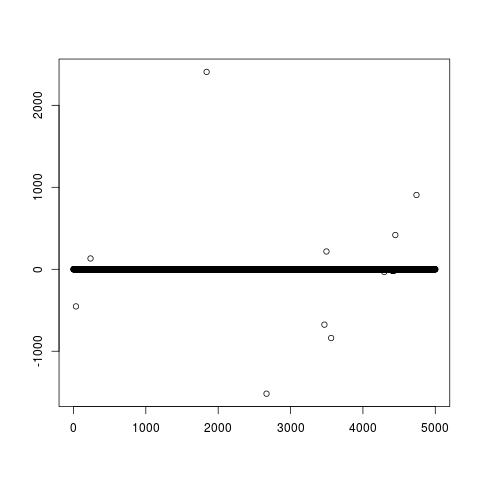} 
			\caption{Strong changes}
		\end{subfigure} 
		\begin{subfigure}[b]{0.5\linewidth}
			\centering
			\includegraphics[width=0.9\linewidth]{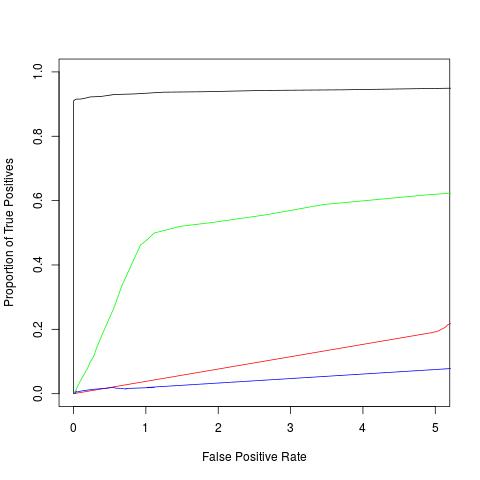} 
			\caption{Strong changes}
		\end{subfigure} 
	\end{subfigure}
	\caption{Data examples and effect of strong point anomalies on ROC curves for the detection of changes in mean of CAPA (black), PELT (red), BreakoutDetection (green), and luminol (blue).}
	\label{fig:MEANANOM} 
\end{figure}

\begin{figure} 
	\begin{subfigure}{\linewidth}
		\begin{subfigure}[b]{0.5\linewidth}
			\centering
			\includegraphics[width=0.9\linewidth]{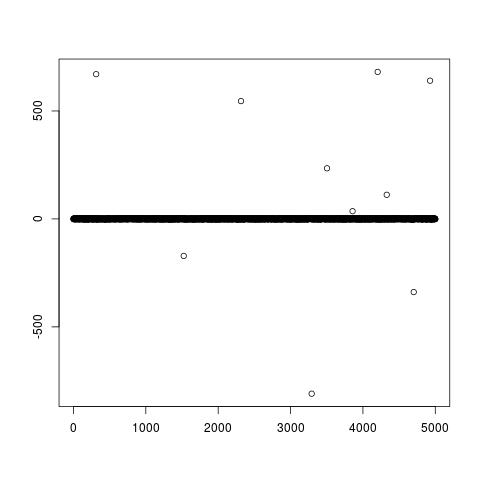}
			\caption{Weak changes}
		\end{subfigure} 
		\begin{subfigure}[b]{0.5\linewidth}
			\centering
			\includegraphics[width=0.9\linewidth]{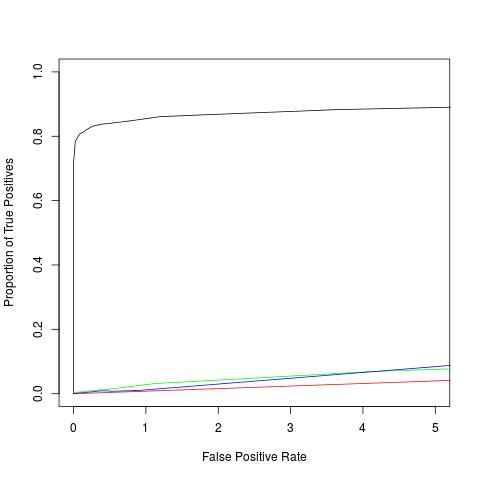} 
			\caption{Weak changes}
		\end{subfigure} 
	\end{subfigure}
	\begin{subfigure}{\linewidth}
		\begin{subfigure}[b]{0.5\linewidth}
			\centering
			\includegraphics[width=0.9\linewidth]{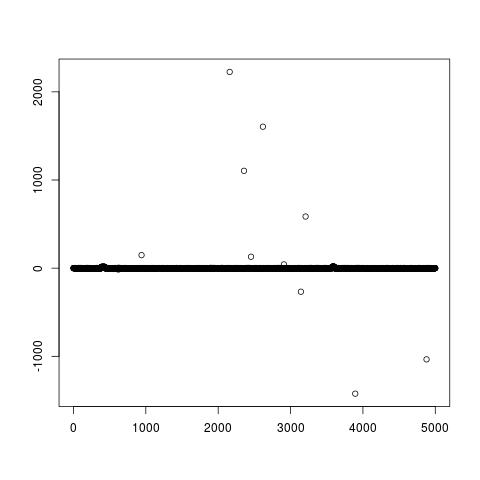} 
			\caption{Strong changes}
		\end{subfigure} 
		\begin{subfigure}[b]{0.5\linewidth}
			\centering
			\includegraphics[width=0.9\linewidth]{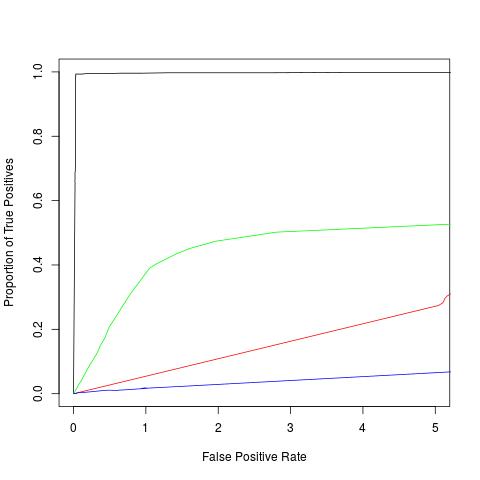} 
			\caption{Strong changes}
		\end{subfigure} 
	\end{subfigure}
	\caption{Data examples and effect of strong point anomalies on ROC curves for the detection of changes in mean and variance of CAPA (black), PELT (red), BreakoutDetection (green), and luminol (blue).}
	\label{fig:MEANVARANOM} 
\end{figure}

The ROC curves for weak and strong changes in variance as well as for weak and strong joint changes in mean and variance can be found in Figures \ref{fig:var_weak}, \ref{fig:VAR}, \ref{fig:meanvar_weak}, and \ref{fig:MEANVAR}. The ROC curves for all three types of collective anomalies in the presence of strong point anomalies can be found in Figures \ref{fig:VARANOM}, \ref{fig:MEANANOM}, and \ref{fig:MEANVARANOM}. The precision of true positives when such point anomalies are present is compared in Figure \ref{tble:Precisionstronganomalies}. We note that CAPA is robust to such point anomalies, unlike PELT and luminol.

\subsection{Application of CAPA to Further Stars}

\begin{figure}  
	\begin{subfigure}[b]{0.5\linewidth}
		\centering
		\includegraphics[width=0.9\linewidth]{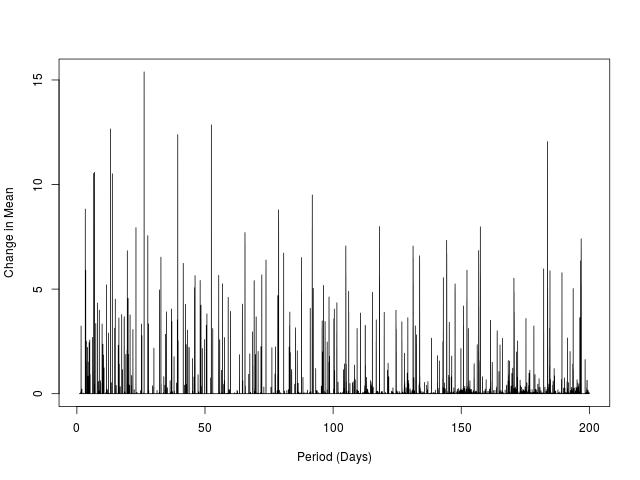} 
		\caption{Kepler 356} 
	\end{subfigure}
	\begin{subfigure}[b]{0.5\linewidth}
		\centering
		\includegraphics[width=0.9\linewidth]{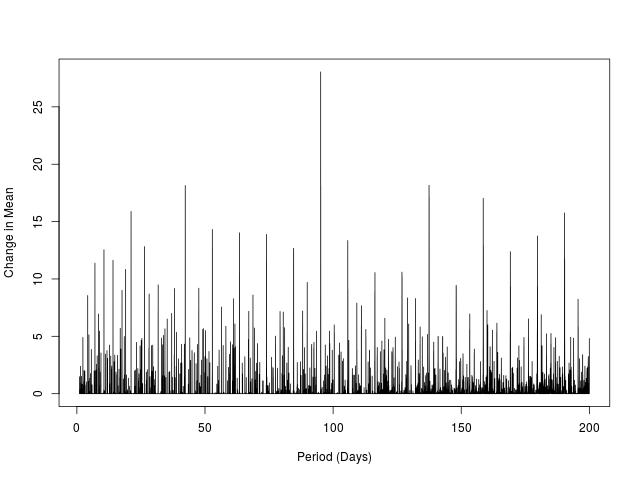} 
		\caption{Kepler 454} 
	\end{subfigure}
	\begin{subfigure}[b]{0.5\linewidth}
		\centering
		\includegraphics[width=0.9\linewidth]{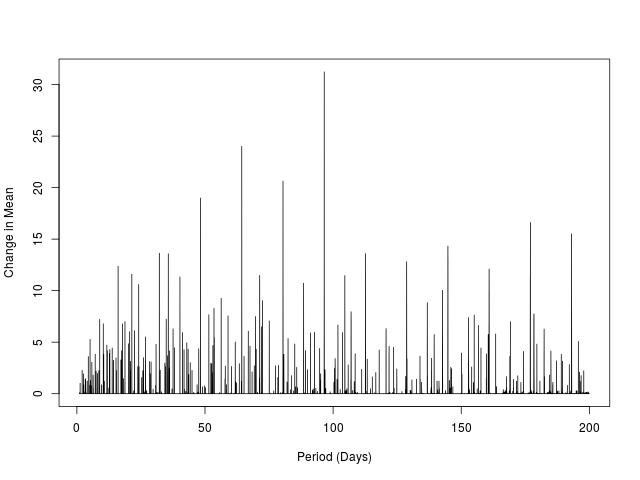} 
		\caption{Kepler 275} 
	\end{subfigure}
	\begin{subfigure}[b]{0.5\linewidth}
		\centering
		\includegraphics[width=0.9\linewidth]{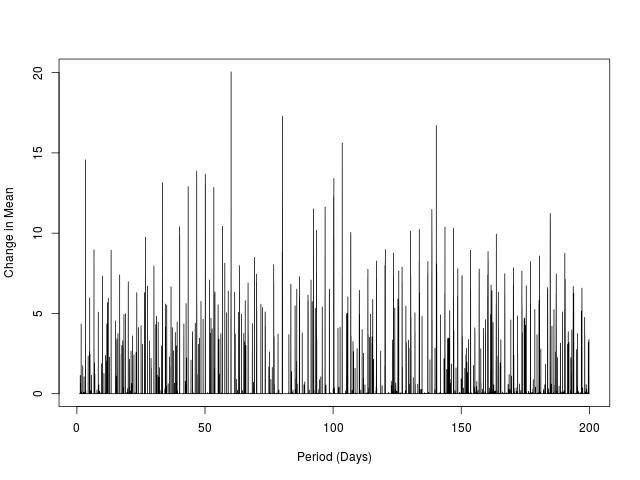} 
		\caption{Kepler 235} 
	\end{subfigure}
	\centering 
	\begin{subfigure}[b]{0.5\linewidth}
		\includegraphics[width=0.9\linewidth]{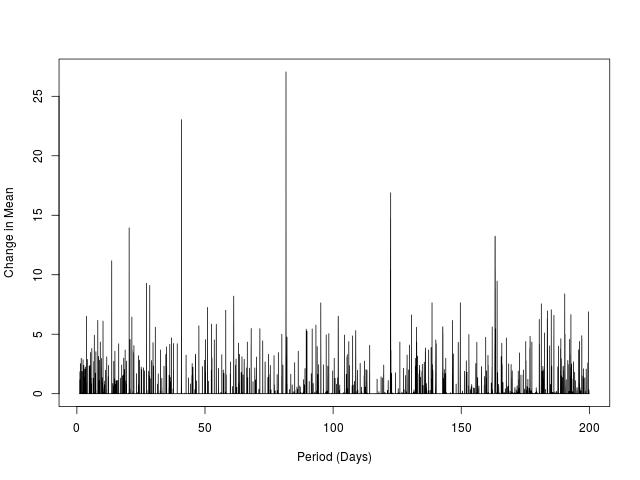} 
		\caption{Kepler 264} 
	\end{subfigure} 
	\caption{The strongest change in mean, as measured by $\max_k\left(\triangle_{\mu,k}\right)$, detected by CAPA for the lightcurves of five stars with known exoplanets. All periods from 1 to 200 days at 0.01 day increment were examined}
	\label{fig:Spectrogrammany} 
\end{figure}

\begin{figure} 
	\begin{center}
		\begin{tabular}{||c |c | c | c  ||} 
			\hline
			Star & Planet & Period & Period (or integer fraction thereof) in top 20 modes \\ [0.5ex] 
			\hline
			Kepler 275 & Kepler-275-b & 10.3007 & No   \\
			Kepler 275 & Kepler-275-c & 16.0881 & Yes  \\
			Kepler 275 & Kepler-275-d & 35.6761 & Yes  \\
			\hline
			Kepler 264 & Kepler-264-b & 40.806     & Yes \\
			Kepler 264 & Kepler 264-c & 140.101261 & No \\
			\hline
			Kepler 356 & Kepler 356-b & 13.1216 &  Yes\\
			Kepler 356 & Kepler 356-c &  4.6127 &  No \\
			\hline
			Kepler 454 & Kepler 454-b & 10.5738 & Yes\\
			Kepler 454 & Kepler 454-c & 523.90  & No \\
			\hline
			Kepler 235 & Kepler 235-b & 3.340   & Yes\\
			Kepler 235 & Kepler 235-c & 7.824   & No \\
			Kepler 235 & Kepler 235-d & 20.0605 & Yes \\
			Kepler 235 & Kepler 235-e & 46.1836 & Yes \\
			\hline
		\end{tabular}
		\caption{Five stars orbited by known exoplanets and whether their period or an integer fraction thereof was in the 20 periods with strongest change in mean according to CAPA.} 
		\label{tble:Planets}
	\end{center}
\end{figure}

We applied the approach detailed in Section 6 to the light curves of five further stars with known exoplanets (\cite{Kepler}). Figure \ref{fig:Spectrogrammany} depicts the largest detected change in mean as measured by $\max_k \left(\triangle_{\mu,k}\right)$ per period for the five stars. We found that the 20 periods exhibiting the largest change in mean correspond to integer fractions of the orbital period of a known exoplanet in all cases. We thus observed no false positives. The results are summarised in Figure \ref{tble:Planets}. We note that not all planets appear in the 20 periods with largest change in mean. This is due to the fact that their signal is weaker than the resonance of the signal of larger planets. CAPA can nevertheless detect the transit signal of the missing planet at their orbital period, with the exception of Kepler 454-c. This planet however was discovered by a different method than the transit method.

\end{document}